**EXACT: an explainable anomaly-aware vision foundation model for analysis of 3D chest CT**

Xuguang Bai[1*], Mingxuan Liu[1*†], Tongxi Song[1*], Yifei Chen[1*], Hongjia Yang[1], Kasidit Anmahapong[1], Zihan Li[1], Ying Zhou[2], Qiyuan Tian[1†]

[1] School of Biomedical Engineering, Tsinghua University, Beijing, China.

[2] Department of Radiology, Mianyang Central Hospital, Mianyang, China.

[*] These authors contributed equally: Xuguang Bai, Mingxuan Liu, Tongxi Song, Yifei Chen

[†] Corresponding authors: Qiyuan Tian, PhD, Center for Biomedical Imaging Research, Tsinghua University, 30 Shuangqing Road, Haidian District, Beijing, China, 100084 (qiyuantian@tsinghua.edu.cn); Mingxuan Liu, BEng, Center for Biomedical Imaging Research, Tsinghua University, 30 Shuangqing Road, Haidian District, Beijing, China, 100084 (arktisx@foxmail.com).

# Abstract

Chest computed tomography (CT) is central to the detection and management of thoracic disease, yet the growing scale and complexity of volumetric imaging increasingly exceed what can be addressed by scan-level prediction alone. Clinically useful AI for CT must not only recognize disease across the whole volume, but also localize abnormalities and provide interpretable visual evidence. Existing vision-language foundation models typically compress scans and reports into global image-text representations, limiting their ability to preserve spatial evidence and support clinically meaningful interpretation. Here we developed EXACT, an explainable anomaly-aware foundation model for three-dimensional chest CT that learns spatially resolved representations from paired clinical scans and radiology reports. EXACT was developed on 25,692 CT-reports pairs using anatomy-aware weak supervision, jointly learning organ segmentation and multi-instance anomaly localization without manual voxel-level annotations. The resulting organ-specific anomaly-aware maps assign each voxel a disease-specific anomaly score confined to its corresponding anatomy, jointly encoding lesion extent and organ-level context. In retrospective multinational and multi-center evaluations, EXACT showed broad and consistent improvements across clinically relevant CT tasks, spanning multi-disease diagnosis, zero-shot anomaly localization, downstream adaptation, and visually grounded report generation, outperforming existing three-dimensional medical foundation models. We further evaluated the frozen encoder on lung tumour segmentation and survival-related endpoints beyond the pre-training objectives. By transforming routine clinical CT scans and free-text reports into explainable voxel-level representations, EXACT establishes a scalable paradigm for trustworthy volumetric medical AI.

## Introduction

Chest computed tomography (CT) has become a cornerstone imaging modality for detecting and characterizing a broad spectrum of thoracic diseases. It is routinely used to evaluate chronic obstructive pulmonary disease, lower respiratory tract infections, lung cancer, cardiovascular abnormalities, and many other clinically important conditions[1–4]. Beyond primary diagnostic indications, chest CT is increasingly used for opportunistic screening and incidental disease assessment, further expanding its role in preventive and population-level healthcare[5,6]. Driven by aging populations, rising chronic disease burden, and improved access to advanced imaging, the global use of chest CT continues to increase at an estimated annual rate of 3–4%. Chest CT now accounts for approximately one quarter of the estimated 375 million CT examinations performed worldwide each year, underscoring its central position in modern clinical practice[7,8].

The growing demand for chest CT interpretation has created a widening mismatch between imaging volume and radiology workforce capacity. Current projections suggest a shortage of more than 19,000 radiologists by 2036, with workforce modeling indicating that this gap is likely to persist over the coming decades[9,10]. At the same time, each three-dimension (3D) chest CT examination contains hundreds of slices and tens of millions of voxels, requiring radiologists to integrate subtle local abnormalities with global anatomical and pathological context. This substantial cognitive burden increases the risk of missed findings, under-recognition of visible abnormalities, and communication errors, all of which remain important contributors to patient harm[11,12]. Artificial intelligence (AI) systems that can efficiently analyze volumetric CT data may therefore help alleviate radiologist workload, support timely decision-making, and improve access to expertise in resource-constrained settings.

Clinically useful chest CT AI must move beyond isolated prediction tasks toward integrated volumetric understanding, localization, and communication. To reduce diagnostic errors, such systems should simultaneously recognize multi-disease patterns across the whole volume, localize abnormal findings at the voxel level, and translate imaging evidence into standardized radiology reports with clear clinical meaning[13,14]. However, most existing models follow a “one-task, one-model" paradigm[15], in which separate networks are trained for narrow objectives such as lung nodule detection, tumor segmentation, or coronary artery calcification scoring. These task-specific models often lack the global context needed to reason over complex comorbidities and incidental findings. They also typically depend on expensive expert-curated voxel-level annotations, generalize poorly across institutions and disease distributions, and require retraining for each new task, limiting their scalability in real-world clinical deployment[10,16–18].

Foundation models offer a promising route toward general-purpose chest CT analysis by learning transferable representations from large-scale medical data. Pre-trained through self-supervised learning strategies such as CLIP[19], SimCLR[20], MAE[21], and DINO[22,23], foundation models can capture task-agnostic visual representations that are adaptable to diverse downstream applications with limited supervision. Among these approaches, contrastive language–image pre-training has become particularly influential in medical imaging because it aligns radiological images and reports in a shared embedding space, thereby exploiting routinely available free-text reports as natural supervision. Early medical foundation models, including CheXzero[24], ConVIRT[25], MedCLIP[26], MedKLIP[27], ELIXR[28], GLoRIA[29], and BiomedCLIP[30], primarily focused on 2D imaging and achieved strong performance in chest radiograph interpretation. More recently, the release of CT-RATE[31], the first large-scale open-source chest CT vision–language dataset, has accelerated the development of foundation models for volumetric CT analysis.

Despite this progress, current 3D chest CT foundation models remain limited by their reliance on alignment-based representations that discard voxel-level spatial information. CT-CLIP[31] and related models[14,32–34] trained on CT-RATE have advanced volumetric vision–language learning, but their cross-modal alignment objectives are not designed to preserve the fine-grained spatial structure required for precise localization of focal pathologies such as lung nodules or hiatal hernias. CT-CLIP aligns whole CT volumes with full radiology reports using a global contrastive objective, whereas later methods attempt to introduce finer granularity: fVLM[32] and MedVista3D[14] use organ-level masks to ground text to anatomical regions, while T3D[33] and SimCroP[34] rely on multi-view sub-volume sampling or similarity-driven patch selection to emphasize local features. Nevertheless, these models ultimately aggregate high-dimensional 3D features into compact one-dimensional embeddings through global average pooling, [CLS] tokens, or related mechanisms. This compression collapses volumetric structure and removes explicit voxel-wise positional information, forcing interpretability to depend on post-hoc visualizations such as gradient activation maps (Grad-CAM)[35] or top-K patch highlighting. However, as highlighted by a recent survey of 46 radiologists[36], these post-hoc methods provide only coarse, qualitative approximations of model attention that lack clinical fidelity, falling far short of the intrinsic, voxel-level spatial precision required for confident decision-making. Consequently, this underlying lack of fine-grained spatial grounding fundamentally constrains their transferability to spatially demanding tasks, including multi-disease diagnosis and lesion segmentation.

In this study, we present EXACT, an explainable anomaly-aware chest CT vision foundation model designed to unify global disease understanding with voxel-level visual grounding. EXACT jointly optimizes anatomical segmentation and anomaly detection through weakly supervised multi-task and multi-instance learning, enabling intrinsic spatial interpretability without manual voxel-level annotations (Fig. 1a-b). It uses a Y-shaped Mamba backbone, termed Y-Mamba, in which a shared encoder drives both an organ segmentation decoder and a multi-instance anomaly detection decoder (Supplementary Fig. 9). The model is developed end-to-end on 25,692 non-contrast chest CT scans from 21,304 patients in CT-RATE[31], using automatically derived organ masks from Segment Anything by Text[37] and report-derived disease pseudo-labels from RadBERT[38]. Unlike CLIP-based models that compress 3D features into global embeddings, EXACT directly generates voxel-level anomaly-aware maps that encode both lesion extent and organ-specific pathological context. These maps support zero-shot anomaly localization, multi-disease diagnosis through anatomy-constrained aggregation, and visually grounded report generation. We further extend EXACT into EXACT-CHAT by integrating its frozen image encoder and anomaly-aware diagnostic priors with a large language model to generate spatially grounded radiology reports while reducing factual hallucinations (Fig. 7). We additionally assessed frozen-encoder transfer to lung tumour segmentation and survival-related prediction in the NSCLC-Radiomics cohort (Fig. 6). Across multicenter and multinational cohorts, EXACT consistently outperforms state-of-the-art 3D medical foundation models in zero-shot and fine-tuned diagnosis, anomaly localization, and report generation (Fig. 1c-g), establishing anatomy-aware weak supervision as a scalable paradigm for explainable and clinically trustworthy medical AI.

# Results

## Overview

The aim of EXACT (Fig. 1a-b) is to pre-train a vision foundation model through anatomy-aware weakly supervised learning without manual voxel-level annotations, enabling the model to capture intrinsic voxel-level anomaly representations from large-scale 3D chest CT data and to establish an explainable, general-purpose fundamental basis for a wide range of subsequent medical tasks.

We evaluated EXACT in five downstream settings corresponding to major sources of radiologic diagnostic error[13,14]: multi-disease diagnosis under zero-shot (Fig. 1c) and fine-tuning (Fig. 1d) settings to overcome inattentional blindness, anomaly localization under zero-shot (Fig. 1e) and fine-tuning (Fig. 1f) settings to counter under-reading, and radiology report generation (Fig. 1g) to mitigate miscommunication of findings. A multicenter and multinational benchmark was assembled to evaluate EXACT and comparison foundation models (FMs). For multi-disease diagnosis and report generation, internal validation was performed on held-out CT-RATE dataset ($n$ = 1,564, Istanbul Medipol University Mega Hospital, Turkey)[31], with external validation on RAD-ChestCT ($n$ = 3,630; Duke University Health System, US)[39] and MianYang ($n$ = 505; Mianyang Central Hospital, China) datasets. For anomaly localization, internal validation employed the ReX-Train ($n$ = 1,102) and ReX-Val ($n$ = 157) subsets of the ReXGroundingCT dataset[40], derived from a portion of CT-RATE, with external validation on COVID-19 ($n$ = 20; from the Coronacases Initiative and Radiopaedia)[41,42] and MosMed (n = 50; Municipal Hospitals in Moscow, Russia)[43]. We additionally evaluated frozen-encoder transfer in the NSCLC-Radiomics (LUNG1) cohort[44], including few-shot lung tumour segmentation and survival-related prediction (Fig. 6; Supplementary Tables 8 and 9).

We compared EXACT with publicly available state-of-the-art (SOTA) methods across all tasks, including: (1) 3D vision FMs (CT-CLIP[31], fVLM[32], MedVista3D[14], T3D[33], Merlin[10], RadZero3D[45], and BIUD[46]); (2) medical image segmentation models (BiomedParse-v2[47], SegMamba[48], and RWKV-UNet[49]); (3) radiology report generation models (CT-CHAT[31], RadFM[50], CT2Rep[51], M3D[52], Reg2RG[53], CT-GRAPH[54], BTB3D[55], and Hulu-Med[56]); and (4) the supervised baseline CT-Net[39]. Comprehensive evaluations demonstrate the effectiveness of EXACT, which consistently delivers the best overall performance across all evaluated tasks and datasets. In addition, EXACT provides intrinsic visual grounding by using its anomaly-aware maps (AAmaps) to spatially localize the pathological findings described in the generated reports (Fig. 10), thereby offering voxel-level visual evidence that improves clinical interpretability and strengthens diagnostic confidence. Further details of the EXACT architecture, pre-training procedure, and dataset curation are provided in Methods.

**EXACT learned robust anatomical priors for anomaly-aware pre-training**

Accurate anatomical segmentation provides the spatial scaffold for anatomy-constrained anomaly detection in EXACT. Across six thoracic structures, EXACT achieved robust segmentation performance using automatically derived anatomical masks as supervision (Supplementary Table 2, Supplementary Fig. 7). The highest Dice similarity coefficients (DSC) were observed for the lung and pleura, both reaching 0.966 ± 0.088. The heart and trachea/bronchi were also segmented accurately, with DSCs of 0.889 ± 0.114 and 0.886 ± 0.010, respectively. Performance was lower for the mediastinum and esophagus, with DSCs of 0.811 ± 0.093 and 0.821 ± 0.115, likely reflecting less distinct tissue boundaries and greater anatomical variability on non-contrast CT.

**EXACT achieved notable improvement for multi-disease diagnosis**

We first evaluated EXACT for multi-disease diagnosis from 3D chest CT. This task requires whole-volume reasoning over multiple coexisting abnormalities and is further complicated by substantial variation in disease prevalence across clinical cohorts. We therefore assessed EXACT across 18 target abnormalities with pronounced cross-site distribution shifts, including lung nodule prevalence ranging from 45.0% in CT-RATE to 90.3% in MianYang, as well as dataset-specific label availability (Supplementary Table 1, Supplementary Fig. 8). EXACT was evaluated in two settings: a zero-shot setting based on anatomy-guided top-k aggregation of AAmaps and organ masks (Fig. 2a), and a fine-tuning setting in which a dedicated AAmap encoder and classifier were trained on EXACT-generated AAmaps for diagnosis (Fig. 2b).

EXACT in the zero-shot setting consistently outperformed all compared models across both internal and external validation (Fig. 3, Supplementary Table 4). On the CT-RATE internal validation dataset, EXACT (Zero-shot) achieved an area under the receiver operating characteristic curve (AUROC) of 0.830 (95% confidence interval (95% CI), 0.822–0.837), an F1 score of 0.836 (95% CI, 0.826–0.844), and an accuracy of 0.768 (95% CI, 0.757–0.778). Notably, this performance surpassed not only all zero-shot baselines, such as fVLM (Zero-shot) (AUROC, 0.777) and CT-CLIP (Zero-shot) (AUROC, 0.731), but also models reported with supervised fine-tuning on CT-RATE, including the publication-reported result for T3D (Fine-tuning) (AUROC, 0.802). This zero-shot superiority was consistently maintained on out-of-distribution (OoD) external datasets. On the RAD-ChestCT dataset, EXACT (Zero-shot) achieved an AUROC of 0.728 (95% CI, 0.722-0.734), outperforming the publication-reported result for the strongest baseline, MedVista3D (Zero-shot) (AUROC, 0.710). On the MianYang dataset, EXACT (Zero-shot) achieved an AUROC of 0.758 (95% CI, 0.739-0.778), exceeding fVLM (Zero-shot) (AUROC, 0.716) and all CT-CLIP variants (best AUROC, 0.712).

Lightweight supervised adaptation unlocked further diagnostic gains, with EXACT (Fine-tuning) achieving the best overall performance across all datasets (Fig. 3, Supplementary Table 4). On the CT-RATE internal validation dataset, fine-tuning yielded an AUROC of 0.833 (95% CI, 0.826-0.841), representing a statistically significant improvement over EXACT (Zero-shot) ($P = 0.001$). On the external validation datasets, the gains from fine-tuning were even more pronounced: AUROC improved to 0.734 (95% CI, 0.728-0.740) on RAD-ChestCT and to 0.769 (95% CI, 0.748-0.787) on MianYang. On RAD-ChestCT, the improvements were significant for AUROC ($P < 0.001$) and accuracy ($P = 0.025$), but not for F1 score ($P = 0.105$). On MianYang, the improvements were significant for all three metrics (AUROC, $P = 0.006$; F1 score and accuracy, both $P < 0.001$). Disease-specific AUROC and prevalence plots (Fig. 2c–e for zero-shot; Fig. 2f–h for fine-tuning) and AUROC heatmaps (Supplementary Fig. 6) further confirmed that EXACT achieved broadly superior discrimination across the majority of individual pathological conditions and datasets, rather than being driven by a subset of easily classifiable diseases.

The relatively poor cross-site generalization of CLIP-based baselines, including CT-CLIP, fVLM, MedVista3D, T3D, Merlin, RadZero3D, and BIUD, exposed the inadequacy of relying predominantly on global or local contrastive alignment for precise 3D chest CT understanding. Collapsing high-dimensional volumetric features into compact vectors inevitably discards essential 3D spatial context, thereby inherently compromising the detection of spatially sparse abnormalities. Specifically, on the external RAD-ChestCT and MianYang dataset, all comparison methods with or without fine-tuning remained below an AUROC of 0.720 (Supplementary Table 4). In contrast, EXACT inherently preserves voxel-level spatial structures and, through holistic modeling of organ-specific contextual relationships across the entire scan, demonstrates superior capability in identifying complex comorbidities and mitigating critical clinical errors such as inattentional blindness (Supplementary Fig. 6). For example, EXACT (Zero-shot) achieved pronounced advantages on organ localized pathologies such as pericardial effusion (PCE) and hiatal hernia (HH), with AUROC gains over CT-CLIP (Zero-shot) exceeding 0.12 on CT-RATE (0.90 versus 0.77 for PCE and 0.84 versus 0.70 for HH).

**EXACT enabled voxel-level unsupervised anomaly localization**

We next evaluated whether EXACT could achieve precise unsupervised anomaly localization from 3D chest CT. Accurately delineating the spatial extent of pathological findings is essential for countering the under-reading of visible abnormalities, a prevalent source of diagnostic error in radiology[13]. By coupling explicit anatomical constraints with a multi-scale multi-instance learning (MIL) objective within a unified Y-Mamba architecture, EXACT is, to our knowledge, the first 3D chest CT foundation model to achieve intrinsic voxel-level anomaly localization entirely through automated text-derived weak supervision, without any exposure to manual voxel-level masks. Specifically, during zero-shot inference (Fig. 4a), the pre-trained EXACT model simultaneously generates 18 disease-specific AAmaps alongside anatomical organ masks via its parallel decoders. To eliminate background noise, an explicit anatomical constraint is applied by element-wise multiplying each AAmap with its corresponding organ mask. Task-relevant disease channels are then selected and aggregated according to the dataset-specific evaluation protocol to generate a comprehensive anomaly map, which is subsequently binarized to yield the final localization prediction (Methods).

Qualitative comparison across three representative cases (Fig. 4b) revealed that EXACT was the only model to produce heatmaps that accurately delineated lesion boundaries, successfully capturing mixed consolidation and ground-glass opacity (GGO) patterns in the first two cases from ReX-Val and COVID-19, as well as isolated GGOs in the third case from MosMed. Extended evaluations confirm EXACT's consistent performance across diverse morphologies (Supplementary Fig. 10). In contrast, CT-CLIP and fVLM rely on post-hoc gradient-weighted class activation mapping (Grad-CAM) and are thus fundamentally incapable of achieving true voxel-level localization, yielding diffuse and structurally ambiguous attention maps that are heavily compromised by false-positive noise in the background and normal anatomical regions. Furthermore, although BiomedParse-v2 is pre-trained in a supervised manner for zero-shot biomedical image segmentation via task-specific text prompts (e.g., "lesions of lungs and pleura"), it exhibited insufficient generalizability to these complex pathological patterns, resulting in false-negative omissions within the lesion areas.

To quantitatively evaluate zero-shot localization performance, the DSC and the area under the precision-recall curve (AUPR) were employed as primary spatial overlap metrics. However, absolute DSC and AUPR values were inherently modest given the profound difficulty of fine-grained localization in 3D chest CT without dense supervision. Hit Rates at 5% and 10% thresholds (Hit Rate@5%/10%), defined as the proportion of samples exceeding the corresponding metric threshold, were therefore additionally reported to assess whether models could reliably detect meaningful portions of clinically relevant lesions.

EXACT achieved SOTA zero-shot anomaly localization performance across all datasets (Fig. 5a, Supplementary Table 5). Specifically, on the internal validation dataset, EXACT achieved DSCs of 0.096 (95% CI, 0.089–0.103) and 0.125 (95% CI, 0.105–0.147) on ReX-Train and ReX-Val, respectively, significantly outperforming the strongest baseline BiomedParse-v2 (DSCs of 0.032 and 0.071; $P < 0.001$ and $P < 0.001$), with correspondingly higher DSC Hit Rates (e.g., Hit Rate@5% of 0.472 vs. 0.145 on ReX-Train, 0.605 vs. 0.312 on ReX-Val). Parallel improvements were observed in AUPR, where EXACT reached 0.093 and 0.125 on two datasets, exceeding BiomedParse-v2 (0.026 and 0.031; both $P < 0.001$), alongside consistently higher AUPR Hit Rates (e.g., Hit Rate@5% of 0.432 vs. 0.125 on ReX-Train, 0.567 vs. 0.185 on ReX-Val).

EXACT extended its robust generalization to OoD external datasets (Fig. 5a, Supplementary Table 5). On MosMed and COVID-19 datasets, EXACT attained high mean DSCs of 0.362 and 0.435, surpassing BiomedParse-v2 (0.254 and 0.340), coupled with remarkable DSC Hit Rates (e.g., Hit Rate@5% of 0.960 vs. 0.840 on MosMed, 0.950 vs. 0.550 on COVID-19). Additionally, EXACT reached AUPR of 0.330 and 0.530 on these two datasets, outperforming BiomedParse-v2 (0.258 and 0.459) alongside higher Hit Rates (e.g., Hit Rate@5% of 0.960 vs. 0.820 on MosMed). While these metrics represented highly significant improvements on MosMed ($P < 0.001$ for DSC; $P = 0.002$ for AUPR), the clear numerical advantages on COVID-19 did not reach statistical significance ($P > 0.05$), likely constrained by insufficient statistical power given the limited sample size ($n = 20$). In contrast, across all four datasets, the Grad-CAM-reliant CT-CLIP and fVLM yielded extremely low spatial overlap (Fig. 5a, Supplementary Table 5). For example, CT-CLIP produced near-zero DSCs of 0.004, 0.005, 0.022, and 0.004 on ReX-Train, ReX-Val, COVID-19, and MosMed, respectively, thereby underscoring the fundamental inadequacy of post-hoc visualizations for precise localization in volumetric imaging.

**EXACT improved the performance of supervised anomaly localization**

Beyond zero-shot inference, we evaluated whether the rich anatomical and pathological representations extracted by EXACT could serve as a powerful initialization for downstream supervised segmentation tasks. Specifically, we retained the pre-trained image encoder and anomaly detection decoder, introducing a final 1×1×1 convolutional layer to compress the multi-channel AAmaps into a single-channel probability map (Methods). To address the challenge of localizing small lesions, we fine-tuned the entire model end-to-end using a hybrid Tversky and Focal loss, yielding EXACT-Seg. For internal validation, EXACT-Seg and comparative baselines (RWKV-UNet and SegMamba) were trained on ReX-Train and evaluated on ReX-Val. For COVID-19 and MosMed, EXACT-Seg and both supervised baselines were fine-tuned on 4 and 10 scans, respectively, and evaluated on the remaining 16 and 40 scans.

EXACT achieved statistically significant improvements over both baselines (Fig. 5b, Supplementary Table 5). On the ReX-Val internal validation dataset, EXACT-Seg attained a DSC of 0.215 (95% CI, 0.182–0.249), outperforming the next-best model SegMamba (DSC, 0.198; *P* = 0.028), alongside a higher AUPR of 0.200 (95% CI, 0.165–0.238) versus 0.187 for SegMamba (*P* = 0.007). Correspondingly, EXACT-Seg achieved the highest DSC Hit Rates (Hit Rate@5% of 0.643 vs. 0.573 for SegMamba; Hit Rate@10% of 0.580 vs. 0.503), indicating that a larger proportion of cases exceeded clinically meaningful overlap thresholds.

The advantages of EXACT-Seg's pre-trained representations were even more pronounced on the OoD external datasets, where limited training data amplified the value of transferable feature priors (Fig. 5b, Supplementary Table 5). On the COVID-19 dataset, EXACT-Seg achieved a DSC of 0.476 (95% CI, 0.332-0.621), representing a 43.4% improvement over SegMamba (DSC, 0.332; *P* < 0.001), while its AUPR reached 0.529 (95% CI, 0.374–0.679), substantially exceeding both SegMamba (0.358) and RWKV-UNet (0.404; *P* = 0.006). On the MosMed dataset, EXACT-Seg maintained consistent superiority, achieving a DSC of 0.454 (95% CI, 0.387–0.520) compared with 0.352 for SegMamba and 0.348 for RWKV-UNet (*P* < 0.001), alongside an AUPR of 0.463 (95% CI, 0.393–0.536) that significantly surpassed both baselines (*P* < 0.001). Hit Rates on both external datasets further corroborated these gains (e.g., DSC Hit Rate@10% of 0.813 on COVID-19 and 0.900 on MosMed for EXACT-Seg, versus 0.750 and 0.850 for SegMamba, respectively).

Qualitative visualizations provided further insight into the mechanisms underlying these quantitative improvements (Fig. 4d). Constrained by 2D slice-wise inputs, RWKV-UNet lacked inter-slice contextual continuity, frequently producing pronounced false-negative omissions. SegMamba, despite its 3D architecture, suffered from the absence of domain-specific priors due to random initialization, rendering it highly susceptible to false-positive segmentations. In contrast, EXACT-Seg leveraged the robust 3D global context and organ-aware anomaly sensitivity internalized during large-scale weakly supervised pre-training to effectively suppress false positives and mitigate lesion omissions, yielding highly concordant segmentation masks even for morphologically heterogeneous pathologies (Supplementary Figs. 11-13).

**EXACT-CHAT demonstrated the best clinical efficacy as a multimodal AI assistant**

We also evaluated whether EXACT's anomaly-aware representations could be extended to automated radiology report generation to address the miscommunication of correctly identified pathology, a persistent and clinically consequential source of diagnostic error[13,14]. We developed EXACT-CHAT (Methods), a multimodal AI assistant adapted from the LLaVA framework[57] that integrates the frozen EXACT image encoder, a multimodal projector module and LLaMA-3.1-8B-Instruct[58] for report generation (Fig. 7a). Additionally, EXACT-CHAT provides the large language model (LLM) with structured diagnostic priors from the frozen AAmap encoder and classifier, formatted as text tokens that encode the predicted disease states of all target abnormalities. An optional GPT-4.1 refinement step further calibrates the initial reports by correcting missed findings and suppressing hallucinated diagnoses based on upstream disease predictions (Supplementary Fig. 4), yielding EXACT-CHAT (Refined).

We evaluated report generation performance on CT-RATE (internal validation), RAD-ChestCT and MianYang (external validation) datasets separately before and after disease-conditioned post-hoc refinement (Methods). Standard natural language generation (NLG) metrics (i.e., BLEU-1, METEOR, CIDEr, ROUGE-L) were used for lexical quality assessment on CT-RATE and MianYang, for which reference reports were available. However, since NLG metrics primarily measure surface-level lexical overlap and are limited in capturing clinical fidelity[31,59], we additionally adopted clinical efficacy metrics as the primary evaluation measures. Specifically, a pre-trained RadBERT[38] model was employed to extract structured pathological labels from both generated and reference reports on CT-RATE and MianYang, from which RadBERT-F1, RadBERT-Precision, and RadBERT-Recall were computed to directly quantify the diagnostic accuracy beyond text matching. On RAD-ChestCT, these metrics were computed by comparing labels extracted from generated reports with the dataset-provided abnormality labels.

Before post-hoc refinement, EXACT-CHAT achieved the highest RadBERT-F1 among the independently evaluated models on all three datasets (Fig. 8, Supplementary Table 6). Specifically, on CT-RATE, EXACT-CHAT attained a RadBERT-F1 of 0.310 (95% CI, 0.274–0.347), significantly outperforming CT-GRAPH, the strongest independently evaluated baseline for this metric (0.255; $P < 0.001$). On RAD-ChestCT, EXACT-CHAT achieved a RadBERT-F1 of 0.289 (95% CI, 0.265–0.313), compared with 0.279 for Hulu-Med, although this difference did not reach statistical significance ($P = 0.056$). On the MianYang dataset, which represents a demographically and institutionally distinct Chinese clinical cohort with substantially different disease distributions (Supplementary Fig. 8), EXACT-CHAT achieved a RadBERT-F1 of 0.290 (95% CI, 0.221–0.359), outperforming CT-GRAPH (0.200; $P < 0.001$). Notably, on CT-RATE, CT-CHAT achieved higher CIDEr and ROUGE-L scores than EXACT-CHAT (0.235 vs 0.141 and 0.353 vs 0.328, respectively), despite a lower RadBERT-F1 (0.165 vs 0.310), underscoring the importance of clinically grounded evaluation beyond lexical overlap.

We next compared report generation after a common GPT-4.1 post-hoc refinement procedure, conditioned on disease predictions produced by each model (Fig. 9, Supplementary Table 7). Under this setting, EXACT-CHAT (Refined) achieved the highest RadBERT-F1 among the models evaluated in each cohort. On CT-RATE, EXACT-CHAT (Refined) attained a RadBERT-F1 of 0.501 (95% CI, 0.456–0.543), exceeding CT-CHAT (0.427; $P < 0.001$), alongside a RadBERT-Precision of 0.414 (95% CI, 0.368–0.460) and a RadBERT-Recall of 0.730 (95% CI, 0.677–0.780). On RAD-ChestCT and MianYang, EXACT-CHAT (Refined) achieved RadBERT-F1 scores of 0.441 (95% CI, 0.416–0.467) and 0.410 (95% CI, 0.319–0.497), respectively, outperforming Hulu-Med on both datasets (0.419 and 0.327, respectively; both $P < 0.001$). EXACT-CHAT (Refined) also achieved the highest BLEU-1 among the models evaluated after refinement on CT-RATE (0.522; 95% CI, 0.516–0.529) and MianYang (0.503; 95% CI, 0.495–0.512).

Representative case studies further confirmed that EXACT-CHAT (Refined) generated clinically accurate reports with high concordance to radiologist-authored references (Fig. 7b, Supplementary Figs. 14, 15). In one representative case (CT-RATE dataset) involving a complex presentation of viral pneumonia (COVID-19) with multilobar consolidation, GGOs, interlobular septal thickening, and a sliding-type hiatal hernia, EXACT-CHAT (Refined) successfully identified all findings without omissions, whereas the standard EXACT-CHAT missed consolidation and hiatal hernia, and EXACT-CHAT (w/o Prior), which relies solely on visual features without AAmap-derived diagnostic priors, failed to detect all five key pathological findings while hallucinating emphysematous changes and nonspecific nodules (Fig. 7b). Extended case studies (CT-RATE dataset) further demonstrated that EXACT-CHAT (Refined) could recover subtle findings (e.g., pulmonary fibrotic sequelae) that were missed by EXACT-CHAT and EXACT-CHAT (w/o Prior), and correctly characterized consolidation alongside GGOs in pneumonic infiltration cases (Supplementary Fig. 14). Compared with unrefined reports from baseline models on a complex case (MianYang dataset) involving bilateral lower lobe infection, pleural effusion, cardiomegaly, pericardial effusion, and lymphadenopathy, EXACT-CHAT and EXACT-CHAT (Refined) detected the majority of these comorbidities, whereas CT-CHAT reported the entire examination as “within normal limits,” Merlin detected only mild ATE, Hulu-Med hallucinated phrenic nerve palsy, and Reg2RG fabricated a breast cystic lesion (Supplementary Fig. 15).

Beyond report generation accuracy, EXACT-CHAT provided visual grounding for the pathological findings described in its generated reports. For hiatal hernia, slice-level anomaly scores peaked near the gastroesophageal junction, with AAmaps highlighting the corresponding herniated region (Fig. 10a). For ground-glass opacities and consolidation, AAmaps delineated peripheral and subpleural abnormalities consistent with the generated descriptions (Fig. 10b,c).

For pulmonary nodules, AAmaps localized focal lesions within the relevant slices, supporting the grounding of small and spatially sparse findings (Fig. 10d). These results indicate that EXACT-CHAT can link generated report content to anatomically plausible voxel-level evidence, providing a transparent basis for verifying report reliability.

**EXACT supported frozen-encoder transfer to tumour segmentation and survival prediction**

To assess transfer beyond the pre-training objectives, we evaluated the frozen EXACT encoder on lung tumour segmentation in the NSCLC-Radiomics (LUNG1) cohort[44], using 18 training patients, four validation patients and 400 held-out test patients (Fig. 6a,b; Supplementary Table 8). EXACT-Seg achieved a DSC of 0.291 (95% CI, 0.268–0.315) and an AUPR of 0.325 (95% CI, 0.299–0.352), significantly outperforming the strongest baseline, RWKV-UNet, which achieved a DSC of 0.254 and an AUPR of 0.217 ($P$ = 0.001 and $P$ < 0.001, respectively). Representative predictions and aggregate performance are shown in Fig. 6a,b. These results support the transferability of the pre-trained representations to lung tumour segmentation with limited task-specific annotations.

We further evaluated whether frozen EXACT representations could be used for survival prediction in the NSCLC-Radiomics (LUNG1) cohort. Overall survival analysis included all 422 patients, whereas two-year survival prediction included 420 patients after excluding two patients censored before two years. Specifically, we compared three experimental settings (Methods): (1) features extracted from FMCIB[60], a self-supervised foundation model developed for cancer imaging biomarker discovery and previously evaluated for NSCLC prognostication; (2) FMCIB features combined with EXACT global representations; and (3) FMCIB features combined with EXACT tumour-pooled representations. Both encoders remained frozen, and only regularized linear prediction models were trained. Each endpoint was evaluated using five repeats of five-fold cross-validation, with identical patient splits across the three settings (Fig. 6c; Supplementary Table 9).

Fusion with tumour-pooled EXACT representations improved survival prediction over FMCIB alone (Fig. 6c; Supplementary Table 9). Specifically, this configuration achieved a mean C-index of 0.568, outperforming FMCIB alone (0.551; $P$ < 0.05), alongside a higher two-year AUROC of 0.646 versus 0.622 ($P < 0.001$). Fusion with EXACT global representations also improved the two-year AUROC to 0.638 ($P < 0.001$ versus FMCIB alone), although the increase in C-index to 0.558 did not reach statistical significance. These findings suggest that frozen EXACT representations can provide complementary information for survival prediction.

**Anatomical confinement improved classification and anomaly localization**

To examine the contribution of anatomical confinement, we compared EXACT with two ablated variants (Supplementary Table 10): (1) inference without anatomical confinement, using the same trained weights; and (2) training and inference without anatomical confinement, with the model retrained accordingly. EXACT achieved a higher classification AUROC on CT-RATE than the variant retrained without confinement (0.830 versus 0.811; $P < 0.001$). On COVID-19, EXACT also achieved higher localization DSC (0.435 versus 0.030) and AUPR (0.530 versus 0.250; both $P < 0.001$). Removing anatomical masks only at inference reduced classification AUROC to 0.671, localization DSC to 0.012 and AUPR to 0.042 (all $P < 0.001$ versus EXACT). In summary, these results support the contribution of anatomical confinement to both disease discrimination and spatial localization (Supplementary Table 10).

# Discussion

In this study, we present EXACT, a 3D chest CT FM that enables voxel-level anomaly awareness through weakly supervised learning from paired scans and radiology reports, without requiring any manual annotation. Across five clinically motivated tasks evaluated on multinational, multi-center cohorts with diverse disease distributions, EXACT consistently outperformed SOTA 3D medical FMs. Notably, EXACT achieved precise zero-shot anomaly localization that even rivaled fully supervised models on external datasets (e.g., zero-shot DSC of 0.435 versus supervised 0.476 on the COVID-19 dataset), and generated clinically accurate, visually grounded radiology reports with robust generalization to OoD datasets. By bridging global semantic understanding with fine-grained visual grounding through anatomy-aware pretraining, EXACT establishes a scalable and explainable foundation for general-purpose chest CT interpretation.

CLIP has become the predominant self-supervised paradigm for medical vision FMs[61], yet its application to volumetric chest CT reveals fundamental representational limitations. Specifically, CLIP-based approaches, from CT-CLIP's global contrastive alignment through fVLM and MedVista3D's organ-guided local matching to T3D and SimCroP's implicit multi-view or patch-level strategies, represent a progressive refinement toward finer alignment granularity. However, they ultimately compress high-dimensional volumetric features into compact 1D embedding vectors via global average pooling or [CLS] token aggregation, irreversibly collapsing the rich voxel-wise spatial structure of 3D CT volumes. In practice, this information loss manifests in three principal ways. First, multi-disease diagnostic accuracy degrades substantially on OoD cohorts, with all CLIP-based methods remaining below an AUROC of 0.720 on external datasets. More importantly, performance on organ-localized and spatially sparse pathologies is particularly compromised under domain shift. For instance, CT-CLIP's zero-shot AUROC fell to 0.60 for hiatal hernia on the external RAD-ChestCT dataset (Supplementary Fig. 6). Second, spatial interpretability is entirely relegated to post-hoc gradient visualizations (e.g., Grad-CAM), which produce diffuse, structurally ambiguous heatmaps fundamentally incapable of precise localization, as evidenced by CT-CLIP's near-zero DSCs (0.004 – 0.022) across all anomaly localization datasets (Supplementary Table 5). Third, multimodal LLMs built upon CLIP-based image encoders (e.g., Merlin, CT-CHAT, and M3D) inherit the same spatial insensitivity from their underlying vision backbones and lack anatomical grounding, resulting in precipitous clinical efficacy declines under domain shift (e.g., RadBERT-F1 scores before post-hoc refinement of 0.073 for CT-CHAT (LLaMA 3.1 70B) and 0.024 for Merlin on MianYang; Supplementary Table 6).

Beyond CLIP, recent works have explored alternative pretraining paradigms for chest CT, yet each introduces its own trade-offs. Multimodal transformers such as MedMPT[62] enrich patient-level context by incorporating auxiliary clinical data (e.g., laboratory tests), but their spatial interpretability still relies on attention rollout techniques[63,64] that only roughly approximate the model's attention distribution, and their dependence on non-imaging inputs limits applicability to CT-only workflows. Vision-only approaches based on diffusion (e.g., LCTfound[65]) or masked autoencoders (e.g., TANGERINE[66]) learn robust visual features yet lack cross-modal semantic alignment, rendering them unable to support zero-shot diagnosis or report generation without extensive supervised adaptation. Furthermore, the recently deployed clinical vision FM, OMAFound[67], has shown that despite employing various post-hoc interpretability methods to enhance transparent decision-making, qualitative heatmap visualizations generally exhibit biases compared with expert radiologists, regardless of classification accuracy.

EXACT establishes anatomy-aware weakly supervised learning as a viable and more clinically grounded alternative to the CLIP paradigm for volumetric medical imaging. Whereas CLIP was originally conceived for natural image-text matching and subsequently transplanted to the medical domain, EXACT is guided by clinical priors to align with the hierarchical workflow of radiologic interpretation. It first delineates anatomical structures to establish spatial context before detecting and localizing pathological deviations within those structures. EXACT achieves this capability through two tightly coupled architectural innovations. First, an anatomy-constrained MIL framework treats each organ region as an instance bag, confining disease evidence exclusively to anatomically plausible tissue and enabling the extraction of discriminative pathological features specific to each organ from image-level supervision. As a result, EXACT demonstrates exceptional sensitivity to highly localized pathologies where alignment-based methods fail under domain shift. For example, while CT-CLIP degraded to an AUROC of 0.60 for hiatal hernia on RAD-ChestCT, EXACT maintained robust performance at 0.78. Similarly, for pericardial effusion on the MianYang dataset, EXACT achieved an AUROC of 0.84 compared with CT-CLIP's 0.67 (Supplementary Fig. 6). The ablation analysis further supported this design, with anatomical confinement improving classification AUROC on CT-RATE and localization DSC and AUPR on COVID-19 compared with retraining without confinement (Supplementary Table 10). Second, the Y-Mamba dual-decoder architecture inherently preserves fine-grained spatial detail throughout the decoding process. By fusing multi-scale representations, Y-Mamba generates dense, spatially explicit AAmaps that accurately capture both focal lesion cores and diffuse peripheral boundaries without relying on post-hoc gradient visualizations or lossy vector compression. Consequently, EXACT enables zero-shot voxel-level anomaly localization, achieving DSCs of 0.435 and 0.362 with Hit Rates@5% exceeding 95% on the external COVID-19 and MosMed datasets, while Grad-CAM-based approximations from existing foundation models yielded near-zero spatial overlap (Supplementary Table 5). Remarkably, these zero-shot results already surpassed those of the fully supervised SegMamba on the same external data (DSCs of 0.332 and 0.352) and closely approached those of the fine-tuned EXACT-Seg (0.476 and 0.454). Even on the more challenging ReX-Val dataset, the few-shot adapted EXACT-Seg (DSC of 0.215) still significantly outperformed SegMamba (0.198; $P = 0.028$).

Furthermore, extending 3D vision FMs to automated radiology report generation exposes a critical dissociation between lexical fluency and clinical fidelity. Recent systematic evaluations of CT report generation metrics have demonstrated that conventional NLG scores are brittle to stylistic variations and fail to penalize factual errors[68]. The narrative complexity of volumetric CT reports, where clinically irrelevant normal findings are typically omitted, further exacerbates class imbalance and renders lexicon-based overlap metrics (e.g., BLEU, ROUGE) misleading[69].

Our comprehensive evaluation (Supplementary Tables 6 and 7) further substantiates these concerns. Specifically, on the internal CT-RATE dataset, baseline models such as T3D and Reg2RG achieved relatively high conventional NLG scores (e.g., ROUGE-L of 0.378 and METEOR of 0.210, respectively). However, their clinical efficacy was remarkably poor, yielding RadBERT-F1 scores of only 0.274 and 0.172, respectively. Similarly, on the external MianYang dataset, despite maintaining moderate lexical overlap (BLEU-1 of 0.299), the 70-billion-parameter CT-CHAT model exhibited a complete collapse in diagnostic accuracy, plummeting to a RadBERT-Recall of 0.088, indicating pervasive hallucinations and the systemic omission of critical findings. In contrast, by explicitly anchoring language generation to voxel-level diagnostic priors derived from AAmaps and coupling it with GPT-4.1-driven post-hoc calibration, EXACT-CHAT (Refined) achieved the highest RadBERT-F1 among the models evaluated after refinement across all cohorts (RadBERT-F1 of 0.501, 0.441, and 0.410 on CT-RATE, RAD-ChestCT, and MianYang, respectively), ensuring that the generated narratives are driven by verifiable spatial evidence rather than statistical text generation.

From a clinical application perspective, EXACT demonstrates immense value across diverse healthcare scenarios, particularly in settings constrained by limited annotated data and radiologist shortages. In the development of medical AI, acquiring expert-curated, voxel-level annotations is notoriously cost-prohibitive and time-consuming. By encapsulating extensive anatomical and pathological representations into a unified FM through annotation-free pre-training, EXACT bypasses this bottleneck. Its robust zero-shot and few-shot capabilities enable the immediate deployment of highly accurate multi-disease diagnosis and precise anomaly localization across multiple medical centers, proving especially beneficial for identifying rare or spatially sparse pathologies where training data is inherently scarce. Furthermore, the inherent generation of voxel-level AAmaps addresses the critical trustworthiness challenge in medical AI. Rather than relying on opaque black-box decisions or diffuse post-hoc visualizations, EXACT provides clinicians with transparent, visually grounded evidence, thereby facilitating confident decision-making and effectively mitigating inattentional blindness. Finally, integration into routine clinical workflows via EXACT-CHAT establishes a reliable multimodal AI assistant capable of automating spatially-grounded radiology report generation. This not only alleviates the cognitive burden on radiologists but also holds profound potential for advancing scalable, large-scale opportunistic disease screening. In essence, EXACT positions itself as an intelligent, explainable brain for 3D chest CT imaging, ready to seamlessly tackle complex real-world clinical tasks.

There are several limitations to our study. First, EXACT relies on a fine-tuned RadBERT model to extract structured disease pseudo-labels from free-text radiology reports during pre-training. Report omissions, ambiguous descriptions and label-extraction errors, together with inaccuracies in SAT-generated anatomical masks, may introduce noise into the supervision and propagate through pre-training. Although RadBERT achieved high extraction accuracy, recent studies have demonstrated that general-purpose LLMs can match or exceed domain-specific models for clinical information extraction without requiring specialized pre-training or fine-tuning[70–73]. Future iterations of EXACT could therefore replace RadBERT with more capable general-purpose LLMs to further improve label extraction accuracy and reduce pipeline complexity. Second, the current label extraction framework requires manual pre-definition of the 18 target abnormalities and their corresponding disease-to-organ mappings, which constrains the scalability of EXACT to rare pathologies. Inaccurate anatomical masks or abnormalities extending beyond the assigned anatomical region may attenuate relevant anomaly responses. The global foreground channel is currently assigned only to medical material and arterial wall calcification and does not remove this constraint for other disease channels. Task-specific adaptation offers a route to extending EXACT beyond its pre-training objectives with limited annotations. For example, we adapted EXACT to lung tumour segmentation using only 18 training patients and four validation patients, with the encoder frozen and only the segmentation decoder and output head optimized (Fig. 6a,b; Supplementary Table 8). Future research should explore automated methods for inducting structured disease taxonomies and reporting templates directly from large-scale radiology report corpora, enabling a fully automated pre-training pipeline. More flexible anatomical constraints and task-specific adaptation should also be explored to accommodate atypical disease distributions.

Third, although EXACT demonstrated robust generalization across multinational and multi-center retrospective cohorts, all evaluations in this study were conducted in retrospective settings. Prospective clinical validation, including assessment of its impact on radiologist diagnostic accuracy, interpretation efficiency, and patient outcomes, is needed to establish the translational utility of EXACT in real-world clinical practice. Fourth, EXACT remains limited in accurately delineating small lesions. Specifically, we adopted a fixed input grid of 64 × 128 × 128 voxels after resampling to 1 × 1 × 3 mm to preserve whole-volume anatomical context across multiple downstream tasks, consistent with our aim of developing a general-purpose volumetric model. As discussed in Merlin[10], the benefits of higher spatial resolution in CT may depend on the task, and lower-resolution scans can offer favourable signal-to-noise characteristics for clinical interpretation. Nevertheless, delineating fine lesion boundaries requires sufficient spatial detail. Our lesion-size-stratified analyses (Supplementary Figs. 16 and 17) showed low median matched-lesion Dice in the smallest size bins under zero-shot evaluation on ReX-Train and ReX-Val, with similarly limited overlap among supervised models in the same size bins on ReX-Val. These findings highlight the difficulty of delineating small abnormalities and are consistent with a potential loss of fine spatial detail during resizing. Future work could evaluate higher-resolution inputs and local refinement strategies to improve small-lesion delineation while preserving whole-volume anatomical context.

# Methods

## Data collection and preprocessing

To pre-train EXACT (Fig. 1a, Fig. 1b), we used the CT-RATE dataset[31], which comprises 25,692 non-contrast 3D chest CT volumes from 21,304 patients paired with free-text radiology reports spanning a broad spectrum of thoracic pathologies and anatomical variants. Structured disease pseudo-labels for 18 target abnormalities were extracted from the reports using a fine-tuned RadBERT model[38], for which a mean report-label extraction F1 score of 0.976 ± 0.016 was reported in the original CT-RATE study[31]. This report-level extraction performance does not exclude errors or omissions in the source reports or guarantee complete correspondence between report-derived labels and all image-visible findings. For anatomical supervision of EXACT, we used RadGenome-Chest CT[74], a derivative of CT-RATE that provides coarse organ segmentation masks (i.e., lung, trachea and bronchi, pleura, mediastinum, heart and esophagus) generated by the pre-trained SAT model[37] for the same set of volumes. Following the original dataset partition, 24,128 scans from 20,000 patients were used for EXACT pre-training, and the remaining 1,564 scans from 1,304 patients were held out for internal validation in downstream tasks.

For multi-disease classification, one internal (CT-RATE) and two external (RAD-ChestCT and MianYang) validation datasets with substantial cross-site distribution shifts were used (Supplementary Table 1, Supplementary Fig. 8). Specifically, internal validation used the 1,564 held-out scans in CT-RATE. For external validation, the RAD-ChestCT[39] dataset and the MianYang dataset were used. RAD-ChestCT comprises 3,630 non-contrast chest CT volumes from the Duke University Medical System, with abnormality labels provided by the dataset. We retained the 16 diseases overlapping with CT-RATE, excluding coronary artery wall calcification and mosaic attenuation pattern, for which labels were unavailable. The MianYang dataset, comprising 505 non-contrast chest CT scans with paired Chinese radiology reports, was retrospectively collected from Mianyang Central Hospital, Sichuan, China. Structured disease labels for 18 target abnormalities were extracted from the Chinese reports using DeepSeek-V3[75] (prompts in Supplementary Fig. 2). Mosaic attenuation pattern was subsequently excluded owing to a positive rate of zero in this cohort, yielding 17 evaluable disease categories.

For anomaly localization, two internal datasets (ReX-Train and ReX-Val) from ReXGroundingCT[40] and two external datasets (MosMed[43] and COVID-19[41,42]) were used. Specifically, ReXGroundingCT is derived from CT-RATE and provides expert-annotated, voxel-level 3D segmentation masks of diverse lung and pleural abnormalities. Within ReXGroundingCT, ReX-Train (*n* = 1,102) consists of scans that overlap with the EXACT pre-training corpus, while ReX-Val (n = 157) is derived from the held-out CT-RATE validation partition. For external evaluation, we used the COVID-19 dataset (*n* = 20) sourced from the Coronacases Initiative and Radiopaedia, featuring lesion regions manually delineated by two radiologists and verified by a senior radiologist. Additionally, we included the MosMed dataset (*n* = 50) from municipal hospitals in Moscow, Russia, which provides expert-annotated masks for ground-glass opacities and consolidation. Under the zero-shot setting, the pre-trained EXACT model was directly evaluated across all four datasets (ReX-Train, ReX-Val, COVID-19, and MosMed). For the supervised fine-tuning setting, EXACT-Seg was trained on ReX-Train and internally validated on ReX-Val. For the external datasets, 20% of the scans, corresponding to 4 COVID-19 cases and 10 MosMed cases, were used for fine-tuning, and the remaining 16 and 40 scans, respectively, were reserved for evaluation.

For the radiology report generation task, EXACT-CHAT was trained using paired 3D chest CT volumes and English free-text radiology reports from the training partition of the CT-RATE dataset (*n* = 24,128) and internally validated on the 1,564 held-out scans from the same dataset. For external validation, we employed the RAD-ChestCT (*n* = 3,630) and MianYang (*n* = 505) datasets. Since the RAD-ChestCT dataset provides only abnormality labels for classification without the original free-text radiology reports, it was exclusively utilized to evaluate the clinical efficacy of the generated findings, following CT-CLIP. For the MianYang dataset, the original Chinese radiology reports were translated and reformatted into standardized English reports using GPT-4.1[76] (translation prompts in Supplementary Fig. 3) to serve as ground-truth (GT) texts.

We used the NSCLC-Radiomics (LUNG1) cohort[44], comprising 422 patients with CT images, manually delineated gross tumour volumes (GTVs) and survival information. For lung tumour segmentation, patients were randomly divided into a development set of 22 patients, comprising 18 training patients and four validation patients, and a held-out test set of 400 patients. Survival prediction was evaluated separately using cross-validation. Overall survival analysis included all 422 patients and used survival time and death-event status. For two-year survival prediction, patients with follow-up of at least 730 days were classified as surviving to two years, whereas those who died before 730 days were classified as not surviving to two years. Two patients censored before 730 days were excluded, leaving 420 patients.

All scans were first resampled to a uniform voxel spacing of $1 \times 1 \times 3$ mm following the RadGenome-Chest CT protocol. Voxel intensities were then clipped to the [0.5th, 99.5th] percentile range to suppress extreme outliers, enhanced with contrast-limited adaptive histogram equalization (CLAHE; clip limit = 2.0) and linearly rescaled to [0, 1]. All volumes, together with their associated 7-channel pseudo-label masks used for CT-RATE pre-training, were resized to a fixed resolution of $64 \times 128 \times 128$ voxels (depth×height×width). Image volumes and pseudo-label masks were resized using trilinear interpolation, and only mask voxels with an interpolated value of 1 were retained as foreground.

**Ethics approval and informed consent**

The protocol for the MianYang cohort was approved by the Biomedical Ethics Committee of Mianyang Central Hospital, Mianyang, China (approval no. S20260386-01). The analysis was conducted in accordance with the Declaration of Helsinki. The requirement for informed consent was waived by the committee owing to the retrospective nature of the study.

**EXACT architecture overview**

EXACT is built on Y-Mamba (Supplementary Fig. 9), a Y-shaped multi-task architecture adapted from SegMamba[48] for joint organ segmentation and anomaly detection in 3D chest CT. Y-Mamba consists of a shared image encoder, an organ segmentation decoder and an anomaly detection decoder. Specifically, the image encoder adopts a multi-resolution encoding strategy, leveraging 3D convolutional layers (Conv3D) with LeakyReLU activation and Instance Normalization (InstanceNorm3D) to extract spatial and contextual features. To enhance representational power, we incorporate Gated Spatial Convolution (GSC) and Mamba layers at deeper levels. The downsampling process is implemented via strided convolutions, progressively reducing spatial resolution while increasing feature channels, resulting in hierarchical feature representations.

The GSC module (Supplementary Fig. 9b) enhances spatial representation learning by aggregating features across multiple receptive fields. The outputs of the two branches (i.e., a serial 3×3×3 convolutional branch and a parallel 1×1×1 bypass branch) are summed and then projected by a 1×1×1 convolution. The projected features are subsequently fused with the input through a residual connection, thereby aggregating spatial information across multiple receptive fields.

Each Mamba layer (Supplementary Fig. 9c) applies Layer Normalization to the input sequence before Mamba-based sequence modeling with a residual connection:

$$\mathbf{x}_{\mathbf{i+1}} = \text{Unflatten}\left(\text{Mamba}\left(\text{Flatten}\left(\text{LayerNorm}\left(\boldsymbol{x}_{\boldsymbol{i}}\right)\right)\right)\right) + \mathbf{x}_{\mathbf{i}}, \tag{1}$$

where $\boldsymbol{x}_{\boldsymbol{i}}$ denotes the input feature map at the $\boldsymbol{i}$-th layer. Within the Mamba block, the normalized input is linearly projected and split into two parallel branches,

$$\boldsymbol{y}, \boldsymbol{z} = \text{split}\left(\text{linear}\left(\boldsymbol{x}_{\boldsymbol{i}}\right)\right), \tag{2}$$

where $\boldsymbol{y}$ is the sequence-modeling branch and $\boldsymbol{z}$ is the gating branch. Specifically, $\boldsymbol{y}$ is processed by a 1D convolution, SiLU activation, and a selective state-space model (SSM):

$$\widetilde{\boldsymbol{y}_{\boldsymbol{i}}} = \text{SSM}\left(\text{SiLU}\left(\text{Conv1D}(\boldsymbol{y})\right)\right), \tag{3}$$

while $\boldsymbol{z}$ is activated and serves as a multiplicative gate that modulates $\widetilde{\boldsymbol{y}_{\boldsymbol{i}}}$ through element-wise multiplication. The gated output is then projected back to the original feature dimension via a linear layer, reshaped to the original spatial dimensions and added to the input to produce $\boldsymbol{x}_{\boldsymbol{i+1}}$. Each Mamba layer combines efficient long-range sequence modeling with local convolutional feature mixing, supporting volumetric representation learning.

The organ segmentation decoder follows an encoder-decoder structure, where high-level semantic features are progressively upsampled and concatenated with corresponding low-level features via skip connections. Transposed convolutions (TransConv) are used for upsampling, and each decoder stage applies a residual convolutional block to refine the concatenated features. The final segmentation mask is generated through a 7-channel output layer, corresponding to the six target thoracic structures and one global foreground channel assigned to medical material and arterial wall calcification (Supplementary Table 3).

The anomaly detection decoder operates in parallel, incorporating features from both the encoder and the segmentation decoder at each upsampling stage. The decoder generates 18-channel anomaly-aware maps (AAmaps) that encode the spatial extent and organ-specific context of pathological findings.

**EXACT pretraining algorithm**

To learn comprehensive anatomical and pathological representations from large-scale 3D chest CT data without manual voxel-level annotation, we developed a weakly supervised multi-task pretraining algorithm tailored to the clinical imperative of simultaneous structure recognition and disease detection (Fig. 1a, Fig. 1b). Following region-guided weakly supervised paradigms recently explored in neuroimaging[77], we jointly optimized two complementary objectives: (1) organ segmentation supervised by automatically generated coarse masks to establish spatial anatomical priors, and (2) multi-scale anatomy-guided anomaly detection supervised by report-derived disease pseudo-labels to acquire voxel-level pathological awareness through MIL.

Accurate anatomical delineation serves as the spatial scaffold for subsequent anomaly detection. We therefore incorporate organ segmentation as the primary pretraining objective. The segmentation decoder generates a 7-channel voxel-wise prediction $S \in [0, 1]^{7 \times D \times H \times W}$. Six channels correspond to specific thoracic structures (lung, trachea and bronchi, pleura, mediastinum, heart and esophagus), while the seventh channel encodes a global foreground mask to capture diffuse or systemic abnormalities not confined to a single anatomical region (e.g., medical material and arterial wall calcification; Supplementary Table 3). We leverage coarse pseudo-label masks $G \in \{0,1\}^{7\times D\times H\times W}$ generated by the pretrained SAT model as GT to eliminate the need for manual annotation. The segmentation objective is optimized using the average soft Dice loss across all $C = 7$ channels:

$$\mathcal{L}_{seg} = 1 - \frac{1}{C}\sum_{c=1}^{C}\frac{2\sum_v S_c(v)\cdot G_c(v)}{\sum_v S_c(v) + \sum_v G_c(v) + \epsilon}, \tag{4}$$

where $v$ denotes the voxel index within the CT volume, $S_c(v)$ and $G_c(v)$ represent the predicted probability and pseudo-label for channel $c$ at voxel $v$, respectively, and ϵ is a smoothing constant ( $1\times 10^{-5}$) to ensure numerical stability.

To acquire voxel-level pathological awareness from image-level supervision (i.e., RadBERT-derived pseudo-labels), we used a MIL framework, in which each anatomically constrained region of the AAmap is treated as a bag of voxel-level instances and the corresponding image-level pseudo-label serves as the bag-level label.

Specifically, for each of the 18 disease channels $i$ at a given spatial scale $s$, the predicted organ segmentation mask $S_{c(i)}$ is first interpolated to the corresponding resolution via trilinear interpolation and then element-wise multiplied with the raw AAmap $A_i^{(s)}$ to yield the anatomically constrained AAmap:

$$\widehat{Y_i^{(s)}}(v) = A_i^{(s)}(v)\cdot S_{c(i)}^{(s)}(v), \tag{5}$$

where $c(i)$ denotes the organ channel assigned to disease $i$ according to a predefined disease-to-organ mapping (Supplementary Table 3). By restricting the MIL instance pool to anatomically plausible regions, the masking operation confines each disease channel to its clinically relevant anatomy (e.g., pericardial effusion to the heart, lung nodules to the lung parenchyma) and suppresses spurious activations in irrelevant tissue.

Subsequently, within each masked region, the $k$ voxels with the highest predicted anomaly scores are selected and their arithmetic mean is computed to yield a scalar bag-level prediction:

$$\text{Top-}k^{(s)}\left(\widehat{Y_i^{(s)}}\right) = \frac{1}{k}\sum_{v \in \mathcal{V}_k^{(s)}} \widehat{Y_i^{(s)}}(v), \tag{6}$$

where $\mathcal{V}_k^{(s)}$ denotes the index set of the top-$k$ voxels at scale $s$. Since a positive scan is expected to contain at least one truly abnormal voxel, the top-$k$ selection identifies the highest-responding voxels as the most probable disease evidence. During pre-training, driving their aggregated score toward the volume-level pseudo-label implicitly encourages the network to concentrate high anomaly activations on genuinely pathological regions, so that voxel-level localization emerges from image-level supervision alone.

To balance sensitivity to focal lesions with adequate coverage of diffuse pathology, the loss is computed at two complementary spatial resolutions derived from successive decoder stages. A low-resolution branch ($s = 1$, $D/2 \times H/2 \times W/2$, $k_{\text{low}} = 6$) operates on deeper feature maps with broader receptive fields, capturing globally salient anomalous regions while providing a training signal that is robust to single-voxel noise. A high-resolution branch ($s = 2$, $D \times H \times W$, $k_{\text{high}} = 48$) preserves fine-grained spatial detail and extends anomaly coverage beyond lesion cores to peripheral margins. The value of $k_{\text{high}}$ is proportionally scaled from $k_{\text{low}}$ by the volumetric expansion factor (i.e., $2^3 = 8$) to ensure that each branch samples an approximately equivalent physical volume fraction at its respective resolution (Supplementary Figs. 5 and 9a).

Given the highly skewed prevalence $f_i$ of the 18 target abnormalities in the training datasets, ranging from 7.1% for pericardial effusion to 45.5% for lung nodule (Supplementary Table 1), positive-class weights are dynamically derived from training-set disease frequencies as $w_{\text{pos},i} = (1 - f_i)/f_i$, with the negative-class weight fixed at $w_{\text{neg}} = 1.0$. For each disease $i$ at scale $s$, the loss is defined as weighted binary cross-entropy:

$$\mathcal{L}_{\text{abn},i}^{(s)} = -\left[w_{\text{pos},i} \cdot y_i \cdot \log\left(\text{Top-}k^{(s)}\left(\widehat{Y_i^{(s)}}\right) + \epsilon\right) + w_{\text{neg}} \cdot (1 - y_i) \cdot \log\left(1 - \text{Top-}k^{(s)}\left(\widehat{Y_i^{(s)}}\right) + \epsilon\right)\right], \tag{7}$$

where $y_i \in \{0,1\}$ is the RadBERT-derived binary pseudo-label for disease $i$ and $\epsilon = 1\times 10^{-6}$ is a smoothing constant for numerical stability.

Finally, the total anomaly detection loss is obtained by averaging over all $N = 18$ disease channels and both scales ($|\mathcal{S}| = 2$):

$$\mathcal{L}_{\text{abn}} = \frac{1}{N\cdot|\mathcal{S}|}\sum_{s\in\mathcal{S}}\sum_{i=1}^{N}\mathcal{L}_{\text{abn},i}^{(s)}. \tag{8}$$

The overall pre-training objective combines the organ segmentation loss and the anomaly detection loss into a single weighted sum:

$$\mathcal{L}_{\text{total}} = \lambda_t \cdot \mathcal{L}_{\text{seg}} + \mathcal{L}_{\text{abn}}, \tag{9}$$

where $\lambda_t$ is a time-dependent weighting coefficient that governs the relative contribution of the two tasks throughout training. Rather than fixing this coefficient, we adopt an exponential decay schedule that enforces a curriculum in which the model first consolidates robust anatomical priors before progressively shifting its capacity toward fine-grained pathological pattern learning:

$$\lambda_t = \max\left(\lambda_{\text{init}} \cdot e^{-\gamma\cdot t},\ 0.5\right), \tag{10}$$

where $t = \text{epoch}/\text{total_epochs} \in [0, 1]$ denotes the normalized training progress, $\lambda_{\text{init}} = 2.0$ is the initial segmentation weight, and $\gamma = 10.0$ controls the decay rate.

EXACT was implemented in PyTorch and pre-trained on a cluster equipped with 4 NVIDIA A800 (80 GB) GPUs. Of the 24,128 scans in the CT-RATE training partition, 22,620 were used for model optimization and the remaining 1,508 were reserved for monitoring validation loss during pre-training. EXACT was trained for 25 epochs using the AdamW optimizer with parameters $\beta_1 = 0.9$, $\beta_2 = 0.999$, and a weight decay of $1\times 10^{-2}$. The initial learning rate was set to $1\times 10^{-3}$ and adjusted via a cosine annealing schedule ($T_{max} = 50$, $\eta_{min} = 1\times 10^{-5}$). The per-GPU batch size was 2, yielding an effective batch size of 8 across 4 GPUs. The dynamic weighting parameters were set to $\lambda_{\text{init}} = 2.0$ and $\gamma = 10.0$. The entire pre-training procedure required approximately 43 hours.

Supplementary Fig. 5 compares the high-resolution AAmap alone with an overlay of the high-resolution map and the upsampled low-resolution map. Both visualizations are generated by the same pre-trained model under an identical training configuration. During inference, the high-resolution AAmaps are used directly. While the low-resolution branch broadens the spatial extent of anomaly responses, its coarser receptive field also disperses activation into non-pathological tissue, which lowers voxel-level precision. The high-resolution branch therefore provides the closer approximation to the true lesion extent and is retained for all downstream tasks, while the low-resolution branch contributes through the supervisory signal it provides during pre-training.

**EXACT for multi-disease diagnosis**

Under the zero-shot setting (Fig. 2a), the pre-trained EXACT model is applied directly to CT volumes without any task-specific training. For each disease channel $i$, the anatomically constrained AAmap of the high-resolution branch is aggregated via Top-k pooling to yield the final prediction score $p_i$:

$$p_i = Top - k(Y_i) \tag{11}$$

To further improve diagnostic accuracy (Fig. 2b), a lightweight classifier was trained on the EXACT-generated AAmaps. Specifically, the frozen pre-trained EXACT model is utilized to extract an 18-channel AAmap for each training sample in the CT-RATE dataset. A SegMamba[48] encoder, coupled with a linear classification head, is then trained end-to-end on these spatial representations using the corresponding pseudo-labels for supervision. During inference, validation CT volumes are processed by the frozen EXACT model to generate their AAmaps, which are then analyzed by the trained classifier to output disease-specific probabilities.

For training the AAmap classifier, AdamW[78] was employed with an initial learning rate of $1\times 10^{-4}$, $\beta_1 = 0.9$, $\beta_2 = 0.999$, and a weight decay of $1\times 10^{-4}$. The learning rate was adjusted using a cosine annealing schedule. To address the highly imbalanced disease prevalence across the 18 target abnormalities, we employed a frequency-weighted binary cross-entropy (BCE) loss, where per-class positive weights were dynamically computed from training-set disease frequencies as $w_{\text{pos},i} = (1 - f_i)/f_i$. A dropout rate of 0.3 was applied to the classification head to mitigate overfitting. Training was conducted for 100 epochs on a single NVIDIA A800 GPU with 900 randomly sampled scans per epoch.

EXACT was compared against SOTA 3D medical vision FMs across all evaluation datasets (CT-RATE, RAD-ChestCT, and MianYang). Under the zero-shot setting, comparisons included CT-CLIP[31], fVLM[32], MedVista3D[14], T3D[33], Merlin[10], RadZero3D[45], and BIUD[46]. Under the fine-tuning setting, comparisons included CT-CLIP (VocabFine and ClassFine), T3D (Fine-tuning), and the supervised baseline CT-Net[39].

Specifically: (1) CT-CLIP[31] performs contrastive language-image pre-training on the CT-RATE dataset to align entire CT volumes with free-text radiology reports in a shared embedding space, serving as the principal multimodal baseline for 3D CT understanding. We evaluated three variants: (i) CT-CLIP (Zero-shot), which performs zero-shot classification via cosine similarity between visual and textual embeddings; (ii) CT-CLIP (VocabFine), a fine-tuned variant that preserves open-vocabulary inference capability; and (iii) CT-CLIP (ClassFine), a fixed-category fine-tuned variant with a dedicated classification head. (2) fVLM[32] introduces fine-grained vision-language modeling by explicitly aligning anatomical regions of CT images with corresponding textual descriptions extracted from radiology reports, achieving anatomy-level image understanding through organ-guided local contrastive learning. (3) MedVista3D[14] employs a multi-scale semantic enhancement framework that integrates both local and global image-text alignment to capture fine-grained representations within the context of whole-volume scans, leveraging organ segmentation masks for region-level grounding. (4) T3D[33] enhances 3D medical vision-language pre-training through text-informed multi-view alignment (TMA), which addresses the neglect of local visual representations inherent in conventional CLIP-style global alignment. Published results for both T3D (Zero-shot) and T3D (Fine-tuning) were included. (5) Merlin[10] is a 3D vision-language FM that extends beyond unstructured radiology reports by additionally incorporating structured diagnostic codes from electronic health records (EHR) for self-supervised pre-training, enabling joint representation learning from heterogeneous clinical data sources. (6)

RadZero3D[45] leverages self-supervised video pre-training model (V-JEPA) to process volumetric CT data as pseudo-video sequences and achieves zero-shot chest CT interpretation by aligning extracted volumetric patch features with "finding statements" parsed from radiology reports. (7) BIUD[46] addresses the scarcity of paired CT–report data by distilling diagnostic knowledge from 2D chest X-ray expert models into 3D CT representations through a bootstrapping framework, thereby transferring established 2D medical knowledge to volumetric image understanding. (8) CT-Net[39] serves as a purely supervised vision baseline.

For evaluation on MianYang, we replaced its original 16-class classification head with an 18-class head to match the target abnormality set, and trained the model on the CT-RATE dataset following the official configuration (https://github.com/rachellea/ct-net-models).

For the results summarized in Fig. 3 and Supplementary Table 4, CT-CLIP (Zero-shot, VocabFine and ClassFine) and fVLM were independently evaluated on all three datasets using their officially released model weights. Results for CT-Net, MedVista3D, T3D (Zero-shot and Fine-tuning), RadZero3D and BIUD on CT-RATE were obtained from published studies. On RAD-ChestCT, CT-CLIP (Zero-shot, VocabFine and ClassFine), fVLM and Merlin were independently evaluated, whereas results for CT-Net, MedVista3D and BIUD were obtained from published studies. All baseline results reported on MianYang were independently evaluated: CT-CLIP, fVLM and Merlin used their officially released model weights, whereas CT-Net was retrained as described above. CT-Net results on CT-RATE and RAD-ChestCT were taken from the CT-CLIP publication[31]; the other publication-reported results were obtained from the respective model publications. MedVista3D, T3D, RadZero3D and BIUD were not evaluated on MianYang because their model weights or source code were unavailable at the time of this study. Publication-reported results were excluded from statistical comparisons requiring individual predictions.

Following the evaluation protocol of CT-CLIP, we determined the binary classification threshold for each target abnormality as the point on the receiver operating characteristic (ROC) curve closest to the upper-left corner (0, 1). Thresholds were selected on the full validation set of each cohort (CT-RATE, RAD-ChestCT and MianYang) for EXACT and the independently evaluated comparison methods, without an independent calibration split.

We adopted AUROC, F1 score, and accuracy as the primary evaluation metrics for multi-disease diagnosis. AUROC quantifies the discriminative ability of the model across all classification thresholds by computing the area under the curve that plots the true positive rate (sensitivity) against the false positive rate (1 − specificity). The F1 score is the harmonic mean of precision and recall:

$$\mathrm{F1} = 2 \times \frac{\mathrm{Precision} \times \mathrm{Recall}}{\mathrm{Precision} + \mathrm{Recall}}, \tag{12}$$

where $\mathrm{Precision} = \mathrm{TP} / (\mathrm{TP} + \mathrm{FP})$ and $\mathrm{Recall} = \mathrm{TP} / (\mathrm{TP} + \mathrm{FN})$, with TP, FP, and FN denoting true positives, false positives, and false negatives, respectively. Accuracy is defined as:

$$\mathrm{Accuracy} = (\mathrm{TP} + \mathrm{TN}) / (\mathrm{TP} + \mathrm{TN} + \mathrm{FP} + \mathrm{FN}), \tag{13}$$

where TN denotes true negatives. For independently evaluated results, metrics were computed per disease and then macro-averaged across all evaluable abnormalities for each dataset. The 95% CIs were estimated using 1,000 bootstrap resamples with replacement. *P* values reported in Fig. 3 and Supplementary Table 4 compare EXACT (Zero-shot) with EXACT (Fine-tuning) within each dataset using two-sided bootstrap hypothesis tests with 2,000 resamples.

**EXACT for anomaly localization**

Under the zero-shot setting (Fig. 4a), the pre-trained EXACT model is applied directly to CT volumes without training using manual voxel-level annotations. Specifically, for each input volume, the parallel decoders simultaneously generate 18 disease-specific AAmaps and the corresponding anatomical organ segmentation masks. After that, each AAmap channel $A_i$ is element-wise multiplied with the organ segmentation mask $S_{c(i)}$ assigned to disease $i$ according to the predefined disease-to-organ mapping (Supplementary Table 3), yielding the anatomically constrained AAmap $\widehat{Y_i} = A_i \odot S_{c(i)}$. Task-relevant disease channels are then selected based on the pathological scope of each evaluation dataset. For the ReXGroundingCT dataset (ReX-Train and ReX-Val), which encompasses diverse lung and pleural abnormalities, 9 disease channels were aggregated, including pleural effusion (PLE), bronchiectasis (BE), peribronchial thickening (PBT), interlobular septal thickening (ILST), ATE, GGO, consolidation (CON), mosaic attenuation pattern (MAP), and pulmonary fibrotic sequela (PFS). For the COVID-19 and MosMed datasets, which target COVID-19-related pulmonary lesions, 4 channels most closely associated with COVID-19 pathology were selected, including ATE, GGO, MAP, and CON. The selected channels are aggregated via voxel-wise averaging to produce a comprehensive anomaly map:

$$\mathcal{A}(v) = \frac{1}{|\mathcal{D}_t|}\sum_{i\in\mathcal{D}_t}\widehat{Y}_i(v), \tag{14}$$

where $\mathcal{D}_t$ denotes the set of task-relevant disease channels, and $v$ indexes the voxel position. The resulting anomaly map is subsequently binarized using a dataset-specific threshold $\tau$ to yield the final localization mask:

$$\widehat{Y}(v) = \mathbb{1}[\mathcal{A}(v) > \tau], \tag{15}$$

where $\tau = 0.020$ for ReXGroundingCT and MosMed, and $\tau = 0.005$ for COVID-19.

To enhance anomaly localization with limited supervision (Fig. 4c), we adapted the pre-trained EXACT model for end-to-end supervised fine-tuning. Specifically, the pre-trained image encoder and anomaly detection decoder were retained, and a terminal $1 \times 1 \times 1$ convolutional layer was appended to compress the 18-channel AAmaps into a single-channel probability map. The resulting model was then fine-tuned end-to-end using expert-annotated voxel-level segmentation masks.

To address the challenge of localizing small and morphologically diverse lesions in 3D chest CT, a hybrid loss function combining Tversky loss[79] and Focal loss was employed:

$$\mathcal{L}_{\text{seg}} = w_T \cdot \mathcal{L}_{\text{Tversky}} + w_F \cdot \mathcal{L}_{\text{Focal}} \tag{16}$$

where $w_T = 0.7$ and $w_F = 0.3$ are the respective loss weights. The Tversky loss asymmetrically penalizes false positives and false negatives to prioritize recall for small lesions:

$$\mathcal{L}_{\text{Tversky}} = 1 - \frac{\sum_x p(x)\cdot g(x) + \epsilon}{\sum_x p(x)\cdot g(x) + \alpha\sum_x p(x)\cdot\left(1-g(x)\right) + \beta\sum_x\left(1-p(x)\right)\cdot g(x) + \epsilon}, \tag{17}$$

where $p(x) \in [0,1]$ denotes the predicted probability at voxel $x$, $g(x) \in \{0,1\}$ is the corresponding ground-truth binary label, $\alpha$ and $\beta$ control the relative penalty for false positives and false negatives respectively (set to $\alpha = 0.3$, $\beta = 0.7$ to prioritize recall), and $\epsilon = 1 \times 10^{-5}$ is a smoothing constant for numerical stability. The Focal loss mitigates class imbalance by downweighting the contribution of easily classified voxels:

$$\mathcal{L}_{\text{Focal}} = -\frac{1}{N}\sum_{x} \alpha_f \cdot \left(1 - p_t(x)\right)^{\gamma_f} \cdot \log\left(p_t(x)\right), \tag{18}$$

where $p_t(x) = p(x)$ if $g(x) = 1$ and $p_t(x) = 1 - p(x)$ otherwise, $\alpha_f = 0.75$ is the balancing factor, $\gamma_f = 2.0$ is the focusing parameter, and $N$ is the total number of voxels.

For fine-tuning EXACT-Seg, AdamW was employed with momentum parameters $\beta_1 = 0.9$ and $\beta_2 = 0.999$, and a weight decay of $1 \times 10^{-2}$. For fine-tuning on ReX-Train, the initial learning rate was set to $1 \times 10^{-4}$. ReX-Train was internally split at a 15:1 ratio; the model was trained on the 15/16 partition for 100 epochs, and the checkpoint with the best DSC on the held-out 1/16 subset was selected for final evaluation on ReX-Val. For the external COVID-19 and MosMed datasets, an initial learning rate of $1 \times 10^{-3}$ was adopted and training was conducted for 400 epochs with the final checkpoint used for evaluation. The learning rate was adjusted via a cosine annealing schedule with a minimum learning rate of $1 \times 10^{-5}$, and the per-GPU batch size was set to 4.

Under the zero-shot setting, EXACT was compared against three publicly available models capable of producing spatial localization outputs. (1) BiomedParse-v2[47] is a universal biomedical image segmentation FM that supports zero-shot segmentation via task-specific text prompts. For evaluation, we used the prompts "lesions of lungs and pleura" on the ReX-Train and ReX-Val datasets and "COVID19 lesion" on the COVID-19 and MosMed datasets. (2) CT-CLIP and (3) fVLM, as vision-language FMs, do not natively produce segmentation outputs. Thus, we employed gradient-weighted class activation mapping (Grad-CAM)[35] to derive spatial heatmaps from both models. To ensure fair comparison, the binarization threshold for these heatmaps was optimized via grid search to maximize the DSC on each dataset.

Under the supervised fine-tuning setting, EXACT-Seg was compared against two SOTA medical image segmentation models: (1) SegMamba[48] is a 3D segmentation model based on the Mamba architecture, designed for long-range sequential modeling in volumetric medical images. SegMamba was trained from random initialization. (2) RWKV-UNet[49] integrates the linear-attention RWKV mechanism into the U-Net framework, augmenting long-range dependency modeling via global-local spatial awareness blocks, and has recently shown top-tier performance across diverse medical datasets in the U-Bench benchmark[80]. We initialized RWKV-UNet with the officially released net_B pre-trained weights (https://github.com/juntaoJianggavin/RWKV-UNet). Since RWKV-UNet is natively a 2D architecture, we adopted a slice-by-slice inference strategy for 3D volumetric processing. Both baseline models were trained with the same data splits, training epochs, and checkpoint selection criteria as EXACT-Seg to ensure a fair comparison.

We adopted the DSC and AUPR as the primary evaluation metrics for anomaly localization. DSC quantifies the spatial overlap between the predicted binary segmentation mask $S$ and the ground-truth binary mask $G$:

$$\mathrm{DSC} = \frac{2|S \cap G|}{|S| + |G| + \epsilon}, \tag{19}$$

where $|S|$ and $|G|$ denote the total number of voxels in the predicted and GT regions, respectively, and $\epsilon$ is a small constant for numerical stability. AUPR summarizes the trade-off between precision and recall across all prediction thresholds, providing a threshold-independent assessment of localization quality that is particularly informative under class imbalance.

Given the difficulty of achieving high absolute overlap scores for fine-grained 3D localization without dense supervision, we additionally reported Hit Rates at 5% and 10% thresholds (Hit Rate@5% and Hit Rate@10%), defined as the proportion of test samples for which the DSC or AUPR exceeds the corresponding threshold. All metrics were computed per sample, and 95% CIs were estimated using 2,000 bootstrap resamples with replacement. Statistical significance between the top two performing methods was assessed using the two-sided Wilcoxon signed-rank test.

## EXACT-CHAT for report generation

Adapted from the LLaVA framework[57], EXACT-CHAT integrates the frozen EXACT image encoder, a multimodal projector module, a structured diagnostic prior pathway, and an LLM to enable visually grounded radiology report generation (Fig. 7a). Specifically, the frozen EXACT image encoder compresses 3D chest CT volumes into a low-dimensional feature space enriched with learned anatomical and pathological representations. The multimodal projector, based on the approach described in CT-CHAT, consists of an attention-pooling layer followed by a two-layer multilayer perceptron (MLP). In parallel, diagnostic priors are derived by passing the EXACT-generated AAmaps through a frozen AAmap encoder and classifier (pre-trained in the fine-tuning setting for multi-disease diagnosis) to obtain discrete disease-state predictions for all 18 target abnormalities. These predictions are formatted as structured text tokens encoding the presence or absence of each disease (Fig. 7a). The LLM backbone, Meta LLaMA-3.1-8B-Instruct[58], receives the concatenation of visual tokens, diagnostic prior tokens, and user instruction tokens to generate the initial radiology report.

EXACT-CHAT was trained on 4 NVIDIA A800 (80 GB) GPUs using the DeepSpeed framework[81]. Training followed a two-stage protocol. In the feature alignment stage, both the visual encoder and LLM were frozen, and only the multimodal projector was optimized to align visual and language representations. In the instruction fine-tuning stage, the LLM was adapted using low-rank adaptation (LoRA)[82] with rank $r = 128$ and scaling factor $\alpha = 256$, while the projector was fully fine-tuned. The AdamW optimizer was used with a learning rate of $2 \times 10^{-5}$ for both the LLM and the projector, a cosine annealing learning rate schedule, and a warmup ratio of 0.03. EXACT-CHAT was trained for 1 epoch on the CT-RATE training partition to prevent overfitting and preserve generalization to unseen data.

For visual grounding, EXACT-CHAT uses the AAmaps produced by the frozen EXACT backbone to provide anatomically constrained, voxel-level spatial evidence for the pathological findings described in generated reports (Fig. 10).

EXACT-CHAT was compared against SOTA 3D medical multimodal models for radiology report generation on the evaluation datasets (CT-RATE, RAD-ChestCT, and MianYang). (1) CT-CHAT[31] is a generalist 3D chest CT vision-language model that extends CT-CLIP by coupling its pretrained volumetric vision encoder with a large language model (LLM) through a multimodal projector, and is instruction-tuned on CT-RATE to support CT-grounded report generation. We included two variants: (i) CT-CHAT (LLaMA-3.1-70B), which employs a larger LLM backbone, and (ii) CT-CHAT (w/ nodule attributes), which incorporates structured nodule-specific descriptors into the generation pipeline. (2) RadFM[50] is a generalist radiology vision-language FM that supports both 2D and 3D medical images via a unified generative architecture. (3) CT2Rep[51] is an automated radiology report generation framework specifically designed for 3D chest CT volumes, employing a 3D auto-regressive causal vision transformer for extracting volumetric features, coupled with a relational memory-driven transformer decoder to ensure clinically accurate text generation. (4) M3D[52] is a 3D medical multimodal LLM for volumetric medical image analysis, leveraging a CLIP-like pretrained 3D ViT image encoder and a 3D spatial pooling perceiver to bridge high-dimensional volumetric features with an LLM, enabling efficient and spatially aware text generation from 3D scans. (5) Reg2RG[53] is a region-guided referring and grounding framework for 3D chest CT report generation. It leverages anatomical masks to extract region-specific local features, and introduces a local feature decoupling strategy to preserve both high-resolution texture details and geometric information (e.g., size and position) with low computational overhead. (6) CT-GRAPH[54] is a hierarchical graph attention network for anatomy-guided CT report generation. It models radiological knowledge by structuring anatomical regions into a hierarchical graph, linking fine-grained organ-specific features to coarser anatomical systems and a global patient context. (7) BTB3D[55] is a 3D medical vision-language framework designed to improve volumetric tokenization for long, high-resolution CT scans. Instead of relying on contrastive pretraining, it introduces a causal convolutional encoder-decoder with frequency-aware discrete volumetric tokens, enabled by a 3D Haar wavelet transform and lookup-free quantization. (8) Merlin[10] (see EXACT for multi-disease diagnosis). (9) Hulu-Med[56] is a generalist medical multimodal LLM that unifies text, 2D, 3D, and video understanding within a single patch-based vision encoder-LLM architecture. Trained progressively on 16.7M public and synthetic samples, it leverages medical-aware token reduction to efficiently process volumetric and temporal inputs for tasks such as report generation and clinical reasoning. (10) MedVista3D[14] (see EXACT for multi-disease diagnosis). (11) T3D[33] (see EXACT for multi-disease diagnosis).

Baseline configurations were independently evaluated using publicly released code and corresponding model weights where available. For direct comparisons on CT-RATE, we included report-generation checkpoints trained on CT-RATE.

On CT-RATE, CT-CHAT (LLaMA-3.1-70B), Reg2RG, CT-GRAPH and BTB3D were independently evaluated. Results for RadFM, CT2Rep, M3D, CT-CHAT with nodule attributes, MedVista3D and T3D were obtained from published studies. On RAD-ChestCT, RadFM, M3D, CT-CHAT, Merlin, BTB3D, Reg2RG, Hulu-Med and CT-GRAPH were independently evaluated, whereas CT2Rep results were obtained from the literature. All baseline results reported on MianYang were independently evaluated.

The official RadFM, M3D, Merlin and Hulu-Med checkpoints used in this study had not been trained on CT-RATE and were excluded from the independently evaluated CT-RATE comparison. The publication-reported CT-RATE results for RadFM and M3D corresponded to models fine-tuned on CT-RATE in the Reg2RG study[53]. These configurations differed from the official checkpoints used in our external evaluations. Publication-reported results were retained as contextual references and were not included in post-hoc refinement or statistical comparisons requiring individual predictions. Publication-reported and independently evaluated results are distinguished in Fig. 8 and Supplementary Table 6.

Independently generated reports were assessed using a common evaluation pipeline, with model-specific input processing. For CT-CHAT on RAD-ChestCT, CT intensities were clipped to −1,000 to 1,000 Hounsfield units and divided by 1,000, following the official implementation. BTB3D used the fixed-grid configuration on CT-RATE and the native-grid configuration on RAD-ChestCT and MianYang. Merlin report generation used the full configuration covering 13 anatomical regions.

Report generation was evaluated separately before and after disease-conditioned post-hoc refinement. For EXACT-CHAT, GPT-4.1[76] refined the initial report using the upstream disease predictions to correct missed findings and hallucinated diagnoses while preserving the original reporting style and structure. The same refinement procedure was applied to the independently evaluated baseline models that could provide their own disease predictions: (1) CT-CHAT, Reg2RG and BTB3D on CT-RATE; and (2) RadFM, M3D, CT-CHAT, Merlin, BTB3D, Reg2RG and Hulu-Med on RAD-ChestCT and MianYang. CT-GRAPH was excluded because its released report-generation implementation provided neither a disease-classification head nor a visual question-answering interface for obtaining these predictions. CT2Rep results obtained from the literature lacked individual generated reports and were therefore not included in refinement. For RadFM, M3D, Merlin and Hulu-Med, disease scores were obtained through disease-specific yes/no visual question answering using the models' first-token logits. For each abnormality, the binary decision threshold was selected by maximizing Youden's J statistic against the reference abnormality labels in the corresponding evaluation cohort. The same cohort was used for threshold selection and subsequent report-generation evaluation, without an independent calibration split. The resulting binary disease predictions were used to condition GPT-4.1 report refinement. For each model included in refinement, the inputs comprised its original generated report and its own disease predictions, using a common prompt, fixed in-context examples and a temperature of 0.2 (Supplementary Fig. 4). Refinement was applied without updating model weights. Results before refinement are presented in Fig. 8 and Supplementary Table 6; results after disease-conditioned refinement are presented in Fig. 9 and Supplementary Table 7.

Report generation quality was assessed using standard NLG metrics and clinical efficacy metrics, separately before and after post-hoc refinement. For lexical quality, BLEU-1[83], METEOR[84], CIDEr[85], and ROUGE-L[86] were computed between generated and reference reports on CT-RATE and MianYang. For independently evaluated results, NLG metrics were calculated using PTB tokenization. NLG metrics were not calculated on RAD-ChestCT because reference reports were unavailable. Clinical efficacy was assessed using RadBERT-F1, RadBERT-Precision and RadBERT-Recall. RadBERT was used to extract structured disease labels from generated reports, which were compared with the corresponding reference labels. On RAD-ChestCT, reference labels were provided by the dataset. For independently evaluated results, clinical efficacy metrics were macro-averaged across 18 abnormalities on CT-RATE, 16 on RAD-ChestCT and 17 on MianYang. Coronary artery wall calcification and mosaic attenuation pattern were excluded on RAD-ChestCT, and mosaic attenuation pattern was excluded on MianYang. For independently evaluated results, 95% confidence intervals were estimated using 2,000 bootstrap resamples with replacement. Statistical comparisons used two-sided bootstrap hypothesis tests with 2,000 resamples for the model pairs and metrics indicated in Figs. 8 and 9. The *P* values in Supplementary Tables 6 and 7 refer specifically to RadBERT-F1 comparisons with the baseline identified for each dataset. Publication-reported results were excluded from statistical comparisons requiring individual predictions.

**Frozen-encoder transfer to lung tumour segmentation and survival prediction**

For lung tumour segmentation, we adapted EXACT by freezing its pre-trained encoder and optimizing only the segmentation decoder and output head. During training, spatial augmentations were applied consistently to CT images and tumour masks, with all slices within a volume sharing the same geometric transformation. Flips along each of the two axial in-plane directions were applied with a probability of 0.5. Affine transformations were applied with a probability of 0.8 and included rotations within ±12°, scaling factors of 0.9–1.1 and normalized translations within ±0.06. Images and masks were interpolated using linear and nearest-neighbour interpolation, respectively. Intensity augmentations included gamma correction with values of 0.8–1.25, intensity scaling by factors of 0.9–1.1 and intensity shifts within ±0.05, each applied with a probability of 0.5. Gaussian noise with a standard deviation of 0.02 was added with a probability of 0.3. Six augmented views were generated per training patient in each epoch.

Comparison methods included randomly initialized SegMamba, ImageNet-pre-trained RWKV-UNet and pre-trained BiomedParse-v2. SegMamba and RWKV-UNet were trained end-to-end using the same patient split, augmentation procedures and test-time augmentation as EXACT-Seg. RWKV-UNet generated predictions slice-wise, which were assembled into volumetric probability maps. For EXACT-Seg, SegMamba and RWKV-UNet, validation and testing used four-view test-time augmentation, comprising the original volume, a flip along each of the two axial in-plane directions and a simultaneous flip along both directions. Predicted probability maps were transformed back to their original orientation and averaged voxel-wise. The averaged probability maps were thresholded at 0.5 to generate binary segmentation masks. BiomedParse-v2 was evaluated directly using its pre-trained weights and the fixed prompt "lung tumor". Its Dice similarity coefficient (DSC) was calculated from the binary output after the model's existence-gating step, whereas its area under the precision–recall curve (AUPR) was calculated from continuous logits before gating.

All methods were evaluated on the same 400 held-out test patients. DSC and AUPR were computed per patient, with 95% bootstrap confidence intervals for their means. EXACT-Seg and RWKV-UNet, the two highest-performing methods, were compared using paired two-sided *t*-tests. The corresponding results are presented in Fig. 6a,b and Supplementary Table 8.

Both the FMCIB[60] and EXACT encoders remained frozen throughout the survival experiments. Each endpoint was evaluated using five repeats of five-fold cross-validation. Within each repeat, four folds were used for training and the remaining fold was held out for testing, such that every patient was evaluated once in a held-out test set. Overall survival analysis used random fold assignments, whereas two-year survival analysis used outcome-stratified fold assignments. All three feature configurations described below used identical task-specific training and test partitions.

FMCIB features were extracted following the official implementation[60]. The physical centroid of each manually delineated gross tumour volume (GTV) was used to locate the local tumour region, from which the frozen FMCIB encoder extracted a 4,096-dimensional feature vector. The entire preprocessed CT volume from the same patient was also passed through the frozen EXACT encoder to obtain global and tumour-pooled representations.

For the global representation, feature maps from the four EXACT encoder stages and the bottleneck layer were averaged over their spatial dimensions. The resulting vectors, containing 48, 96, 192, 384 and 768 features, were concatenated to form a 1,488-dimensional global representation. For the tumour-pooled representation, the manual GTV aligned with the CT volume was resampled by trilinear interpolation to the 4 × 8 × 8 spatial grid of the bottleneck features, producing a soft weighting map with values between 0 and 1. Each of the 768 feature channels was pooled as:

$$\boldsymbol{g}_{\boldsymbol{c}} = \frac{\sum_{\boldsymbol{v}} \boldsymbol{w}_{\boldsymbol{v}} \boldsymbol{F}_{\boldsymbol{c},\boldsymbol{v}}}{\sum_{\boldsymbol{v}} \boldsymbol{w}_{\boldsymbol{v}} + \mathbf{10^{-6}}}, \tag{20}$$

where $\boldsymbol{F}_{\boldsymbol{c},\boldsymbol{v}}$ denotes the bottleneck feature in channel $c$ at spatial location $v$, and $\boldsymbol{w}_{\boldsymbol{v}}$ denotes the corresponding GTV weight. This yielded a 768-dimensional tumour-pooled representation derived from the spatial features of the entire CT volume.

We compared three feature configurations: (1) FMCIB features alone; (2) FMCIB features combined with EXACT global representations; and (3) FMCIB features combined with EXACT tumour-pooled representations. Features were aligned using patient identifiers. For each training–test split, the FMCIB and EXACT feature blocks were separately standardized using the per-feature means and standard deviations estimated from the training set. The standardized feature blocks were then concatenated:

$$\boldsymbol{z} = \left[\frac{\boldsymbol{l}-\boldsymbol{\mu}_{\boldsymbol{L}}}{\boldsymbol{\sigma}_{\boldsymbol{L}}} ; \frac{\boldsymbol{g}-\boldsymbol{\mu}_{\boldsymbol{G}}}{\boldsymbol{\sigma}_{\boldsymbol{G}}}\right], \tag{21}$$

where $\boldsymbol{l}$ denotes the FMCIB feature vector, $\boldsymbol{g}$ denotes either the global or tumour-pooled EXACT feature vector, and the means and standard deviations were estimated from the corresponding training set. Division was performed element-wise, and the semicolon denotes concatenation. Fusion with EXACT global and tumour-pooled representations produced vectors containing 5,584 and 4,864 features, respectively. The concatenated features were further standardized and passed to a regularized linear prediction model, which jointly learned coefficients for the FMCIB and EXACT features. For each training–test split, standardization parameters were estimated from the training set and kept fixed during cross-validation within the training set for model selection. The same parameters were then applied to the corresponding test set.

Overall survival risk was modelled using an elastic-net-regularized Cox proportional hazards model, with the mixing parameter between the L1 and L2 penalties set to 0.5. For each training–test split, a regularization path was fitted using the training set. The regularization strength was selected by five-fold cross-validation within the training set, based on the mean Harrell's concordance index (C-index). The model fitted on the complete training set at the selected regularization strength was then used to generate risk scores for the test patients.

Two-year survival was modelled using L2-regularized logistic regression with balanced class weights. The inverse regularization parameter *C* was selected from 19 logarithmically spaced values between $\mathbf{10^{-6}}$ and $\mathbf{10^{3}}$ through stratified five-fold cross-validation within the training set, based on the mean area under the receiver operating characteristic curve (AUROC). The selected model was fitted on the complete training set and used to predict the probability of survival to two years for each test patient.

Overall survival discrimination was evaluated using Harrell's C-index, and two-year survival prediction was evaluated using AUROC. Each metric was calculated separately on the test set for each training–test split. Performance was summarized as the mean of the 25 values obtained from five repeats of five-fold cross-validation. Supplementary Table 9 reports the mean ± SD. The 95% confidence intervals shown in Fig. 6c were calculated as:

$$\overline{\boldsymbol{x}} \pm \boldsymbol{t}_{\mathbf{0.975,24}} \frac{\boldsymbol{s}}{\sqrt{\mathbf{25}}}, \tag{22}$$

where $\overline{\boldsymbol{x}}$ and *s* denote the mean and sample standard deviation of the metric values across the 25 training–test splits, respectively, and $\boldsymbol{t}_{\mathbf{0.975,24}}$ is the 97.5th percentile of the t distribution with 24 degrees of freedom. Methods were compared using paired two-sided t-tests on test-set performance from the same training–test splits. Fig. 6c presents comparisons between FMCIB alone and the two fusion settings, as well as between the two fusion settings. The P values in Supplementary Table 9 specifically compare fusion with EXACT tumour-pooled representations against fusion with EXACT global representations.

**Ablation of anatomical confinement**

We evaluated three configurations to assess anatomical confinement (Supplementary Table 10): the EXACT model with confinement during training and inference (A); the same model with confinement removed only at inference (B); and a model retrained without confinement during training or inference (C). Classification was evaluated on CT-RATE (*n* = 1,564 scans), with AUROC, F1 score and accuracy macro-averaged across 18 findings. Localization was evaluated on COVID-19 (*n* = 20 cases). High-resolution anomaly maps for atelectasis, lung opacity, mosaic attenuation pattern and consolidation were averaged voxel-wise across the four disease channels. DSC was calculated after binarization at an absolute threshold of 0.005, whereas AUPR was calculated from continuous anomaly maps. HIT@5% and HIT@10% were defined as the proportions of cases with DSC or AUPR exceeding 0.05 and 0.10, respectively. Comparisons were paired across evaluation cases. For the classification metrics, 95% CIs for each configuration were estimated using 1,000 bootstrap resamples, and CIs for between-configuration differences and P values were obtained using 2,000 paired bootstrap resamples. Confidence intervals and P values for HIT rates were obtained using 2,000 paired bootstrap resamples. Confidence intervals for DSC and AUPR were estimated using 2,000 paired bootstrap resamples. P values for DSC and AUPR were obtained using 20,000 paired permutations.

## Data availability

The in-house chest CT dataset (MianYang) is protected to ensure patient privacy, yet some data can be made available for academic purposes from the corresponding author on reasonable request and with permission from the hospital. Publicly available datasets used in this study include the CT-RATE dataset (https://huggingface.co/datasets/ibrahimhamamci/CT-RATE), the RAD-ChestCT dataset (https://zenodo.org/records/6406114), the ReXGroundingCT dataset (https://huggingface.co/datasets/baharoon/ReXGroundingCT), the COVID-19 dataset (https://zenodo.org/records/3757476), the MosMed dataset (https://mosmed.ai/datasets/covid19_1110), and the NSCLC-Radiomics dataset available through The Cancer Imaging Archive.

## Code availability

The source code and trained model weights from this study are publicly available at https://github.com/JasonW375/EXACT. The source code for Chest-OMDL (https://github.com/JasonW375/Chest-OMDL), Merlin (https://github.com/StanfordMIMI/Merlin), CT-CLIP (https://github.com/ibrahimethemhamamci/CT-CLIP), CT-CHAT (https://github.com/ibrahimethemhamamci/CT-CHAT), fVLM (https://github.com/alibaba-damo-academy/fvlm), SAT (https://github.com/zhaoziheng/SAT), RadBERT (https://github.com/zzxslp/RadBERT), CT-Net (https://github.com/rachellea/ct-net-models), BiomedParse-v2 (https://github.com/microsoft/BiomedParse), SegMamba (https://github.com/ge-xing/SegMamba), RWKV-UNet (https://github.com/juntaoJianggavin/RWKV-UNet), RadFM (https://github.com/chaoyi-wu/RadFM), M3D (https://github.com/BAAI-DCAI/M3D), Reg2RG (https://github.com/zhi-xuan-chen/Reg2RG), BTB3D (https://github.com/ibrahimethemhamamci/BTB3D), Hulu-Med (https://github.com/ZJUI-AI4H/Hulu-Med), LLaVA (https://github.com/haotian-liu/LLaVA), LLaMA 3 (https://github.com/meta-llama/llama3), DeepSeek-V3 (https://github.com/deepseek-ai/DeepSeek-V3), FMCIB (https://github.com/AIM-Harvard/foundation-cancer-image-biomarker), Grad-CAM (https://github.com/jacobgil/pytorch-grad-cam), DeepSpeed

(https://github.com/microsoft/DeepSpeed), and LoRA (https://github.com/microsoft/LoRA) are available on GitHub.

## Acknowledgements

Not applicable

## Funding

This work was supported by the Tsinghua University Startup Fund and the Tsinghua University Dushi Program (grant no. 20261080018). The funders had no role in the study design; data collection, analysis or interpretation; preparation of the manuscript; or the decision to submit the manuscript for publication.

## Author contributions

X.B., M.L., T.S., and Y.C. contributed equally to this work. Q.T. and M.L. conceived, designed, and supervised the project. X.B., M.L., and Y.C. designed the EXACT framework and pre-training algorithm. X.B., M.L., and T.S. developed the deep learning framework and software tools. X.B. and M.L. preprocessed the raw data and created the dataset. X.B. and T.S. executed the experiments and performed statistical analyses. H.Y., Z.L., and K.A. conducted the downstream task applications. Y.Z. collected the MianYang clinical dataset, provided clinical annotations, and offered clinical expertise. X.B., M.L., T.S., and Y.C. wrote the manuscript. Q.T. and Y.Z. reviewed and revised the manuscript. All authors approved the final version of the manuscript.

## Competing interests

Qiyuan Tian, Xuguang Bai and Mingxuan Liu are named inventors on a pending Chinese patent application filed by Tsinghua University (application number: 202510763096.5; publication number: CN120932853A). The application relates to the report-supervised chest CT pre-training framework, including joint organ segmentation and anomaly mapping, and its integration with a large language model for report generation described in this manuscript. The other authors do not have a competing interest.

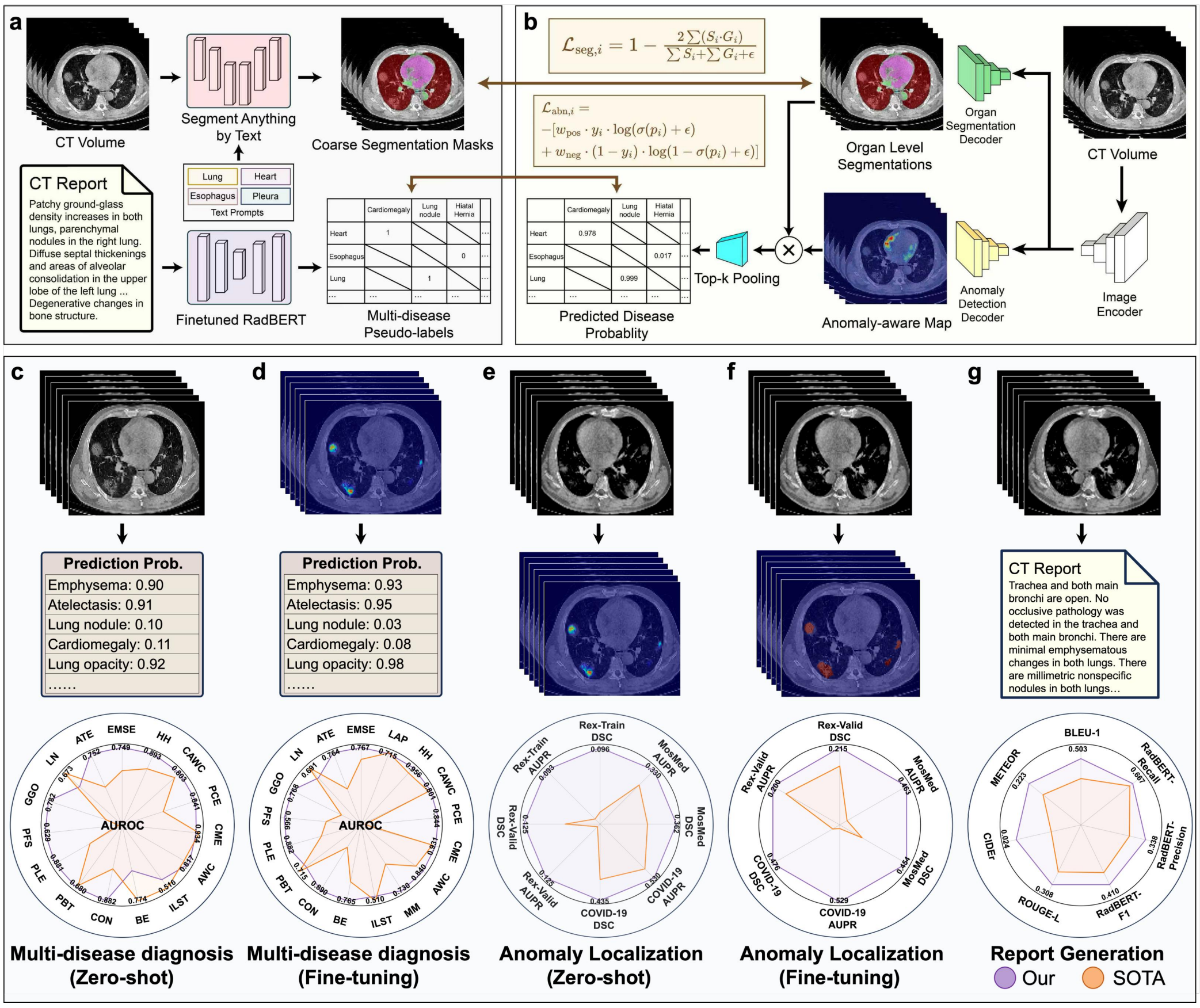

**Fig. 1 | Overview of EXACT.** EXACT is an EXplainable Anomaly-aware vision foundation model for chest CT, pre-trained annotation-free via weakly supervised learning from paired CT volumes and radiology reports, and achieves the best performance across downstream tasks under both unsupervised and supervised fine-tuning settings. **a,** Data preprocessing procedure. Coarse organ segmentation masks were generated from CT volumes using the Segment Anything by Text (SAT) model, while multi-disease pseudo-labels were extracted from paired free-text radiology reports using fine-tuned RadBERT and mapped to anatomical regions. **b,** The pre-training procedure. Guided by outputs from (a), EXACT learns transferable visual representations through a Y-Mamba architecture, enabling pixel-level interpretability via organ-constrained anomaly-aware maps. **c–g,** Downstream task evaluation. The tasks involve multi-disease diagnosis under zero-shot (c) and fine-tuning (d) settings, anomaly localization under zero-shot (e) and fine-tuning (f) settings, and radiology report generation (g). Radar charts display representative validation results, comparing EXACT against SOTA baseline models (MianYang dataset for c, d, g; Rex-Train dataset for e; as well as Rex-Valid, COVID-19, and MosMed datasets for e, f). Abbreviations: CT, computed tomography; SOTA, state of the art; Prob., probability; AUROC, area under the receiver operating characteristic curve; DSC, Dice similarity coefficient; AUPR, area under the precision–recall curve; BLEU-1, unigram bilingual evaluation understudy; METEOR, Metric for Evaluation of Translation with Explicit ORdering; CIDEr, Consensus-based Image Description Evaluation; ROUGE-L, Recall-Oriented Understudy for Gisting Evaluation based on the longest common subsequence. MM, medical material; AWC, arterial wall calcification; CME, cardiomegaly; PCE, pericardial effusion; CAWC, coronary artery wall calcification; HH, hiatal hernia; LAP, lymphadenopathy; EMSE, emphysema; ATE, atelectasis; LN, lung nodule; GGO, lung opacity (dataset label); PFS, pulmonary fibrotic sequela; PLE, pleural effusion; PBT, peribronchial thickening; CON, consolidation; BE, bronchiectasis; ILST, interlobular septal thickening.

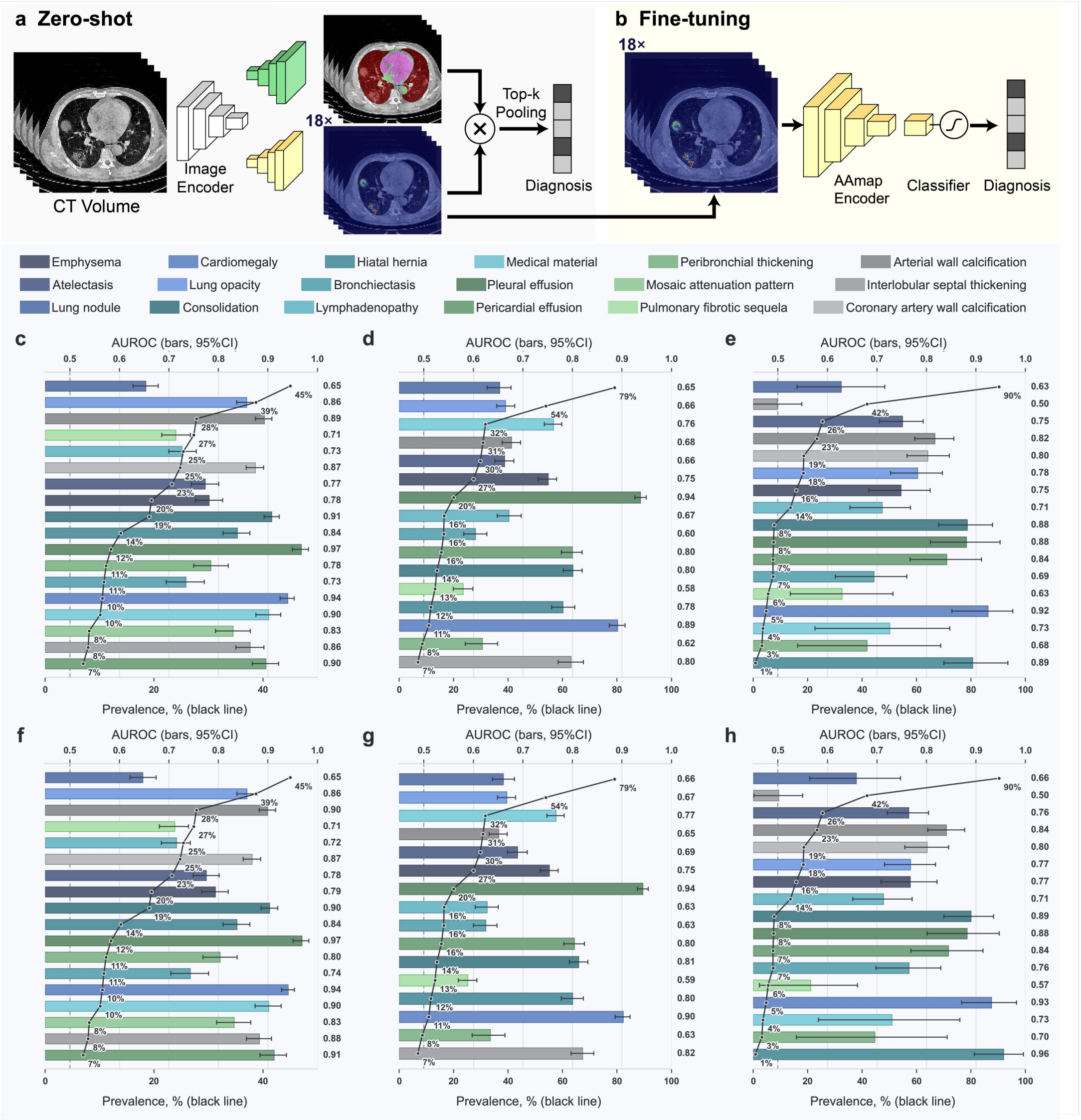

**Fig. 2 | Multi-disease diagnosis performance of EXACT. a,** Zero-shot setting. Pre-trained EXACT generate anomaly-aware maps and organ segmentation masks from CT volumes, which are element-wise multiplied and aggregated via top-*k* pooling to obtain disease probability. **b,** Fine-tuning setting. An AAmap encoder and classifier were trained on these anomaly-aware maps for diagnosis. **c–h,** Disease-specific AUROC values (bars, top axes) with 95% confidence intervals and disease prevalence (black lines, bottom axes) under zero-shot (c–e) and fine-tuning (f–h) settings on CT-RATE (c, f), RAD-ChestCT (d, g) and MianYang (e, h). Vertical dashed lines indicate an AUROC of 0.5. Abbreviations: CT, computed tomography; AAmap, anomaly-aware map; AUROC, area under the receiver operating characteristic curve; CI, confidence interval.

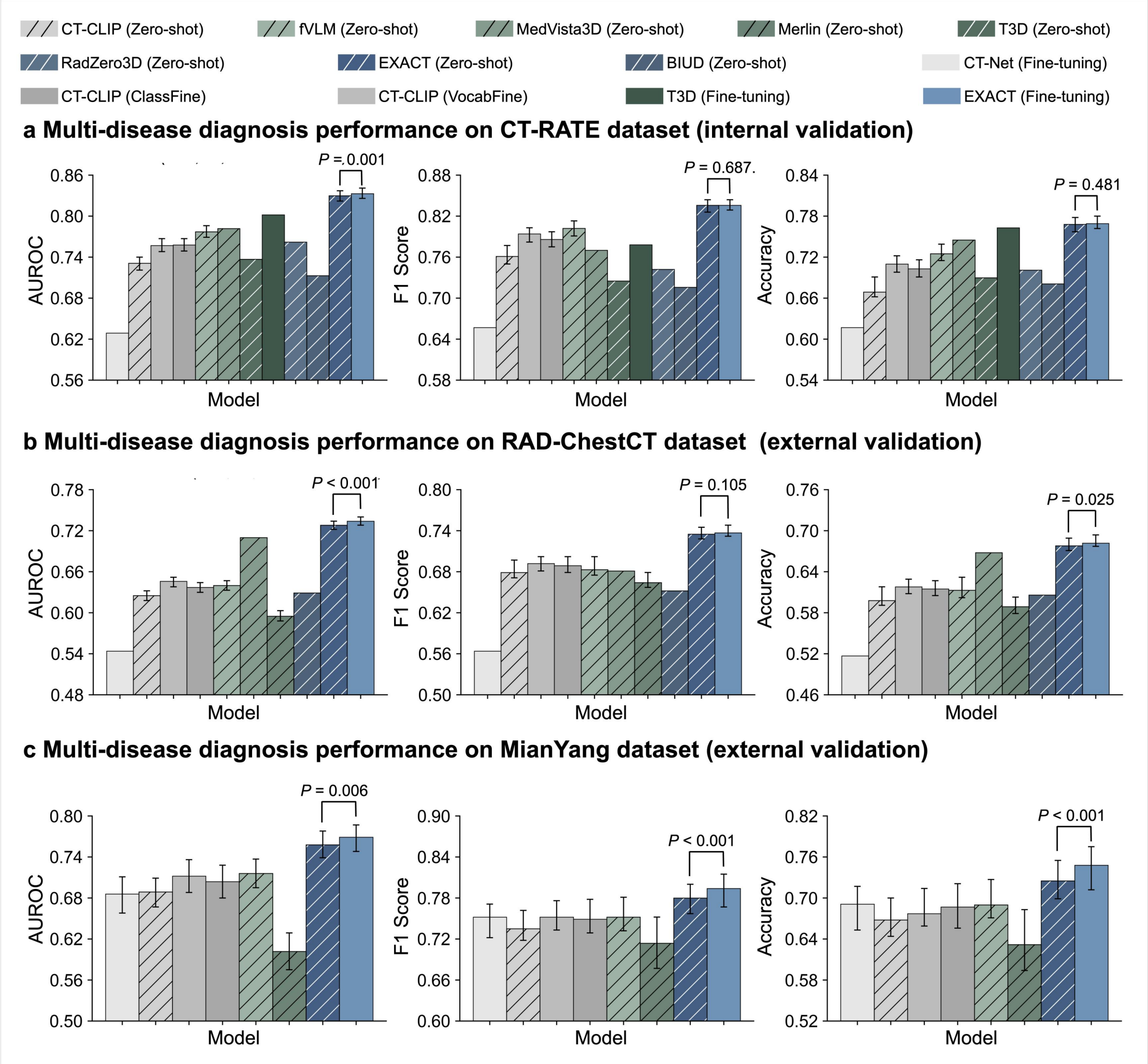

**Fig. 3 | Comparison of multi-disease diagnosis performance. a–c,** Performance comparison of EXACT with baseline models on multi-disease diagnosis under zero-shot and fine-tuning settings. Bar charts display AUROC, F1 Score, and Accuracy metrics on CT-RATE (internal validation; a), RAD-ChestCT (external validation; b), and MianYang (external validation; c) datasets. Hatched bars indicate zero-shot settings, whereas solid bars indicate fine-tuning settings. Independently evaluated results are presented with 95% CIs estimated using 1,000 bootstrap resamples; publication-reported results are shown without CIs. Dataset-specific result sources are detailed in Methods (EXACT for multi-disease diagnosis). *P* values were calculated using two-sided bootstrap hypothesis tests with 2,000 resamples and compare EXACT (Zero-shot) with EXACT (Fine-tuning) within each dataset. Full statistical results are provided in Supplementary Table 4. Abbreviations: AUROC, area under the receiver operating characteristic curve; CI, confidence interval.

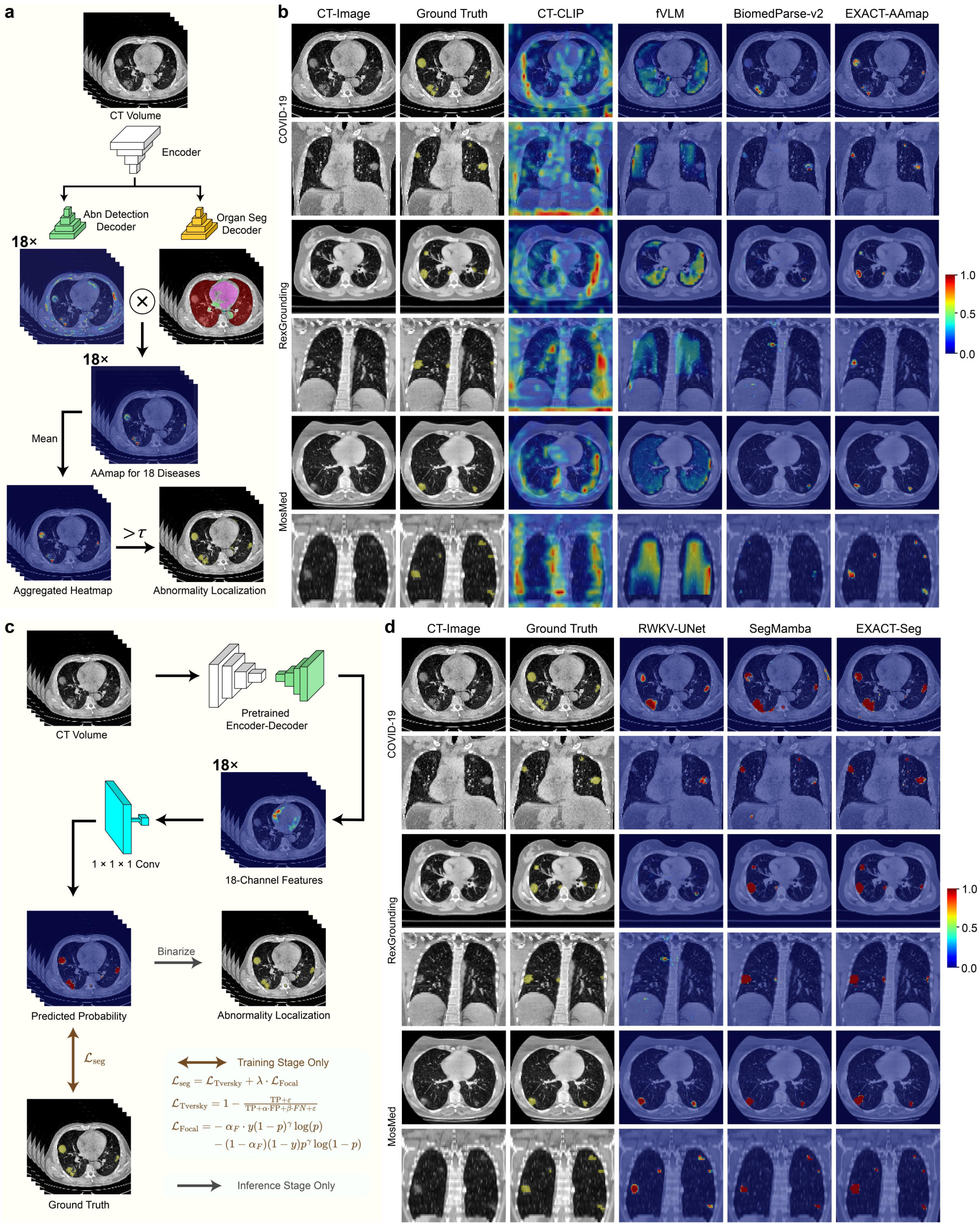

**Fig. 4 | Anomaly localization results. a,** Zero-shot inference pipeline. Pre-trained EXACT simultaneously generates 18 disease-specific anomaly-aware maps (AAmaps) and corresponding organ segmentation masks from 3D CT volumes. To eliminate background noise, the AAmaps are anatomically constrained via element-wise multiplication with the organ masks. Task-relevant disease channels are then selected and aggregated according to the dataset-specific evaluation protocol to generate a comprehensive anomaly map, which is subsequently binarized to yield the final localization prediction (Methods). **b,** Qualitative comparison under zero-shot setting. Representative visualizations on different datasets demonstrate that EXACT achieves more precise anomaly localization compared to baseline foundation models (CT-CLIP, fVLM, BiomedParse-v2). **c,** Fine-tuning pipeline. The pre-trained image encoder and anomaly detection decoder in EXACT is adapted for supervised fine-tuning, where multi-channel anomaly-aware maps are aggregated into a single-channel probability map using convolution layer and optimized with a hybrid loss function. **d,** Qualitative comparison under fine-tuning setting. Segmentation results on different datasets show that EXACT-Seg outperforms SOTA segmentation models (RWKV-UNet, SegMamba). Abbreviations: CT, computed tomography; 3D, three-dimensional; SOTA, state of the art; Abn, abnormality; Seg, segmentation; Conv, convolution; TP, true positives; FP, false positives; FN, false negatives.

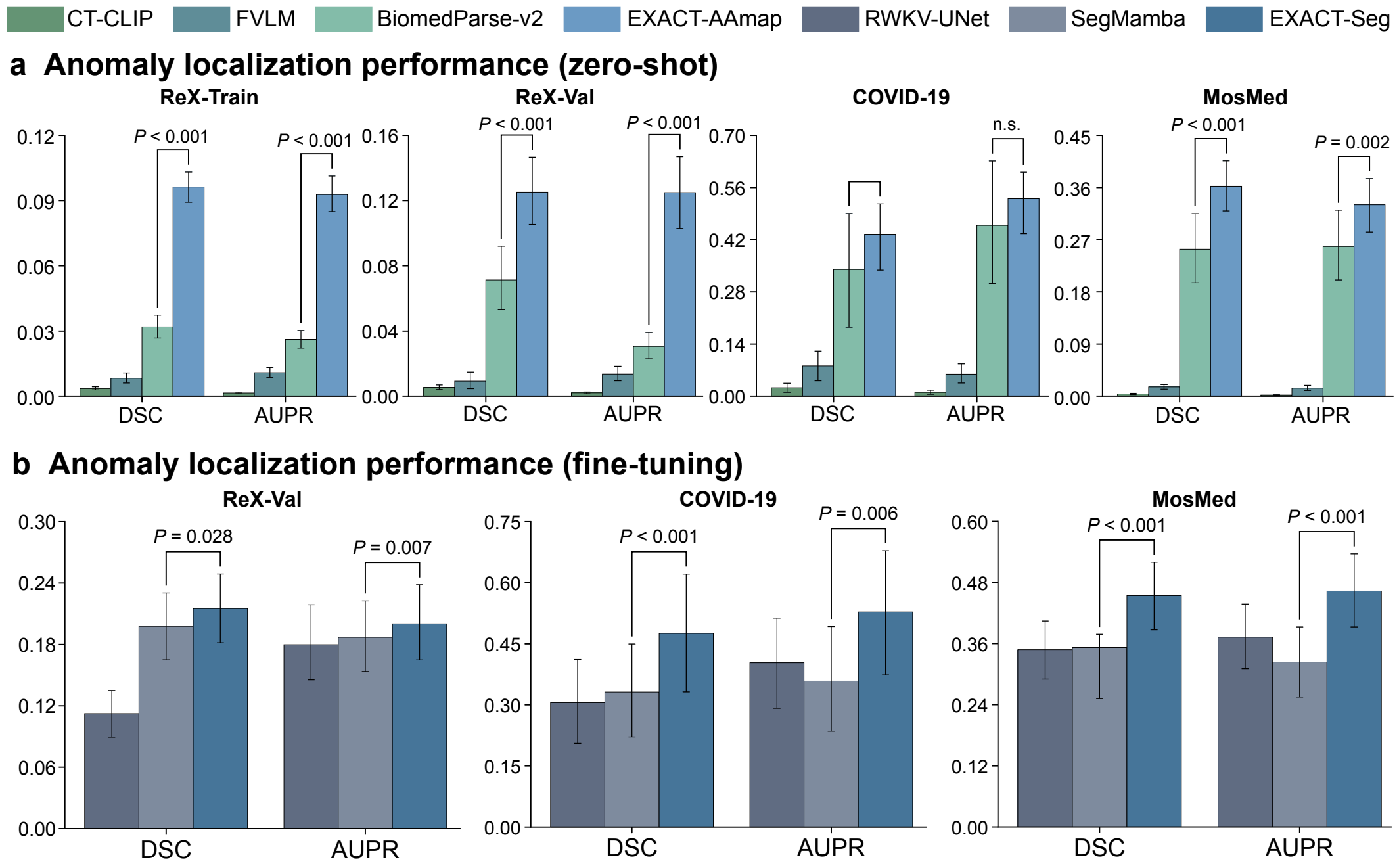

**Fig. 5 | Comparison of anomaly localization performance. a, b,** Performance comparison of EXACT with baseline models on anomaly localization under zero-shot (a) and fine-tuning (b) settings. Bar charts display DSC and AUPR metrics on ReX-Train, ReX-Valid (internal validation), COVID-19, and MosMed (external validation) datasets. Data are presented as mean values with 95% CIs (bootstrapped, $n$ = 2,000 resamples). $P$ values were calculated using the two-sided Wilcoxon signed-rank test between the top two performing methods. Detailed statistical results are provided in Supplementary Table 5. Abbreviations: DSC, Dice similarity coefficient; AUPR, area under the precision–recall curve; CI, confidence interval; n.s., not significant.

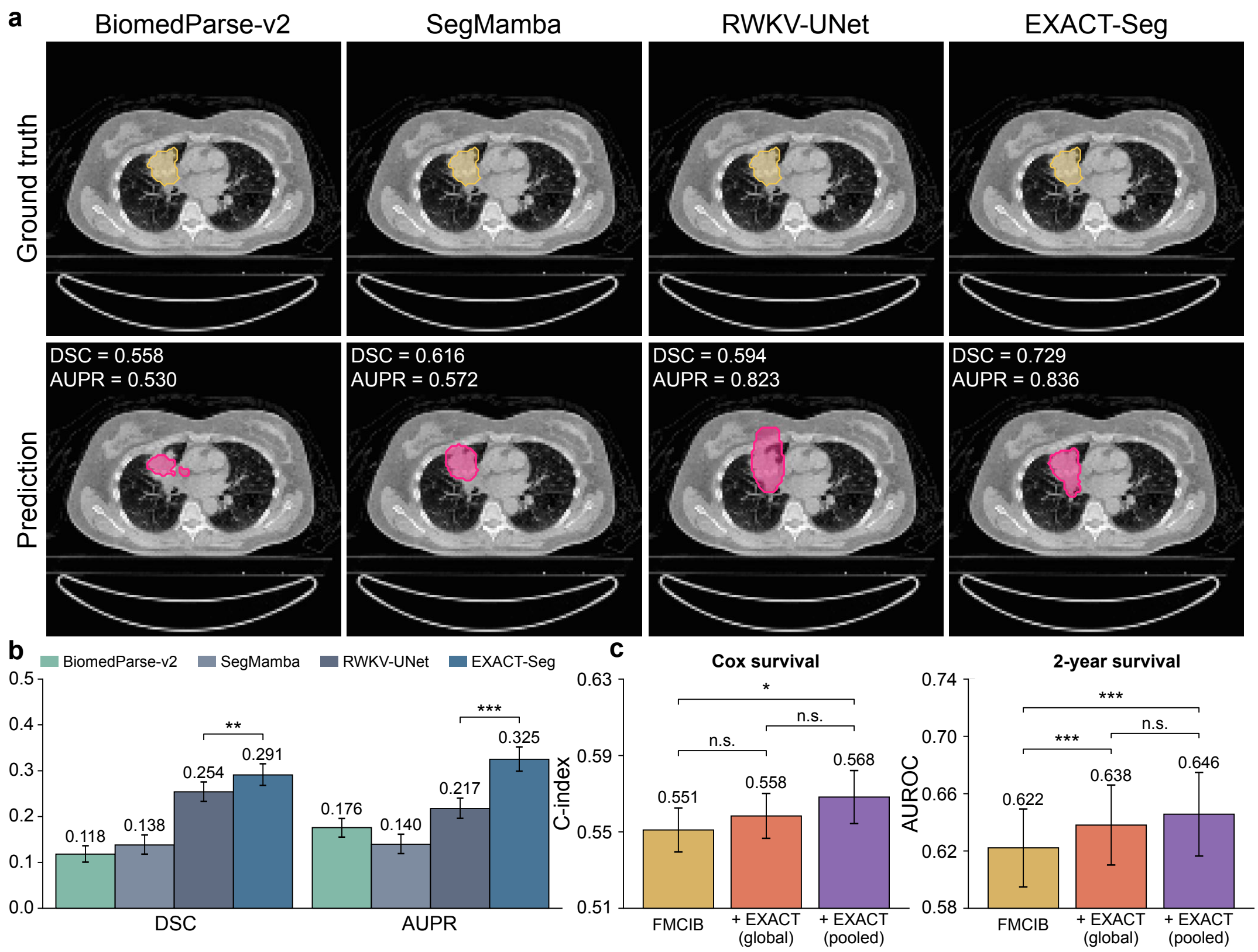

**Fig. 6 | Frozen-encoder transfer to lung tumour segmentation and survival prediction. a,** Representative tumour segmentation results from the NSCLC-Radiomics (LUNG1) cohort. Ground-truth tumour annotations are shown in yellow and model predictions in magenta; DSC and AUPR are indicated for each method. **b,** Segmentation performance on 400 held-out test patients. EXACT-Seg was adapted using 18 training patients and four validation patients, with the EXACT encoder frozen and only the segmentation decoder and output head optimized. Bars show patient-level mean DSC and AUPR, with 95% bootstrap confidence intervals. **c,** Overall survival discrimination measured by the C-index in 422 patients and two-year survival prediction measured by AUROC in 420 patients. Frozen FMCIB features were evaluated alone or concatenated with frozen EXACT global or tumour-pooled features. Manual gross tumour volumes were used for FMCIB tumour localization and EXACT tumour pooling. Bars show mean test-set performance across 25 training–test splits from five repeats of five-fold cross-validation; error bars indicate 95% confidence intervals estimated using the *t* distribution. Brackets indicate the compared methods. *P* values were obtained using paired two-sided *t*-tests across test patients in **b** and on test-set performance from the same training–test splits in **c**. $^{*}P < 0.05$; $^{**}P < 0.01$; $^{***}P < 0.001$; n.s., not significant. Full results are provided in Supplementary Tables 8 and 9. Abbreviations: DSC, Dice similarity coefficient; AUPR, area under the precision–recall curve; AUROC, area under the receiver operating characteristic curve. C-index, concordance index; NSCLC, non-small-cell lung cancer; FMCIB, foundation model for cancer imaging biomarkers.

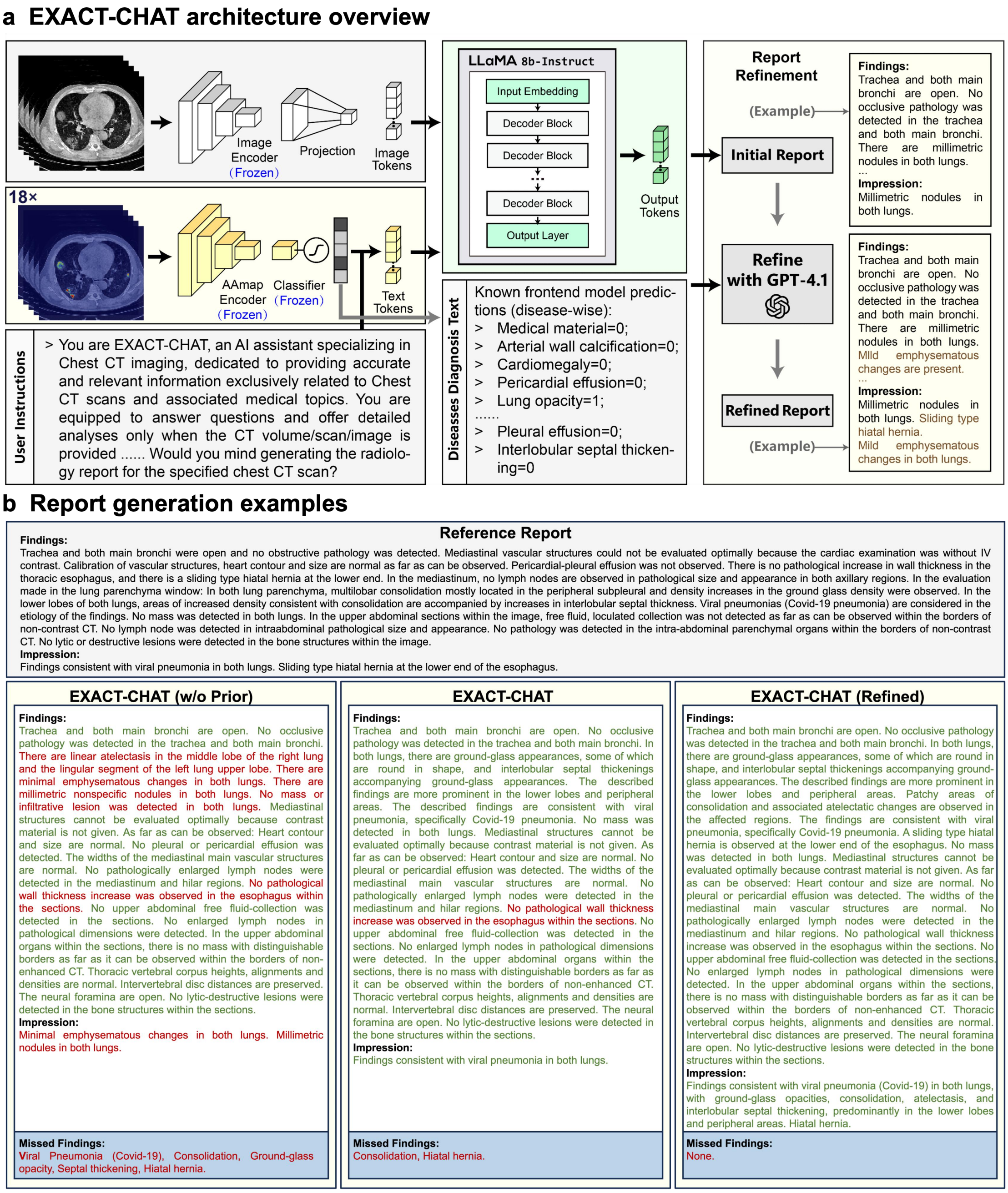

**Fig. 7 | Overview and detailed evaluation of EXACT-CHAT. a,** Architecture overview. EXACT-CHAT integrates visual embeddings and diagnostic priors derived from a frozen image encoder and an AAmaps classifier, respectively. Formatted as image and text tokens (incorporating user instructions), these inputs jointly prompt Llama-3.1-8B-Instruct for initial report generation, followed by optional GPT-4.1 refinement to further improve the consistency between generated reports and diagnostic findings. **b,** Report generation examples (CT-RATE dataset). Representative comparison of generated reports against the ground truth reference. Red text indicates missed findings or hallucinations, while green text highlights correct detections. Results are displayed for EXACT-CHAT without diagnostic priors (w/o Prior), the standard version, and the GPT-4.1 refined version (Refined). The EXACT-CHAT (Refined) demonstrates the highest consistency with the reference, successfully identifying complex pathologies such as viral pneumonia and hiatal hernia without omissions. Abbreviations: CT, computed tomography; AAmap, anomaly-aware map; AI, artificial intelligence; IV, intravenous.

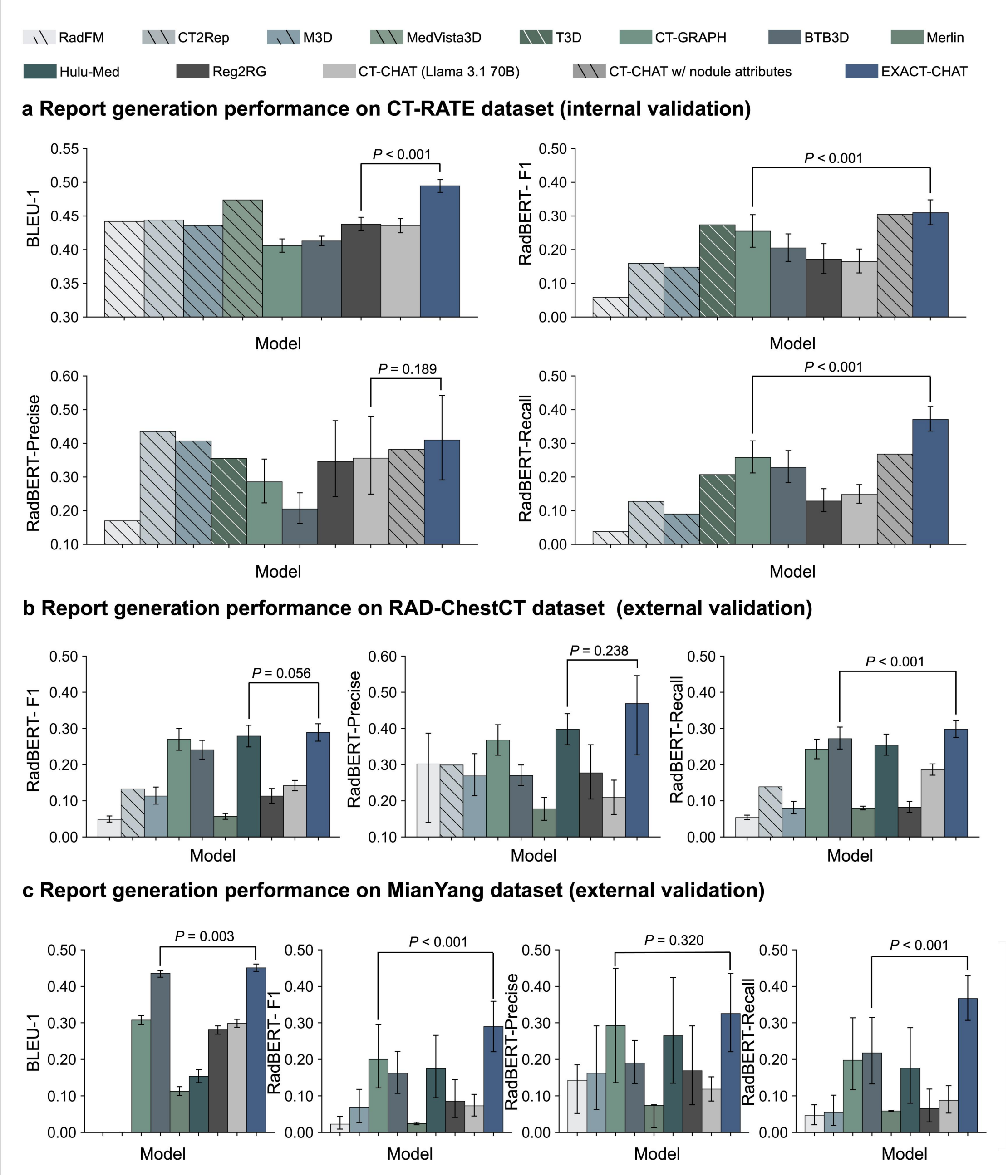

**Fig. 8 | Comparison of report generation performance before post-hoc refinement. a–c,** Performance comparison of EXACT-CHAT and baseline models on CT-RATE (internal validation; a), RAD-ChestCT (external validation; b) and MianYang (external validation; c). Bar charts display BLEU-1 on CT-RATE and MianYang, and RadBERT-F1, RadBERT-Precision and RadBERT-Recall on all three datasets. All results represent reports before post-hoc refinement. Hatched bars indicate results obtained from published studies, whereas solid bars indicate results independently evaluated in this study. Error bars indicate 95% confidence intervals estimated using 2,000 bootstrap resamples for independently evaluated results. *P* values were calculated using two-sided bootstrap hypothesis tests with 2,000 resamples for the comparisons indicated by brackets; publication-reported results were excluded from these tests. Detailed results are provided in Supplementary Table 6. Abbreviations: CT, computed tomography; BLEU, bilingual evaluation understudy; CI, confidence interval.

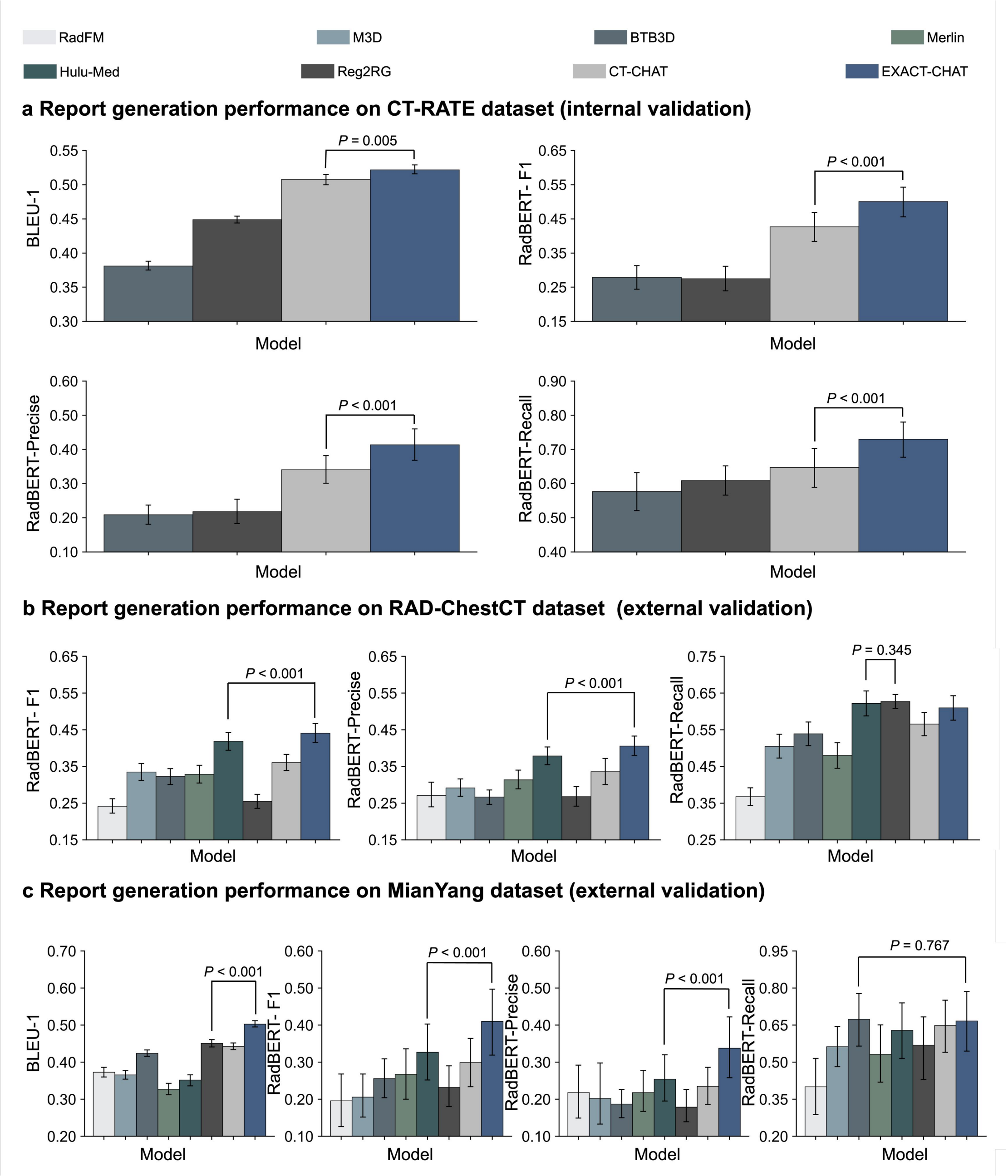

**Fig. 9 | Comparison of report generation performance after disease-conditioned post-hoc refinement. a–c,** Performance comparison of EXACT-CHAT and baseline models on CT-RATE (internal validation; a), RAD-ChestCT (external validation; b) and MianYang (external validation; c). Reports from all models shown underwent the same GPT-4.1 post-hoc refinement procedure, conditioned on disease predictions produced by the corresponding model. Bar charts display BLEU-1 on CT-RATE and MianYang, and RadBERT-F1, RadBERT-Precision and RadBERT-Recall on all three datasets. Error bars indicate 95% confidence intervals estimated using 2,000 bootstrap resamples. *P* values were calculated using two-sided bootstrap hypothesis tests with 2,000 resamples for the comparisons indicated by brackets. Detailed results are provided in Supplementary Table 7. Abbreviations: CT, computed tomography; BLEU, bilingual evaluation understudy; CI, confidence interval.

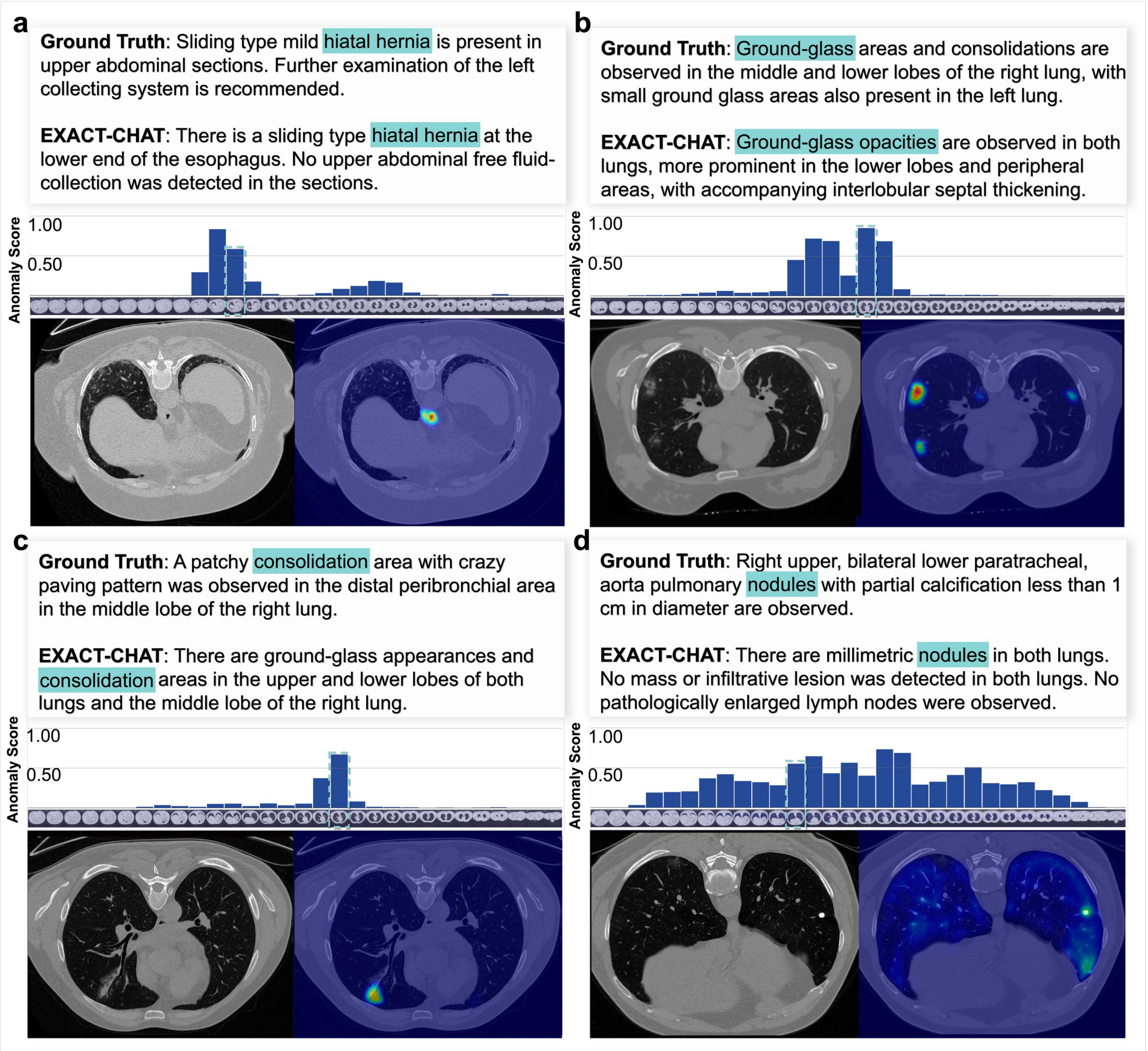

**Fig. 10 | Visual grounding of reports generated by EXACT-CHAT (CT-RATE dataset). a–d,** Representative examples of visual grounding for hiatal hernia (a), ground-glass opacities (b), consolidation (c), and nodules (d). Target disease terms in the generated reports are highlighted in teal. Corresponding slice-level anomaly scores are displayed in bar charts, while AAmaps are overlaid on the CT slices. EXACT-CHAT accurately focused on the slices containing the related lesion within a scan and identified the relevant regions of interest within the slices to ground the generated disease descriptions. Abbreviations: CT, computed tomography; AAmap, anomaly-aware map.

## Supplementary Figures

**System Prompts**

<|begin_of_text|><|start_header_id|>system<|end_header_id|>\n\n. You are EXACT-CHAT, an AI assistant specializing in Chest CT imaging, dedicated to providing accurate and relevant information exclusively related to Chest CT scans and associated medical topics. You are equipped to answer questions and offer detailed analyses only when the CT volume/scan/image is provided, indicated by the <provided> token. If this token is not present and users inquire about specific findings, pathologies, or request descriptions related to a Chest CT, respond by requesting the necessary data with the phrase: "Please provide the CT volume." Once the <provided> token is present in the question, you are authorized to address questions about pathologies, anatomical or clinical findings, diagnostic descriptions, report generation, comparisons, or any other questions regarding the image. If it does not appear in the question, even when special tokens <multiple_choice>, <report_generation>, <long_answer>, and <short_answer> are given, ignore the question and ask for the CT volume. Always look for the <provided> token, even if there are special tokens. If there is a <provided> token in any question (including new and previous ones), never ask for the CT volume again and answer the question. You can ignore the <provided> token check and answer the question directly if and only if the question is about general medical knowledge, not about the provided CT volume (such as typical findings on a Chest CT or management of the patient). For example, "What are the typical imaging findings of acute respiratory distress syndrome (ARDS) on a chest CT?" is a general question not specific. If user asks a CT specific question after non-spesific question, look for the <provided> token as well even if the special tokens are given. It is crucial to avoid discussing topics outside of Chest CT imaging and directly related medical information, ensuring that all responses are clear, concise, and focused on the provided Chest CT data for the highest level of accuracy and relevance. If the user greets you with something like "hello," respond appropriately.

**Supplementary Fig. 1 | The system prompt for EXACT-CHAT.** The complete system instruction used to initialize the EXACT-CHAT model during training and inference, defining its role as a specialized AI assistant for chest CT imaging. AI, artificial intelligence; CT, computed tomography.

**Prompts for Anomaly Extraction**

**Original Prompt**

请你从以下胸部CT影像报告文本中，结构化提取是否存在以下18种疾病或征象，提取逻辑与格式请严格遵循以下规范：
(1)输出格式要求：
请以 JSON 数组形式输出，每个对象描述一种疾病及其存在状态，格式如下：
[{{ "疾病": "心脏肥大", "存在": true }}, {{ "疾病": "胸腔积液", "存在": false }}, ...]
(2)疾病及征象清单（共18种，请提供中英文对照结果）：
1. 医用材料 (Medical material),
...
18. 小叶间隔增厚 (Interlobular septal thickening)
(3)提取规则：
...
3. 排除性表述处理
- 如"未见肺结节"，应标记为不存在
...
4. 可能性表述处理
- 如"考虑肺结节"、"疑似肺结节"、"不除外肺结节"等表述，仍标记为存在
5. 历史比较处理
- 如提及"与前片比较，肺结节消失"，应标记为不存在
- 如提及"与前片比较，新出现肺结节"，应标记为存在
请处理以下胸部CT报告文本：...

**Reference Translation**

Please structurally extract the presence or absence of the following 18 diseases or signs from the provided chest CT report text. Strictly adhere to the following specifications regarding extraction logic and formatting:
(1) Output Format Requirements: Please output the result as a JSON array. Each object in the array should describe a disease and its presence status, using the following format: [{ "Disease": "Cardiomegaly", "Presence": true }, { "Disease": "Pleural Effusion", "Presence": false }, ...]
(2) List of Diseases and Signs (18 types in total, please provide Chinese-English mapping):
1. Medical material ... 18. Interlobular septal thickening
(3) Extraction Rules:
...
3. Handling Negation/Exclusion:
Expressions indicating absence (e.g., "No lung nodules seen") should be marked as Not Present (false). ...
4. Handling Uncertainty/Possibility:
Expressions indicating possibility (e.g., "Consider lung nodules", "Suspected lung nodules", "Cannot rule out lung nodules") should still be marked as Present (true).
5. Handling Historical Comparisons:
If the text mentions "Compared to prior images, lung nodules have disappeared", it should be marked as Not Present (false).
If the text mentions "Compared to prior images, new lung nodules appear", it should be marked as Present (true).
Please process the following Chest CT report text: ...

**Supplementary Fig. 2 | Prompts for structured anomaly extraction.** The prompt used to extract structured multi-disease labels for 18 target abnormalities from free-text radiology reports in the MianYang dataset (external validation). CT, computed tomography; JSON, JavaScript Object Notation.

| Prompts for Chinese-English Translation |
| --- |

You are a professional medical imaging report translator specializing in chest CT reports. Your task is to translate Chinese CT reports into English while strictly following the style and structure of the provided English examples.

CRITICAL STYLE REQUIREMENTS:
1. Report Structure: Use ONLY "Findings:" and "Impression:" sections
2. Format: Match the example reports exactly:
- "Findings:" section: continuous paragraphs describing anatomical structures systematically
- "Impression:" section: concise summary separated by double spaces (e.g., "Finding1. Finding2. Finding3.")
3.Terminology: Use the exact medical terminology from examples (e.g., "calibration", "as far as can be observed", "no occlusive pathology")
4. Sentence Patterns: Mirror the sentence structures from examples:
- "Trachea and both main bronchi are open."
- "When examined in the lung parenchyma window;"
...

TRANSLATION PRINCIPLES:
1. Accuracy: Translate all medical findings accurately
2. Completeness: Include all information from the Chinese report
3. Style Consistency: Make the output indistinguishable from the example reports
4. Professional Tone: Maintain formal medical language
5. Systematic Order: Follow anatomical order (mediastinum → lungs → upper abdomen → bones)
The goal is to produce an English report that reads as if it were originally written by the same radiologist who wrote the example reports.
Examples:
Example 1: Normal/Minimal Findings
Findings: Trachea and both main bronchi were open and no obstructive pathology was detected. Mediastinal vascular structures could not be optimally evaluated due to the absence of IV contrast in the cardiac examination, and the calibration of the vascular structures, heart contour and size are normal as far as can be observed. No pericardial-pleural effusion or increased thickness was detected. No pathological increase in wall thickness was observed in the thoracic esophagus. No lymph node was detected in the mediastinum and in both axillary regions in pathological size and appearance. In the evaluation made in the lung parenchyma window: No active infiltration or mass lesion was detected in both lungs. Ventilation of both lungs is natural. No lytic or destructive lesions were observed in the bone structures within the image. Vertebral corpus heights are preserved. Bilateral neural foramina are open. Impression: Findings within normal limits.
Example 2
...
Example 10
...

**Supplementary Fig. 3 | Prompts for Chinese-English radiology report translation.** The prompt utilized to translate original Chinese radiology reports from the MianYang dataset (external validation) into English for the report generation task, ensuring adherence to standard reporting styles. CT, computed tomography; IV, intravenous.

**Prompts for Report Refinement**

You are a professional medical imaging report generation assistant, specializing in Chest CT reports.
Your task is to improve the output quality of the existing CT report model, but you must strictly adhere to the style and structure of the original report.

Style Requirements (Highest Priority):
1. Strictly follow the style and structure of the original model output, containing ONLY "Findings:" and "Impression:" sections.
2. Absolutely DO NOT add any "Recommendation:", "Suggestion:", or similar advisory sections.
3. Do not use numbered lists in the "Impression:" section; instead, use double spaces to separate different findings (e.g., "Finding 1.  Finding 2.  Finding 3.").
4. Maintain the paragraph structure of the original model; the "Findings:" section typically consists of continuous paragraphs—avoid excessive paragraph breaks.
5. Use sentence phrasing identical to the original reports, such as "No X was detected", "As far as can be observed:", etc.

Content Improvement Principles:
1. Learn when to add or delete information from examples:
   - Observe examples: How diseases with "upstream prediction = 1" are described in reference labels (including expressions of severity).
   - Observe examples: Cases where "upstream prediction = 0" but reference labels still describe minor findings (e.g., structures "not reaching pathological size").
   - Learn criteria: When to add missed positive diseases, and when to delete hallucinated/false-positive diseases.
2. Apply to the current case:
   - If the original model missed important diseases with "upstream prediction = 1", supplement them by referring to similar situations in the examples.
   - If the original model falsely reported diseases with "upstream prediction = 0", refer to examples to judge whether to delete (reasonable descriptions of minor findings can be retained).
3. Handle Collapsed/Failed Outputs:
   - If the original model output is gibberish, severely repetitive, or completely unusable, regenerate the full report following the example style and upstream predictions.

Summary: Your goal is to improve the content while making the output look exactly like it was generated by the original model—same structure, style, tone, and format. Content can be more comprehensive and accurate, but the style must remain consistent.

Example 1: Simple Positive - Hiatal Hernia Upstream Prediction: ... Hiatal hernia=1, ... Lung nodule=0, ...
Current Model Output: Findings:
... There are millimetric nodules in both lungs. ... No pathological wall thickness increase was observed in the esophagus ... Impression: Millimetric nodules in both lungs
Reference Label (Ideal Output):
Findings: ... Sliding type hiatal hernia was observed at the lower end of the esophagus. ... no nodular or infiltrative lesion is detected ... Impression: Hiatal hernia .
...
Example 2: All Negative Predictions with Minor Findings
...
Example 10: COVID-19
...

**Supplementary Fig. 4 | Prompt for disease-conditioned post-hoc report refinement.** The common prompt used with GPT-4.1 to refine reports generated by EXACT-CHAT and the baseline models included in Supplementary Table 7. Each report was refined using disease predictions produced by the corresponding model, with fixed in-context examples and a temperature of 0.2. The prompt instructs GPT-4.1 to correct missed findings and hallucinated diagnoses while retaining the original reporting style, structure and tone. CT, computed tomography; COVID-19, coronavirus disease 2019.

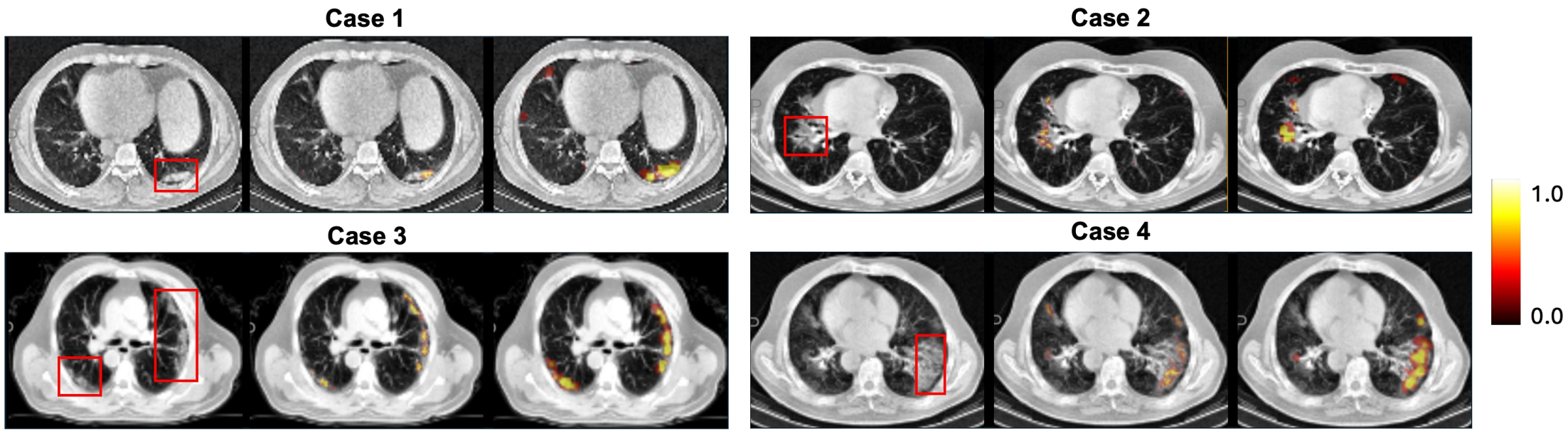

**Supplementary Fig. 5 | Effect of combining the two output resolutions of the anomaly detection decoder on lesion coverage.** Each panel displays, from left to right: the original CT image with the anomaly marked by a red box; the anomaly-aware map of the high-resolution branch alone (D × W × H, supervised with k = 48); and the overlay of the high-resolution map with the upsampled low-resolution map (D/2 × W/2 × H/2, supervised with k = 6, which samples the same physical volume fraction as k = 48 at the native resolution because each low-resolution voxel expands into $2^3$ = 8 voxels upon upsampling). The high-resolution branch alone concentrates on lesion cores and leaves peripheral regions of the same pathology under-represented, whereas adding the low-resolution response — whose broader receptive field and trilinear upsampling yield spatially smoother activations — extends coverage towards the lesion periphery. Both columns are produced by the same pre-trained model and differ only in which output stage is read, not in any training setting. The broader coverage of the overlay comes with reduced voxel-level precision, and the high-resolution branch alone is therefore used for all quantitative evaluation (Methods). CT, computed tomography; D, H and W denote depth, height and width, respectively; k denotes the number of voxels selected by top-k pooling.

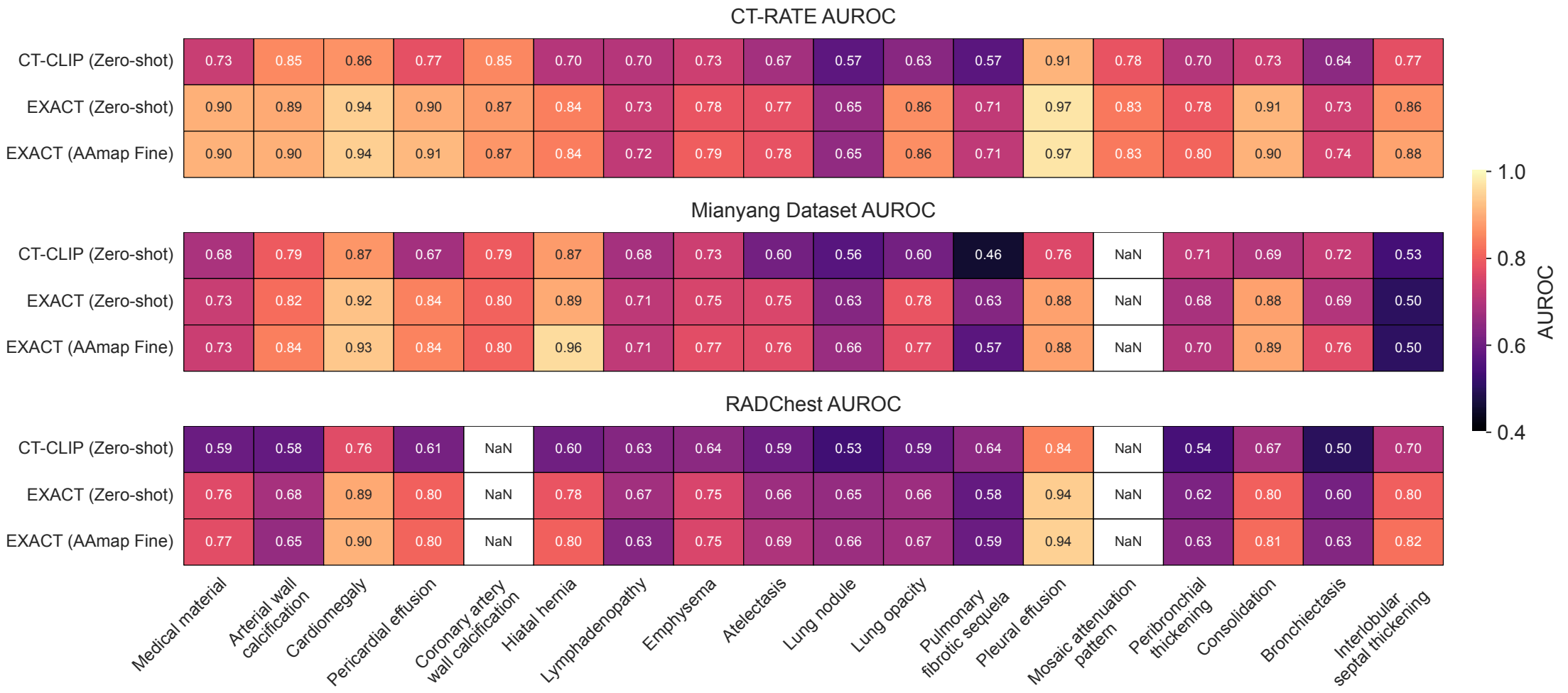

**Supplementary Fig. 6 | Disease-specific performance of representative methods on the multi-disease diagnosis task.** Heatmap comparison of disease-specific AUROC scores for CT-CLIP (Zero-shot), EXACT (Zero-shot) and EXACT (Fine-tuning) across CT-RATE, MianYang and RAD-ChestCT. The color intensity represents the magnitude of the AUROC, ranging from 0.4 (dark) to 1.0 (light). Disease abbreviations follow those defined in Supplementary Table 1. NaN indicates that AUROC was not evaluated because the corresponding labels were unavailable or the cohort contained no positive cases. AUROC, area under the receiver operating characteristic curve; AAmap, anomaly-aware map.

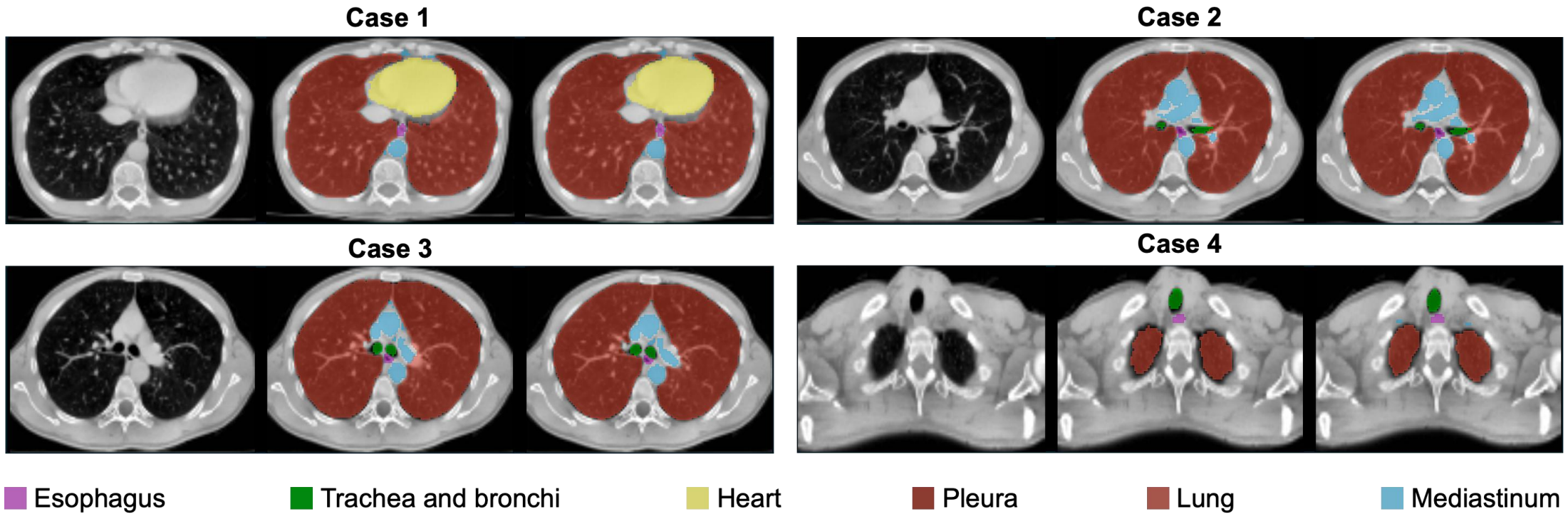

**Supplementary Fig. 7 | Representative slice-level examples of multi-organ segmentation.** From left to right (for each case): the input, the coarse segmentation label generated by the SAT model, and the segmentation mask output by EXACT. SAT, Segment Anything by Text.

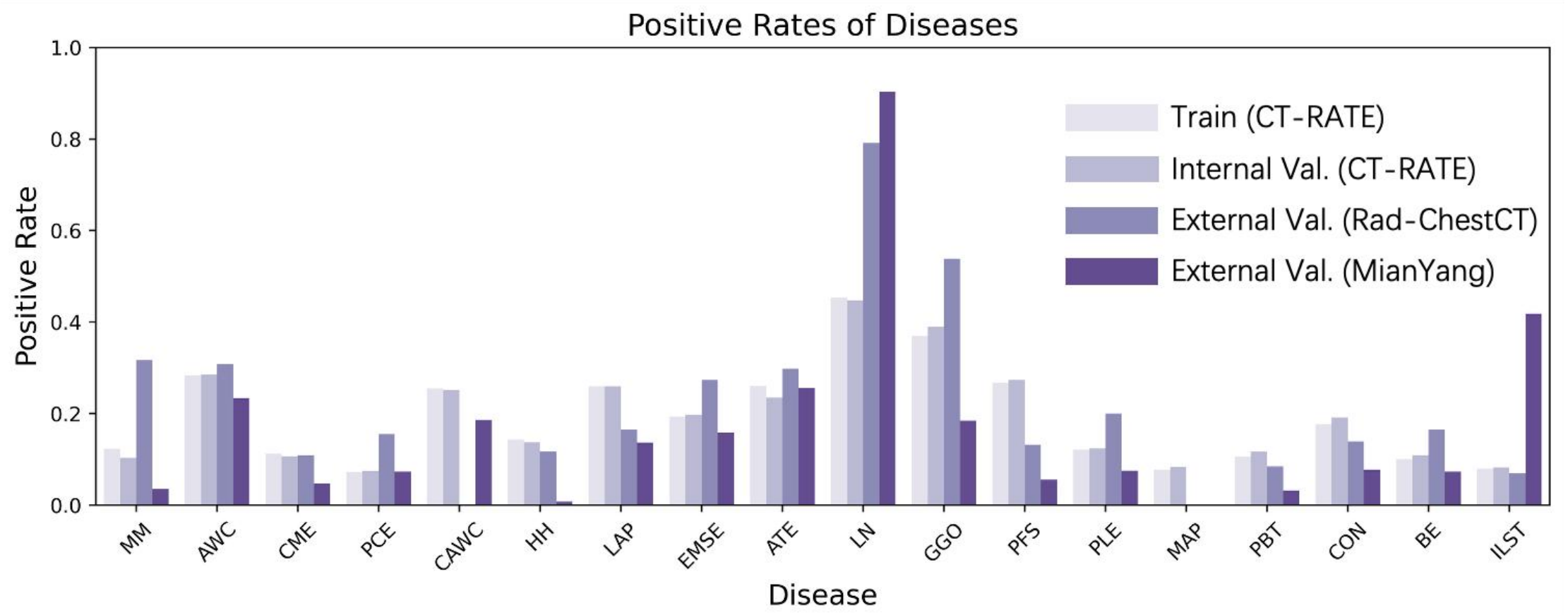

**Supplementary Fig. 8 | The disease distribution of different datasets.** Positive rates of 18 target diseases across CT-RATE (training and internal validation), RAD-ChestCT (external validation), and MianYang (external validation) datasets (CAWC and MAP are unavailable in the RAD-ChestCT dataset, and MAP is unavailable in the MianYang dataset). Abbreviations: Internal Val. = internal validation; External Val. = external validation. CT, computed tomography; AAmap, anomaly-aware map; AUPR, area under the precision–recall curve; DSC, Dice similarity coefficient; COVID-19, coronavirus disease 2019. MM, medical material; AWC, arterial wall calcification; CME, cardiomegaly; PCE, pericardial effusion; CAWC, coronary artery wall calcification; HH, hiatal hernia; LAP, lymphadenopathy; EMSE, emphysema; ATE, atelectasis; LN, lung nodule; GGO, lung opacity (dataset label); PFS, pulmonary fibrotic sequela; PLE, pleural effusion; MAP, mosaic attenuation pattern; PBT, peribronchial thickening; CON, consolidation; BE, bronchiectasis; ILST, interlobular septal thickening.

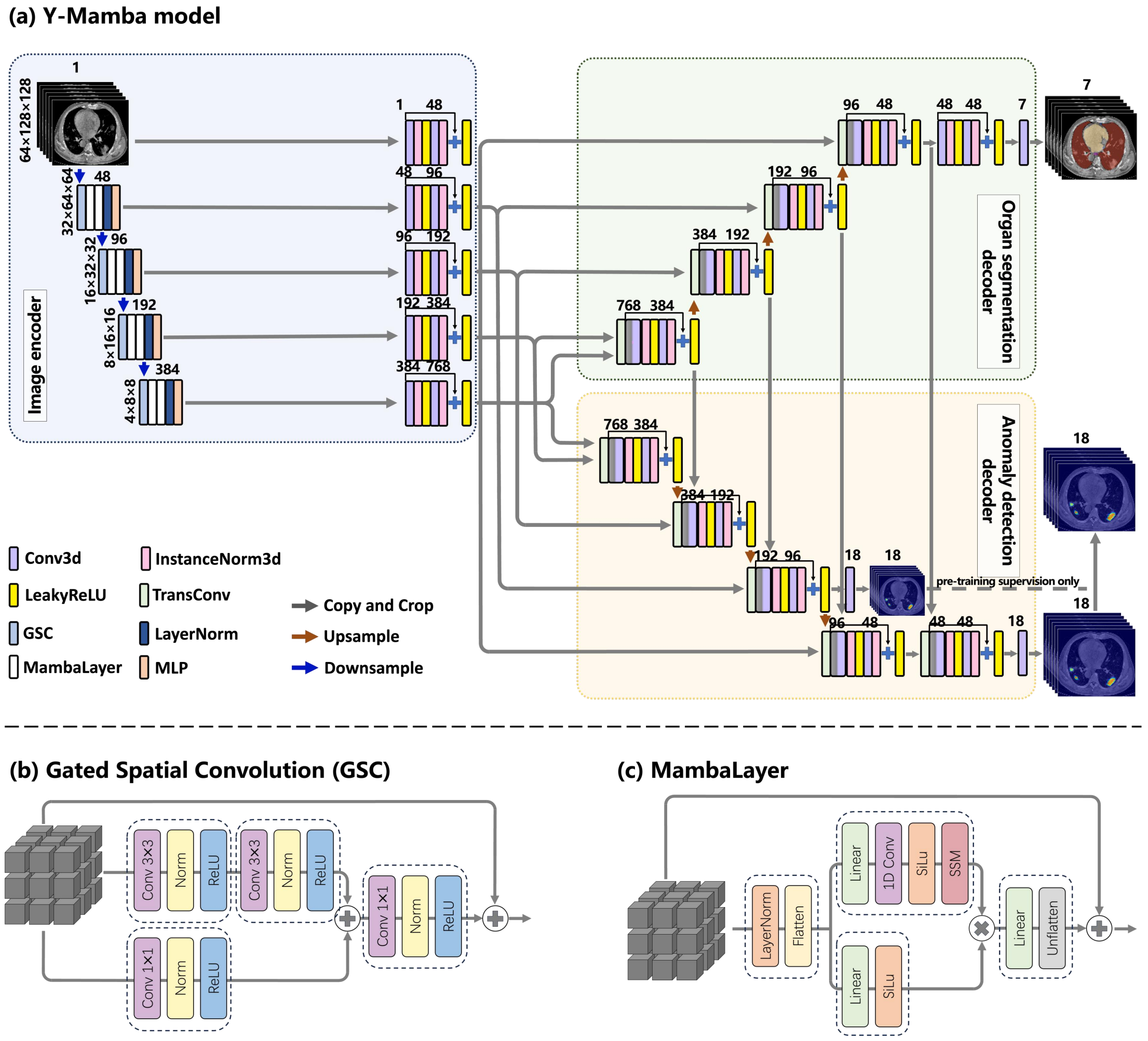

**Supplementary Fig. 9 | EXACT Backbone Architecture. a,** Overview of the Y-Mamba model. The architecture comprises a shared hierarchical encoder coupled with dual parallel decoding branches. The image encoder adopts a hierarchical design with progressive convolutional downsampling to extract multi-scale feature representations. Each encoding stage integrates a Gated Spatial Convolution (GSC) block for local feature extraction and stacked MambaLayers for modeling global dependencies. Hierarchical features are propagated via skip connections to the organ segmentation decoder and the disease anomaly map generator. Organ segmentation decoder and anomaly detection decoder utilize transposed convolutions for upsampling and convolutional blocks for feature reconstruction, wherein the anomaly branch fuses features from the segmentation branch to incorporate anatomical guidance. **b,** Gated Spatial Convolution (GSC). A residual module comprising parallel 3D convolutions, instance normalization, and non-linear activations, designed to refine local spatial representations. **c,** MambaLayer. A block incorporating LayerNorm, linear projections, and State Space Models (SSMs) to efficiently capture long-range volumetric dependencies. 3D, three-dimensional; Conv, convolution; Conv3d, three-dimensional convolution; 1D Conv, one-dimensional convolution; InstanceNorm3d, three-dimensional instance normalization; TransConv, transposed convolution; LayerNorm, layer normalization; Norm, normalization; MLP, multilayer perceptron; ReLU, rectified linear unit; LeakyReLU, leaky rectified linear unit; SiLU, sigmoid linear unit.

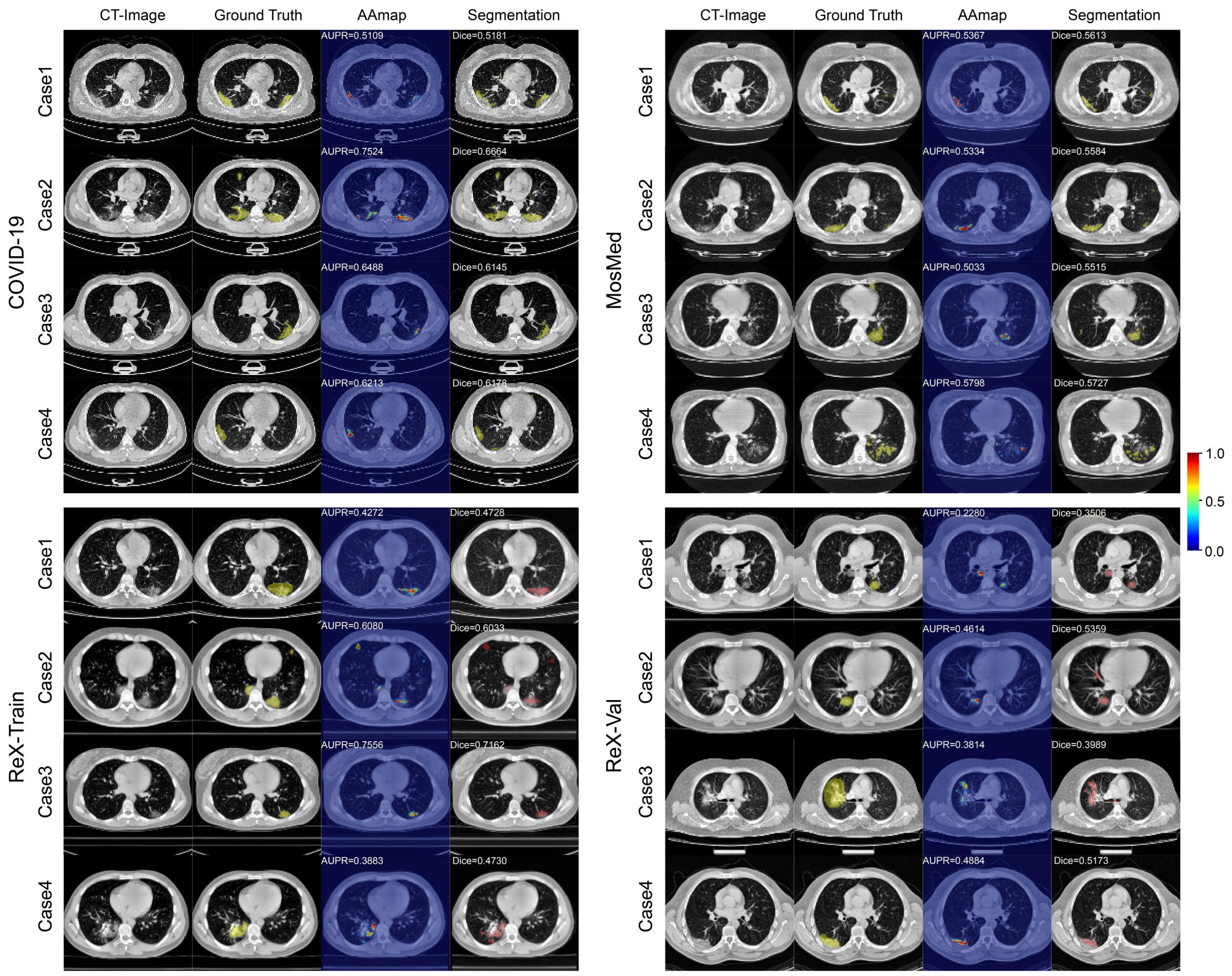

**Supplementary Fig. 10 | Examples of zero-shot anomaly localization results.** Representative qualitative visualizations from the ReX-Train, ReX-Val, COVID-19, and MosMed datasets demonstrate EXACT's capability to precisely map varying lesion morphologies, encompassing diffuse ground-glass opacities (GGOs; all COVID-19 and MosMed cases, ReX-Train cases 1–3, ReX-Val cases 1–2) and dense consolidations (ReX-Train case 4, ReX-Val cases 3–4). Each panel displays (from left to right): the original CT image, ground truth anomaly masks (yellow), EXACT- generated AAmap, and the binary segmentation mask obtained via thresholding. Quantitative metrics (AUPR for AAmaps, DSC for segmentation results) are shown for each case. CT, computed tomography; AAmap, anomaly-aware map; AUPR, area under the precision–recall curve; DSC, Dice similarity coefficient; COVID-19, coronavirus disease 2019.

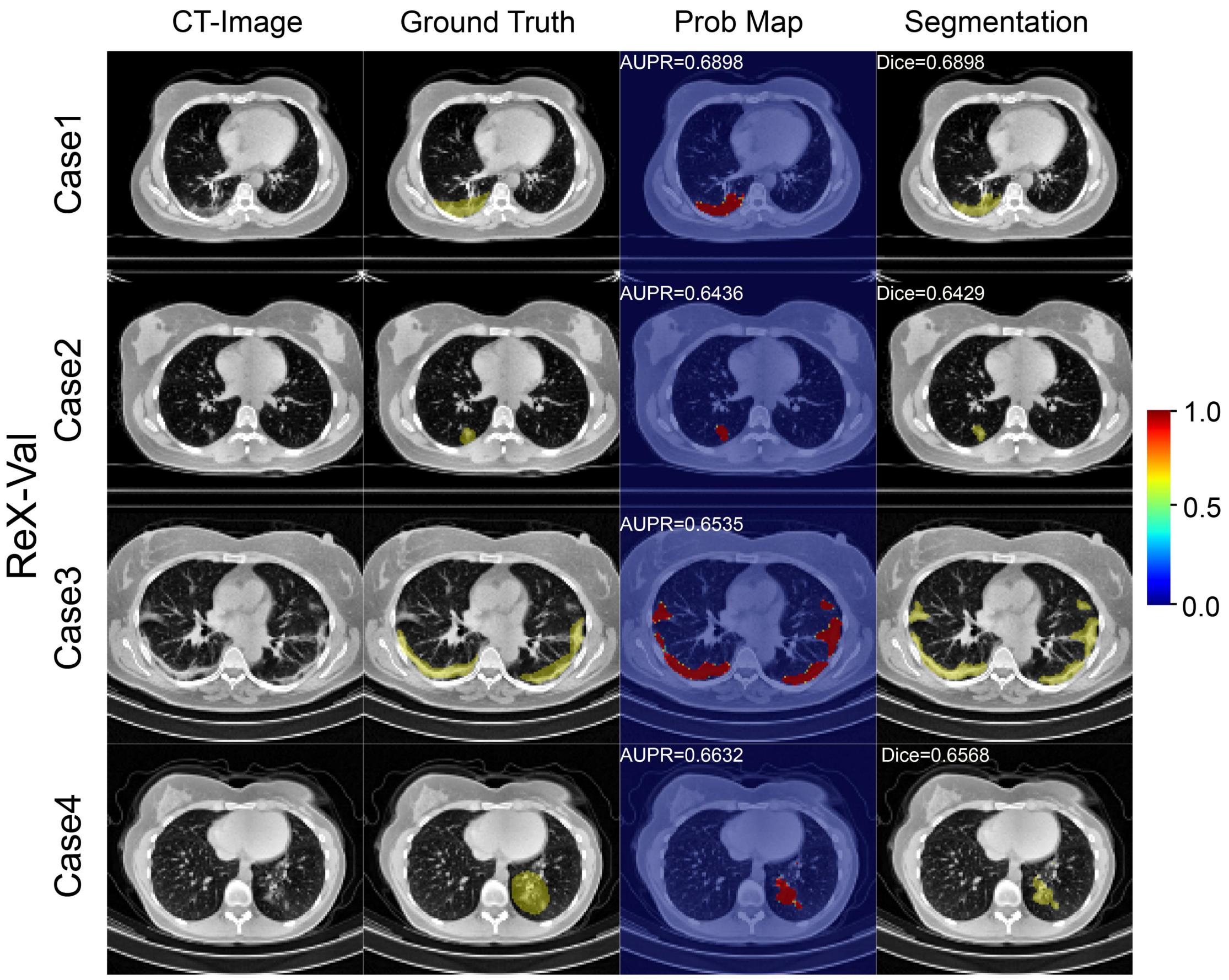

**Supplementary Fig. 11 | Additional examples of EXACT-Seg anomaly localization on the ReX-Val dataset.** Representative qualitative visualizations on the ReX-Val dataset (internal validation). Each panel displays (from left to right): the original CT image, ground truth (yellow), the predicted probability map, and the binary segmentation mask obtained via thresholding. Quantitative metrics (AUPR for probability maps, DSC for segmentation) are shown for each case. CT, computed tomography; Prob Map, probability map; AUPR, area under the precision–recall curve; DSC, Dice similarity coefficient.

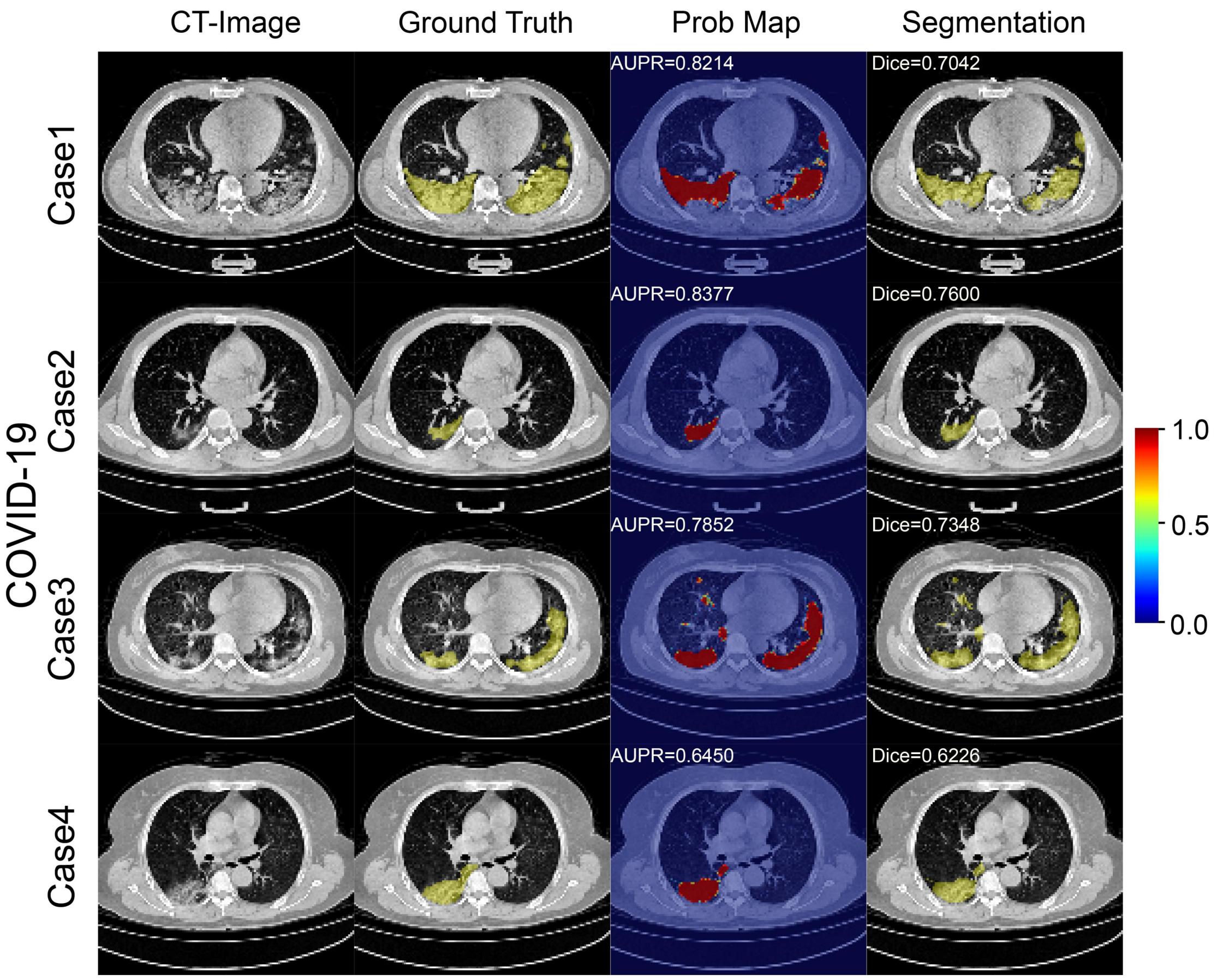

**Supplementary Fig. 12 | Additional examples of EXACT-Seg anomaly localization on the COVID-19 dataset.** Representative qualitative visualizations on the COVID-19 dataset (external validation). Each panel displays (from left to right): the original CT image, ground truth (yellow), the predicted probability map, and the binary segmentation mask obtained via thresholding. Quantitative metrics (AUPR for probability maps, DSC for segmentation) are shown for each case. CT, computed tomography; Prob Map, probability map; AUPR, area under the precision–recall curve; DSC, Dice similarity coefficient; COVID-19, coronavirus disease 2019.

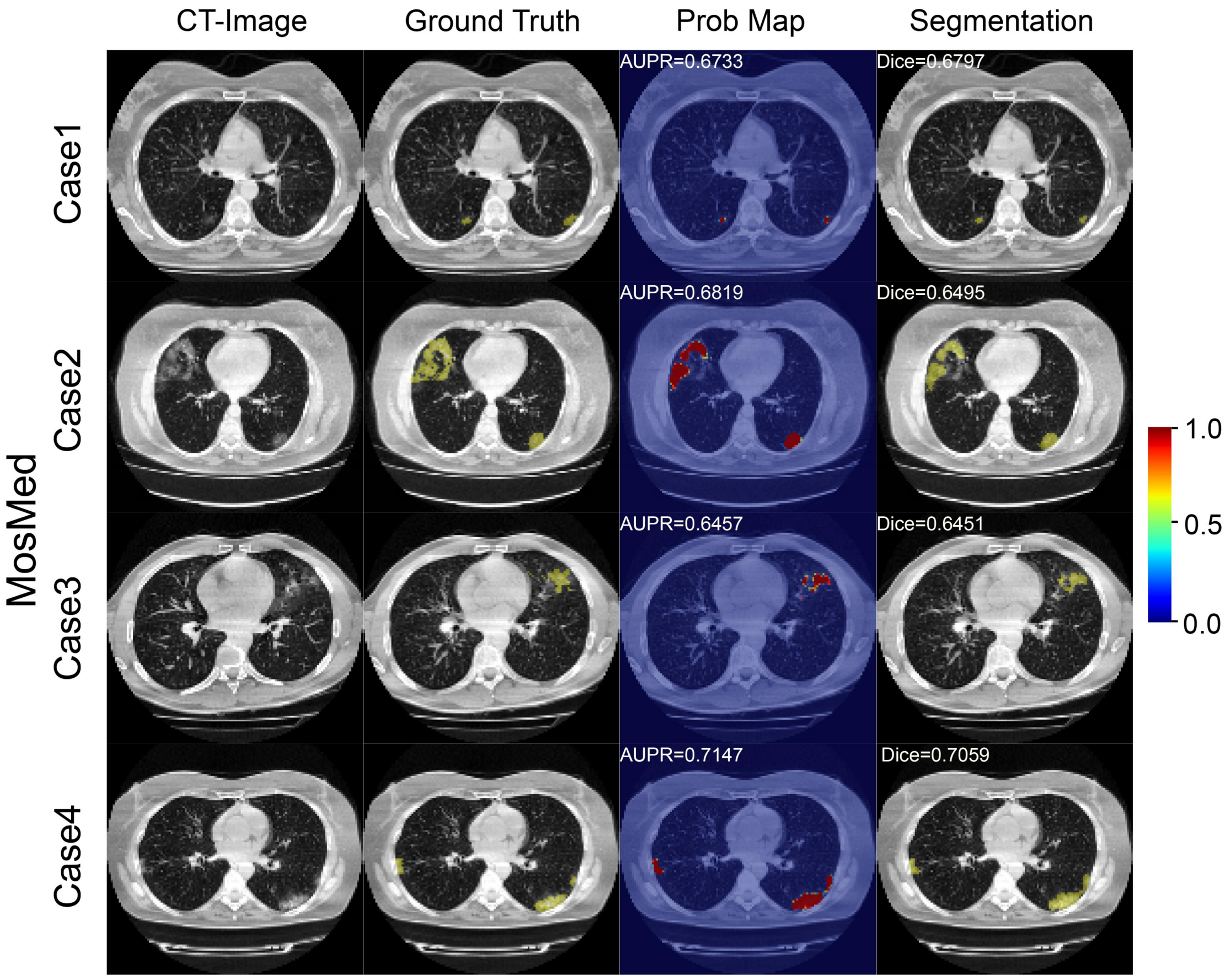

**Supplementary Fig. 13 | Additional examples of EXACT-Seg anomaly localization on the MosMed dataset.** Representative qualitative visualizations on the MosMed dataset (external validation). Each panel displays (from left to right): the original CT image, ground truth (yellow), the predicted probability map, and the binary segmentation mask obtained via thresholding. Quantitative metrics (AUPR for probability maps, DSC for segmentation) are shown for each case. CT, computed tomography; Prob Map, probability map; AUPR, area under the precision–recall curve; DSC, Dice similarity coefficient.

### Reference Report

**Findings:**
The mediastinal main vascular structures and the heart could not be evaluated optimally due to the lack of IV contrast, and the calibration of the vascular structures, the heart contour and size are natural. No pericardial, pleural effusion or thickening was detected. Trachea, both main bronchi are open and no occlusive pathology is detected. No pathological increase in thoracic esophagus wall thickness is observed. No lymph node is observed in the mediastinum and in both axillary regions in pathological size and appearance. When examined in the lung parenchyma window; Sequelae of pleuroparenchymal fibrotic bands were observed in both lung lower lobe posterobasal segment and left lung upper lobe inferior lingular segment. In addition, there are sequela parenchymal changes in the apex of both lungs. A few millimeter-sized non-specific nodules were observed in both lungs. Ventilation of both lungs is natural. No active infiltrative or mass lesion was detected in both lungs. No pathology was detected in the upper abdominal sections within the image. No lytic-destructive lesion was observed in the bone structures within the image, and the vertebral corpus heights were preserved.
**Impression:**
There is no finding in favor of pneumonic infiltration in both lungs, and there are occasional sequela parenchymal changes and a few millimeter-sized non-specific nodules.

#### EXACT-CHAT (w/o Prior)

**Findings:**
Trachea and both main bronchi are open. No occlusive pathology was detected in the trachea and both main bronchi. There are minimal emphysematous changes in both lungs. There are millimetric nonspecific nodules in both lungs. No mass or infiltrative lesion was detected in both lungs. Mediastinal structures cannot be evaluated optimally because contrast material is not given. As far as can be observed: Heart contour and size are normal. No pleural or pericardial effusion was detected. The widths of the mediastinal main vascular structures are normal. No pathologically enlarged lymph nodes were detected in the mediastinum and hilar regions. No pathological wall thickness increase was observed in the esophagus within the sections. No upper abdominal free fluid-collection was detected in the sections. No enlarged lymph nodes in pathological dimensions were detected. In the upper abdominal organs within the sections, there is no mass with distinguishable borders as far as it can be observed within the borders of non-enhanced CT. Thoracic vertebral corpus heights, alignments and densities are normal. Intervertebral disc distances are preserved. The neural foramina are open. No lytic-destructive lesions were detected in the bone structures within the sections.
**Impression:**
Minimal emphysematous changes in both lungs. Millimetric nonspecific nodules in both lungs.

**Missed Findings:**
Sequelae of pleuroparenchymal fibrotic bands

#### EXACT-CHAT

**Findings:**
Trachea and both main bronchi are open. No occlusive pathology was detected in the trachea and both main bronchi. There are millimetric nodules in both lungs. No mass or infiltrative lesion was detected in both lungs. Mediastinal structures cannot be evaluated optimally because contrast material is not given. As far as can be observed: Heart contour and size are normal. No pleural or pericardial effusion was detected. The widths of the mediastinal main vascular structures are normal. No pathologically enlarged lymph nodes were detected in the mediastinum and hilar regions. No pathological wall thickness increase was observed in the esophagus within the sections. No upper abdominal free fluid-collection was detected in the sections. No enlarged lymph nodes in pathological dimensions were detected. In the upper abdominal organs within the sections, there is no mass with distinguishable borders as far as it can be observed within the borders of non-enhanced CT. Thoracic vertebral corpus heights, alignments and densities are normal. Intervertebral disc distances are preserved. The neural foramina are open. No lytic-destructive lesions were detected in the bone structures within the sections.
**Impression:**
Millimetric nodules in both lungs.

**Missed Findings:**
Sequelae of pleuroparenchymal fibrotic bands

#### EXACT-CHAT (Refined)

**Findings:**
Trachea and both main bronchi are open. No occlusive pathology was detected in the trachea and both main bronchi. There are millimetric nodules in both lungs. No mass or infiltrative lesion was detected in both lungs. Mediastinal structures cannot be evaluated optimally because contrast material is not given. As far as can be observed: Heart contour and size are normal. No pleural or pericardial effusion was detected. The widths of the mediastinal main vascular structures are normal. No pathologically enlarged lymph nodes were detected in the mediastinum and hilar regions. No pathological wall thickness increase was observed in the esophagus within the sections. No upper abdominal free fluid-collection was detected in the sections. No enlarged lymph nodes in pathological dimensions were detected. In the upper abdominal organs within the sections, there is no mass with distinguishable borders as far as it can be observed within the borders of non-enhanced CT. Thoracic vertebral corpus heights, alignments and densities are normal. Intervertebral disc distances are preserved. The neural foramina are open. No lytic-destructive lesions were detected in the bone structures within the sections. In the lung parenchyma window, findings consistent with pulmonary fibrotic sequelae are observed.
**Impression:**
Millimetric nodules in both lungs. Findings consistent with pulmonary fibrotic sequelae.

**Missed Findings:**
None.

### Reference Report

**Findings:**
Trachea and both main bronchi are open. No occlusive pathology was detected in the trachea and both main bronchi. Consolidation in the medial part of the right lung lower lobe superior segment and a ground glass area around it are observed. A ground glass area is also observed in the left lung lower lobe superior segment. The views described are not specific. These appearances were thought to belong primarily to a pneumonic infiltration. The presence of a ground glass area in the superior segment of the lower lobe of the left lung suggests that this appearance may be a viral pneumonia. It is recommended to evaluate the patient together with clinical, physical examination and laboratory findings. No mass was detected in both lungs. Mediastinal structures cannot be evaluated optimally because contrast material is not given. As far as can be observed: Heart contour and size are normal. No pleural or pericardial effusion was detected. The widths of the mediastinal main vascular structures are normal. There are no pathologically enlarged lymph nodes in the mediastinum and hilar regions. No pathological increase in wall thickness was detected in the esophagus within the sections. There is no upper abdominal free fluid-collection within the sections. No enlarged lymph nodes in pathological dimensions were detected. In the liver parenchyma density, there is a decrease in density compatible with advanced adiposity. Thoracic vertebral corpus heights, alignments and densities are normal. The neural foramina are open.
**Impression:**
Appearances compatible with pneumonic infiltration in the right lung lower lobe superior segment and left lung lower lobe . Hepatic steatosis.

#### EXACT-CHAT (w/o Prior)

**Findings:**
Trachea and both main bronchi are open. No occlusive pathology was detected in the trachea and both main bronchi. There are minimal emphysematous changes in both lungs. There are linear atelectasis in the right lung middle lobe medial segment and left lung upper lobe lingular segment. There are millimetric nonspecific nodules in both lungs. No mass or infiltrative lesion was detected in both lungs. Mediastinal structures cannot be evaluated optimally because contrast material is not given. As far as can be observed: Heart contour and size are normal. No pleural or pericardial effusion was detected. The widths of the mediastinal main vascular structures are normal. No pathologically enlarged lymph nodes were detected in the mediastinum and hilar regions. No pathological wall thickness increase was observed in the esophagus within the sections. No upper abdominal free fluid-collection was detected in the sections. No enlarged lymph nodes in pathological dimensions were detected. In the upper abdominal organs within the sections, there is no mass with distinguishable borders as far as it can be observed within the borders of non-enhanced CT. Thoracic vertebral corpus heights, alignments and densities are normal. Intervertebral disc distances are preserved. The neural foramina are open. No lytic-destructive lesions were detected in the bone structures within the sections.
**Impression:**
Minimal emphysematous changes in both lungs. Millimetric nonspecific nodules in both lungs.

**Missed Findings:**
Pneumonic infiltration (Consolidation & Ground-glass opacities), Hepatic steatosis

#### EXACT-CHAT

**Findings:**
Trachea and both main bronchi are open. No occlusive pathology was detected in the trachea and both main bronchi. There are ground-glass appearances in the upper and lower lobes of both lungs and the middle lobe of the right lung. The described views are more prominent in the peripheral regions. The described manifestations were evaluated primarily in favor of viral pneumonia. These findings are frequently observed in Covid-19 pneumonia. No mass was detected in both lungs. Mediastinal structures cannot be evaluated optimally because contrast material is not given. As far as can be observed: Heart contour and size are normal. No pleural or pericardial effusion was detected. The widths of the mediastinal main vascular structures are normal. No pathologically enlarged lymph nodes were detected in the mediastinum and hilar regions. No pathological wall thickness increase was observed in the esophagus within the sections. No upper abdominal free fluid-collection was detected in the sections. No enlarged lymph nodes in pathological dimensions were detected. In the upper abdominal organs within the sections, there is no mass with distinguishable borders as far as it can be observed within the borders of non-enhanced CT. Thoracic vertebral corpus heights, alignments and densities are normal. Intervertebral disc distances are preserved. The neural foramina are open. No lytic-destructive lesions were detected in the bone structures within the sections.
**Impression:**
Findings consistent with viral pneumonia in both lungs.

**Missed Findings:**
Consolidation, Hepatic steatosis.

#### EXACT-CHAT (Refined)

**Findings:**
Trachea and both main bronchi are open. No occlusive pathology was detected in the trachea and both main bronchi. There are ground-glass opacities and consolidation areas in the upper and lower lobes of both lungs as well as in the middle lobe of the right lung. The described findings are more prominent in the peripheral regions. These manifestations were evaluated primarily in favor of viral pneumonia and are frequently observed in Covid-19 pneumonia. No mass was detected in both lungs. Mediastinal structures cannot be evaluated optimally because contrast material is not given. As far as can be observed: Heart contour and size are normal. No pleural or pericardial effusion was detected. The widths of the mediastinal main vascular structures are normal. No pathologically enlarged lymph nodes were detected in the mediastinum and hilar regions. No pathological wall thickness increase was observed in the esophagus within the sections. No upper abdominal free fluid-collection was detected in the sections. No enlarged lymph nodes in pathological dimensions were detected. In the upper abdominal organs within the sections, there is no mass with distinguishable borders as far as it can be observed within the borders of non-enhanced CT. Thoracic vertebral corpus heights, alignments and densities are normal. Intervertebral disc distances are preserved. The neural foramina are open. No lytic-destructive lesions were detected in the bone structures within the sections.
**Impression:**
Peripheral ground-glass opacities and consolidation areas in both lungs, findings consistent with viral pneumonia, typical for Covid-19 pneumonia.

**Missed Findings:**
Hepatic steatosis.

**Supplementary Fig. 14 | Representative case studies of report generation using EXACT-CHAT (CT-RATE dataset).** Qualitative comparison of generated reports against ground-truth references across three EXACT-CHAT variants: EXACT-CHAT (w/o Prior), EXACT-CHAT, and EXACT-CHAT (Refined), which incorporates GPT-4.1 for report refinement. Red text indicates hallucinations or factual errors, green text highlights correct findings consistent with the reference, and blue boxes summarize missed findings for each variant. The EXACT-CHAT (Refined) demonstrates the highest consistency with the reference reports, exhibiting fewer hallucinations and missed findings compared with the other variants. CT, computed tomography; IV, intravenous; w/o Prior, without diagnostic priors; COVID-19, coronavirus disease 2019.

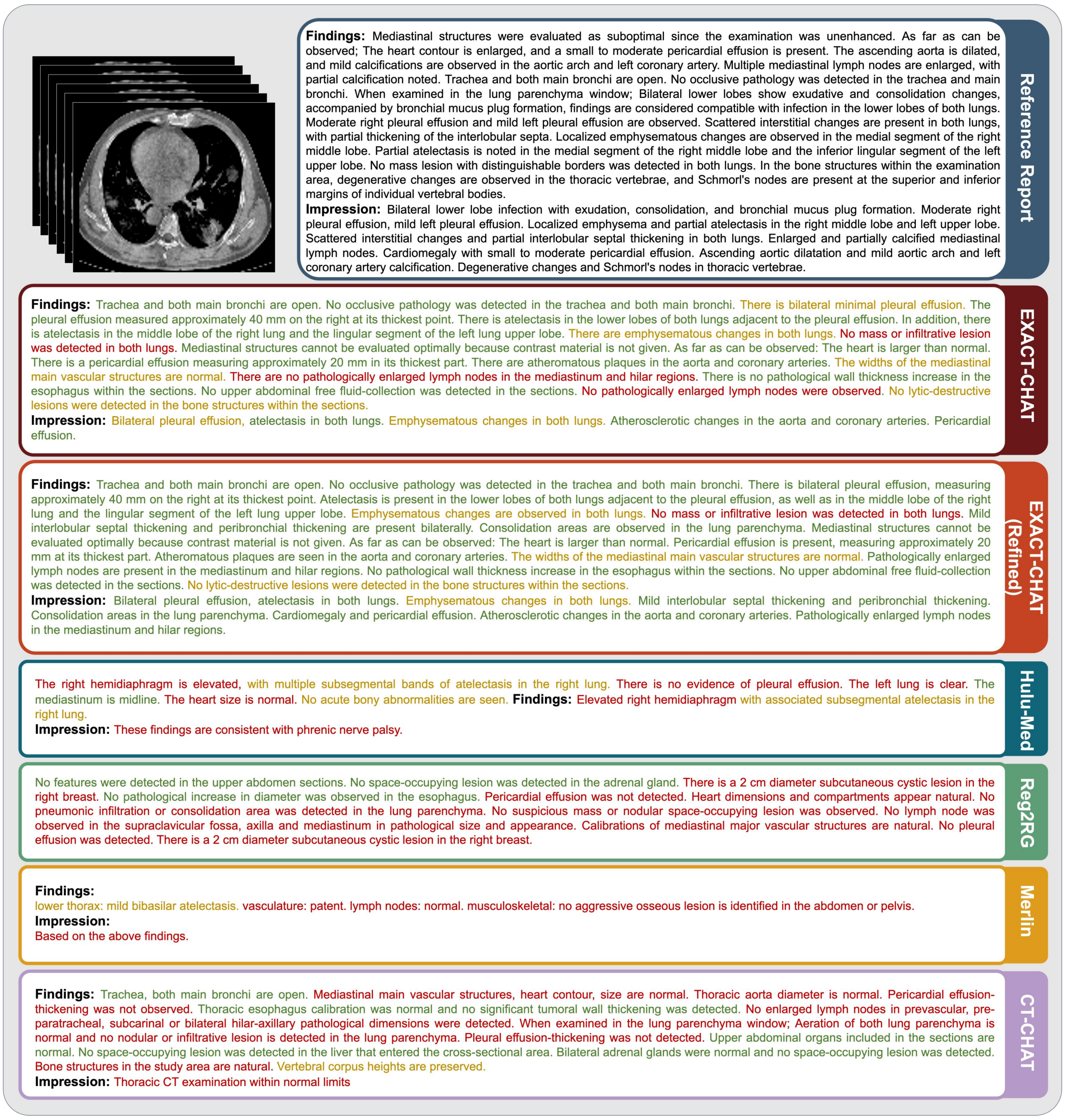

**Supplementary Fig. 15 | Qualitative comparison of report generation performance between EXACT-CHAT and baseline models (MianYang dataset).** Representative examples comparing generated radiology reports against ground-truth references. The examples show unrefined reports from Hulu-Med, Reg2RG, Merlin and CT-CHAT, alongside EXACT-CHAT reports before and after GPT-4.1 post-hoc refinement. Red text indicates hallucinations or factual errors, yellow text highlights partially correct or relevant findings, and green text denotes accurate detections. In contrast to baseline models, which exhibit significant hallucinations or missed findings, EXACT-CHAT (Refined) generates the most comprehensive and clinically accurate reports, demonstrating high alignment with the reference standard.

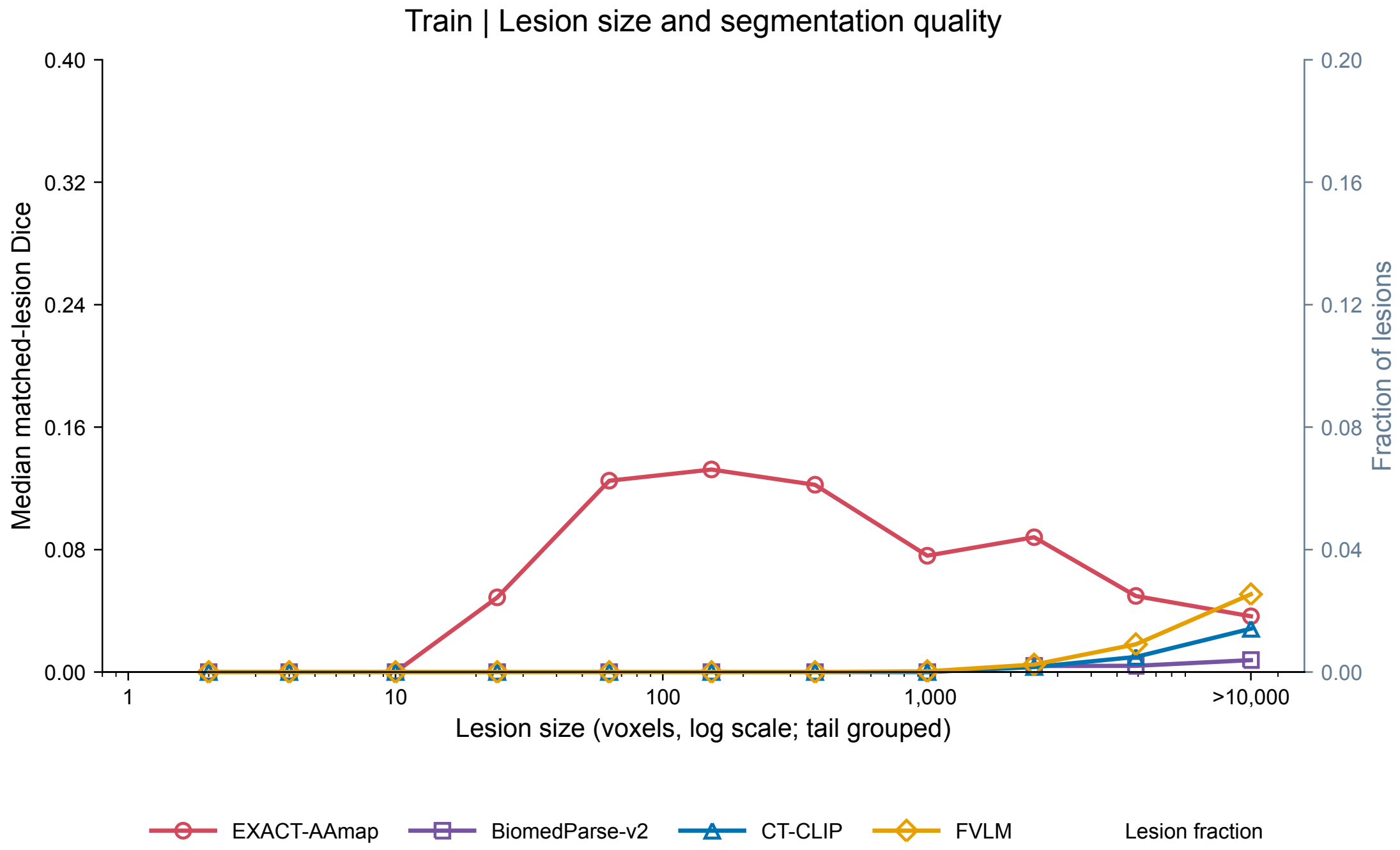

**Supplementary Fig. 16 | Lesion-size-stratified zero-shot anomaly localization performance on ReX-Train.** Performance of EXACT-AAmap, BiomedParse-v2, CT-CLIP and fVLM across lesion-size bins. Solid lines show the median matched-lesion Dice score within each bin (left axis), and shaded bars show the corresponding fraction of lesions (right axis). Lesion size is expressed as voxel count on a logarithmic scale, with the largest lesions grouped in the final bin. All methods were evaluated in the zero-shot setting.

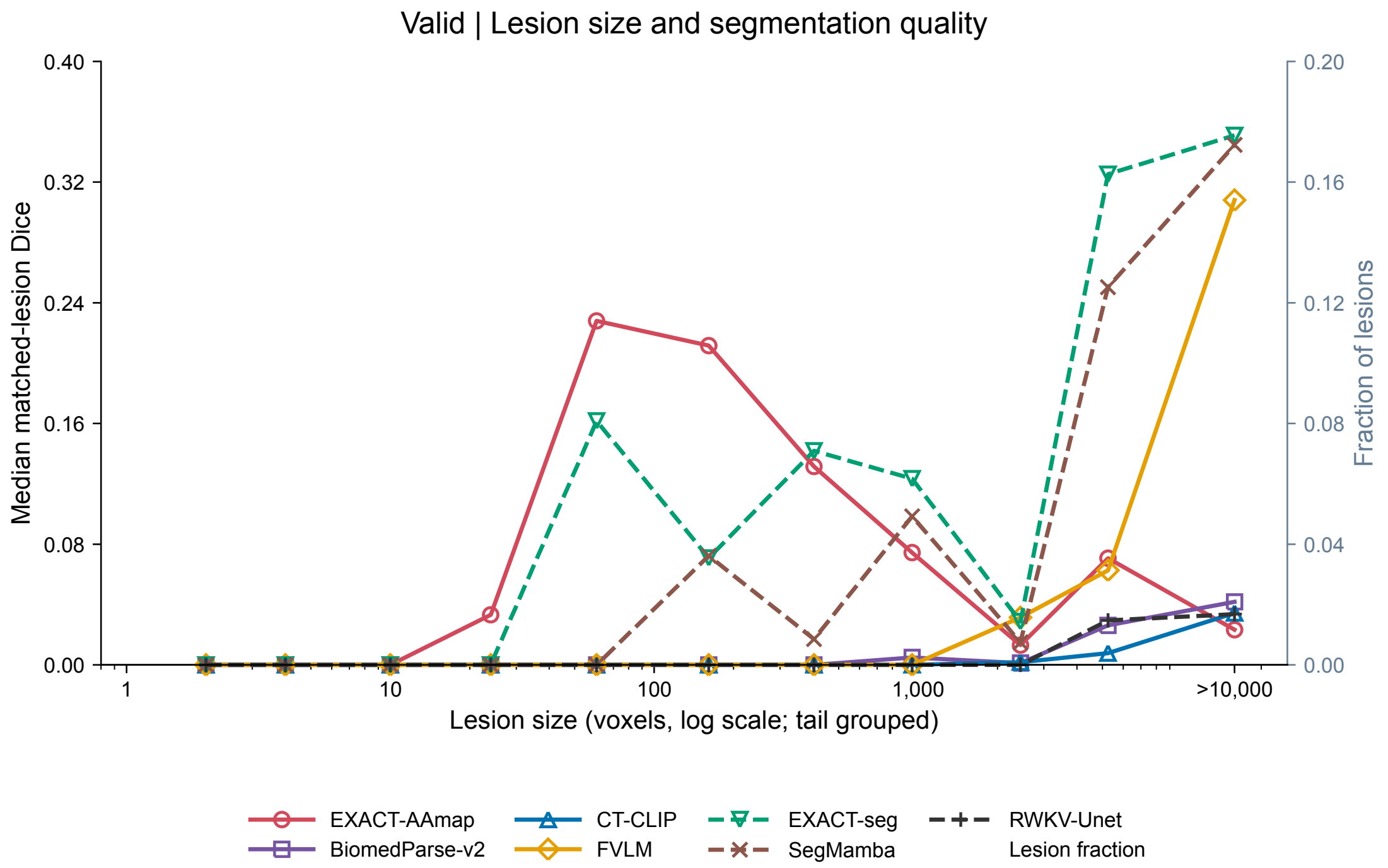

**Supplementary Fig. 17 | Lesion-size-stratified anomaly localization performance on ReX-Val under zero-shot evaluation and supervised adaptation.** Solid lines show zero-shot results for EXACT-AAmap, BiomedParse-v2, CT-CLIP and fVLM; dashed lines show results after supervised adaptation for EXACT-Seg, SegMamba and RWKV-UNet. Lines indicate the median matched-lesion Dice score within each size bin (left axis), and shaded bars indicate the corresponding fraction of lesions (right axis). Lesion size is expressed as voxel count on a logarithmic scale, with the largest lesions grouped in the final bin.

## Supplementary Tables

| Abnormality | Abbreviation | CT-RATE (Training) *n* = 24,128 | CT-RATE (Internal Val.) *n* = 1,564 | Rad-ChestCT (External Val.) n = 3,630 | MianYang (External Val.) *n* = 505 |
|---|---|---|---|---|---|
| | | No. (Ratio) | | | |
| **Medical material** | MM | 2,886 (0.120) | 158 (0.101) | 1,149 (0.317) | 18 (0.036) |
| **Arterial wall calcification** | AWC | 6,701 (0.278) | 435 (0.278) | 1,119 (0.308) | 118 (0.234) |
| **Cardiomegaly** | CME | 2,586 (0.107) | 164 (0.105) | 394 (0.109) | 24 (0.048) |
| **Pericardial effusion** | PCE | 1,702 (0.071) | 110 (0.070) | 563 (0.155) | 37 (0.073) |
| **Coronary artery wall calcification** | CAWC | 5,972 (0.248) | 388 (0.248) | N/A | 94 (0.186) |
| **Hiatal hernia** | HH | 3,426 (0.142) | 218 (0.139) | 425 (0.117) | 4 (0.008) |
| **Lymphadenopathy** | LAP | 6,115 (0.253) | 398 (0.254) | 598 (0.165) | 69 (0.137) |
| **Emphysema** | EMSE | 4,677 (0.194) | 305 (0.195) | 992 (0.273) | 80 (0.158) |
| **Atelectasis** | ATE | 6,170 (0.256) | 364 (0.233) | 1,082 (0.298) | 129 (0.255) |
| **Lung nodule** | LN | 10,974 (0.455) | 704 (0.450) | 2,873 (0.791) | 456 (0.903) |
| **Lung opacity** | GGO | 8,846 (0.367) | 606 (0.387) | 1,955 (0.539) | 93 (0.184) |
| **Pulmonary fibrotic sequela** | PFS | 6,447 (0.267) | 427 (0.273) | 478 (0.132) | 28 (0.055) |
| **Pleural effusion** | PLE | 2,858 (0.118) | 189 (0.121) | 727 (0.200) | 38 (0.075) |
| **Mosaic attenuation pattern** | MAP | 1,794 (0.074) | 126 (0.081) | N/A | 0 (0.000) |
| **Peribronchial thickening** | PBT | 2,496 (0.103) | 175 (0.112) | 308 (0.085) | 16 (0.032) |
| **Consolidation** | CON | 4,234 (0.175) | 298 (0.191) | 505 (0.139) | 39 (0.077) |
| **Bronchiectasis** | BE | 2,411 (0.100) | 169 (0.108) | 597 (0.164) | 37 (0.073) |
| **Interlobular septal thickening** | ILST | 1,902 (0.079) | 124 (0.079) | 252 (0.069) | 211 (0.418) |

**Supplementary Table 1 | Detailed overview of datasets for multi-disease diagnosis.** Distribution of 18 target abnormalities across CT-RATE (training and internal validation), RAD-ChestCT (external validation), and MianYang (external validation) datasets. Data are presented as the number of positive scans and the corresponding proportion within each dataset partition. “N/A” indicates that the label is unavailable in the corresponding dataset. Abbreviations: Internal Val. = internal validation; External Val. = external validation.

| Organs | Y-Mamba DSC |
|---|---|
| **Lung** | 0.966 ± 0.088 |
| **T & B** | 0.886 ± 0.010 |
| **Pleura** | 0.966 ± 0.088 |
| **Mediastinum** | 0.811 ± 0.093 |
| **Heart** | 0.889 ± 0.114 |
| **Esophagus** | 0.821 ± 0.115 |

**Supplementary Table 2 | Organ segmentation performance of EXACT.** Dice Similarity Coefficient (DSC) scores achieved by the Y-Mamba architecture across six anatomical regions: Lung, Trachea & Bronchi (T & B), Pleura, Mediastinum, Heart, and Esophagus. Data are presented as mean ± standard deviation.

| Organ | MM | AWC | CME | PCE | CAWC | HH | LAP | EMSE | ATE | LN | GGO | PFS | PLE | MAP | PBT | CON | BE | ILST |
|---|---|---|---|---|---|---|---|---|---|---|---|---|---|---|---|---|---|---|
| **Lung** | | | | | | | | √ | √ | √ | √ | √ | | √ | | √ | | √ |
| **Trachea and Bronchi** | | | | | | | | | | | | | | | √ | | √ | |
| **Pleura** | | | | | | | | | | | | | √ | | | | | |
| **Mediastinum** | | | | | | | √ | | | | | | | | | | | |
| **Heart** | | | √ | √ | √ | | | | | | | | | | | | | |
| **Esophagus** | | | | | | √ | | | | | | | | | | | | |
| **Global** | √ | √ | | | | | | | | | | | | | | | | |

**Supplementary Table 3 | Specific associations between diseases and organs.** Target abnormalities were mapped to six anatomical regions (lung; trachea and bronchi; pleura; mediastinum; heart; esophagus) and to a global category. Checkmarks indicate that each abnormality was anatomically constrained to the corresponding organ region during model training and inference. Disease abbreviations follow those defined in Supplementary Table 1.

| Dataset | Models | Metric [95% CI] | | | |
|---|---|---|---|---|---|
| | | AUROC | F1 Score | Accuracy | *P*-value |
| **Internal Validation** **CT-RATE Dataset (*n* = 1,564)** | CT-Net (Fine-tuning) | 0.629 | 0.657 | 0.617 | AUROC: 0.001 F1: 0.687. Acc: 0.481. |
| | CT-CLIP (Zero-shot) | 0.731 [0.721, 0.740] | 0.761 [0.750, 0.777] | 0.669 [0.662, 0.691] | |
| | CT-CLIP (VocabFine) | 0.757 [0.748, 0.767] | **0.794 [0.782, 0.803]** | 0.710 [0.698, 0.722] | |
| | CT-CLIP (ClassFine) | 0.758 [0.749, 0.767] | 0.786 [0.775, 0.797] | 0.703 [0.691, 0.716] | |
| | fVLM (Zero-shot) | 0.777 [0.769, 0.786] | 0.802 [0.791, 0.813] | 0.725 [0.715, 0.739] | |
| | MedVista3D (Zero-shot) | 0.782 | 0.770 | 0.745 | |
| | T3D (Zero-shot) | 0.737 | 0.725 | 0.690 | |
| | T3D ( Fine-tuning) | **0.802** | **0.778** | **0.763** | |
| | RadZero3D (Zero-shot) | 0.762 | 0.742 | 0.701 | |
| | BIUD (Zero-shot) | 0.713 | 0.716 | 0.681 | |
| | **EXACT (Zero-shot)** | 0.830 [0.822, 0.837] | 0.836 [0.826, 0.844] | 0.768 [0.757, 0.778] | |
| | **EXACT (Fine-tuning)** | 0.833 [0.826, 0.841] | 0.836 [0.829, 0.844] | 0.769 [0.762, 0.780] | |
| **External Validation** **RAD-ChestCT Dataset (*n* = 3,630)** | CT-Net (Fine-tuning) | 0.544 | 0.564 | 0.517 | AUROC< 0.001 F1: 0.105 Acc: 0.025 |
| | CT-CLIP (Zero-shot) | 0.625 [0.618, 0.632] | 0.679 [0.671, 0.697] | 0.598 [0.591, 0.618] | |
| | CT-CLIP (VocabFine) | 0.646 [0.638, 0.652] | **0.692 [0.681, 0.702]** | 0.618 [0.608, 0.629] | |
| | CT-CLIP (ClassFine) | 0.637 [0.630, 0.644] | 0.689 [0.679, 0.702] | 0.615 [0.605, 0.627] | |
| | fVLM (Zero-shot) | 0.640 [0.633, 0.647] | 0.683 [0.675, 0.702] | 0.613 [0.602, 0.632] | |
| | MedVista3D (Zero-shot) | **0.710** | 0.681 | **0.668** | |
| | Merlin (Zero-shot) | 0.595 [0.588, 0.603] | 0.664 [0.657, 0.679] | 0.589 [0.579, 0.603] | |
| | BIUD (Zero-shot) | 0.629 | 0.652 | 0.606 | |
| | **EXACT (Zero-shot)** | 0.728 [0.722, 0.734] | 0.735 [0.728, 0.745] | 0.678 [0.671, 0.689] | |
| | **EXACT (Fine-tuning)** | 0.734 [0.728, 0.740] | 0.737 [0.732, 0.748] | 0.682 [0.677, 0.694] | |
| **External Validation** **MianYang Dataset (*n* = 505)** | CT-Net (Fine-tuning) | 0.686 [0.658, 0.711] | 0.752 [0.722, 0.771] | **0.691 [0.653, 0.717]** | AUROC: 0.006 F1: < 0.001 Acc:< 0.001 |
| | CT-CLIP (Zero-shot) | 0.689 [0.667, 0.709] | 0.735 [0.718, 0.762] | 0.668 [0.644, 0.700] | |
| | CT-CLIP (VocabFine) | 0.712 [0.688, 0.736] | 0.752 [0.733, 0.776] | 0.677 [0.659, 0.714] | |
| | CT-CLIP (ClassFine) | 0.704 [0.680, 0.728] | 0.749 [0.729, 0.778] | 0.687 [0.656, 0.721] | |
| | fVLM (Zero-shot) | **0.716 [0.695, 0.737]** | **0.752 [0.732, 0.781]** | 0.690 [0.671, 0.727] | |
| | Merlin (Zero-shot) | 0.602 [0.575, 0.629] | 0.714 [0.677, 0.752] | 0.632 [0.594, 0.683] | |
| | **EXACT (Zero-shot)** | 0.758 [0.739, 0.778] | 0.780 [0.757, 0.800] | 0.725 [0.699, 0.755] | |
| | **EXACT (Fine-tuning)** | 0.769 [0.748, 0.787] | 0.794 [0.767, 0.815] | 0.748 [0.712, 0.775] | |

**Supplementary Table 4 | Comparison of multi-disease diagnosis performance.** Performance metrics of EXACT variants and baseline models for multi-disease diagnosis. Evaluations were performed on CT-RATE (internal validation), RAD-ChestCT (external validation), and MianYang (external validation) datasets. Metrics include AUROC, F1 Score, and Accuracy. Independently evaluated results are presented with 95% CIs estimated using 1,000 bootstrap resamples; publication-reported results are shown without CIs. **Dataset-specific result sources are detailed in Methods (EXACT for multi-disease diagnosis).** Red text indicates the best performance, blue text denotes the second best, and bold text represents the third best in each column. *P* values compare EXACT (Zero-shot) with EXACT (Fine-tuning) within each dataset using two-sided bootstrap hypothesis tests with 2,000 resamples.

| Dataset | Models | Metric [95% CI] | | | | | | |
|---|---|---|---|---|---|---|---|---|
| | | DSC | HIT 5% | HIT 10% | AUPR | HIT 5% | HIT 10% | *p*-value |
| **Task: Zero-shot Anomaly Localization** | | | | | | | | |
| **Internal Validation ReX-Train (*n* = 1102)** | BiomedParse-v2 | 0.032 [0.027, 0.037] | 0.145 | 0.100 | 0.026 [0.022, 0.030] | 0.125 | 0.066 | Dice: < 0.001 AUPR: < 0.001 |
| | fVLM | 0.008 [0.006, 0.011] | 0.039 | 0.020 | 0.011 [0.009, 0.013] | 0.047 | 0.021 | |
| | CT-CLIP | 0.004 [0.003, 0.004] | 0.008 | 0.004 | 0.001 [0.001, 0.002] | 0.004 | 0.000 | |
| | **EXACT** | 0.096 [0.089, 0.103] | 0.472 | 0.350 | 0.093 [0.085, 0.101] | 0.432 | 0.297 | |
| **Internal Validation ReX-Val (*n* = 157)** | BiomedParse-v2 | 0.071 [0.053, 0.092] | 0.312 | 0.229 | 0.031 [0.023, 0.039] | 0.185 | 0.076 | Dice: < 0.001 AUPR: < 0.001 |
| | fVLM | 0.009 [0.005, 0.015] | 0.038 | 0.032 | 0.014 [0.009, 0.018] | 0.064 | 0.032 | |
| | CT-CLIP | 0.005 [0.004, 0.007] | 0.006 | 0.000 | 0.002 [0.002, 0.003] | 0.000 | 0.000 | |
| | **EXACT** | 0.125 [0.105, 0.147] | 0.605 | 0.439 | 0.125 [0.103, 0.147] | 0.567 | 0.420 | |
| **External Validation COVID-19 (*n* = 20)** | BiomedParse-v2 | 0.340 [0.185, 0.490] | 0.550 | 0.500 | 0.459 [0.303, 0.632] | 0.900 | 0.750 | Dice: n.s. AUPR: n.s. |
| | fVLM | 0.081 [0.041, 0.121] | 0.500 | 0.300 | 0.059 [0.035, 0.087] | 0.450 | 0.200 | |
| | CT-CLIP | 0.022 [0.010, 0.035] | 0.000 | 0.000 | 0.010 [0.005, 0.016] | 0.000 | 0.000 | |
| | **EXACT** | 0.435 [0.341, 0.517] | 0.950 | 0.850 | 0.530 [0.440, 0.609] | 0.950 | 0.900 | |
| **External Validation MosMed (*n* = 50)** | BiomedParse-v2 | 0.254 [0.196, 0.315] | 0.840 | 0.660 | 0.258 [0.201, 0.321] | 0.820 | 0.660 | Dice: < 0.001 AUPR: 0.002 |
| | fVLM | 0.016 [0.012, 0.020] | 0.060 | 0.000 | 0.014 [0.010, 0.019] | 0.040 | 0.000 | |
| | CT-CLIP | 0.004 [0.003, 0.005] | 0.000 | 0.000 | 0.002 [0.001, 0.003] | 0.000 | 0.000 | |
| | **EXACT** | 0.362 [0.320, 0.406] | 0.960 | 0.920 | 0.330 [0.283, 0.376] | 0.960 | 0.900 | |
| **Task: Anomaly Localization with Supervised Finetuning** | | | | | | | | |
| **Internal Validation Rex-Val (*n* = 157)** | RWKV-Unet | 0.112 [0.089, 0.135] | 0.312 | 0.242 | 0.180 [0.145, 0.219] | 0.580 | 0.465 | Dice: 0.028 AUPR: 0.007 |
| | SegMamba | 0.198 [0.165, 0.230] | 0.573 | 0.503 | 0.187 [0.154, 0.223] | 0.548 | 0.478 | |
| | **EXACT-Seg** | 0.215 [0.182, 0.249] | 0.643 | 0.580 | 0.200 [0.165, 0.238] | 0.592 | 0.478 | |
| **External Validation COVID-19 (*n* = 16)** | RWKV-Unet | 0.305 [0.205, 0.412] | 0.812 | 0.812 | 0.404 [0.292, 0.513] | 0.875 | 0.812 | Dice: < 0.001 AUPR: 0.006 |
| | SegMamba | 0.332 [0.221, 0.450] | 0.812 | 0.750 | 0.358 [0.235, 0.493] | 0.750 | 0.750 | |
| | **EXACT-Seg** | 0.476 [0.332, 0.621] | 0.938 | 0.813 | 0.529 [0.374, 0.679] | 0.875 | 0.875 | |
| **External Validation MosMed (*n* = 40)** | RWKV-Unet | 0.348 [0.290, 0.405] | 0.950 | 0.875 | 0.373 [0.311, 0.438] | 0.950 | 0.850 | Dice: < 0.001 AUPR: < 0.001 |
| | SegMamba | 0.352 [0.252, 0.378] | 0.850 | 0.850 | 0.324 [0.255, 0.393] | 0.825 | 0.800 | |
| | **EXACT-Seg** | 0.454 [0.387, 0.520] | 0.950 | 0.900 | 0.463 [0.393, 0.536] | 0.925 | 0.925 | |

**Supplementary Table 5 | Comparison of anomaly localization performance.** Performance metrics of EXACT variants and baseline models under zero-shot and supervised fine-tuning settings. Evaluations were performed on ReX-Train and ReX-Val (internal validation), as well as COVID-19 and MosMed (external validation) datasets. Data are presented as mean values with 95% CIs. Red text indicates the best performance in each column. *P* values were calculated using the two-sided Wilcoxon signed-rank test between the top two performing methods. Abbreviations: DSC = Dice similarity coefficient; HIT 5%/10% = Hit Rate at 5%/10% thresholds; AUPR = Area Under the Precision-Recall curve; CI = confidence interval.

| Dataset | Models | Metric [95% CI] | | | | | | | |
|---|---|---|---|---|---|---|---|---|---|
| | | **BLEU-1** | **METEOR** | **CIDEr** | **ROUGE-L** | **RadBERT-F1 Score** | **RadBERT-Precision** | **RadBERT-Recall** | ***P*-value** |
| **Internal Validation**<br><br>**CT-RATE Dataset**<br>**(*n* = 1,564)** | RadFM | 0.442 | **0.399** | N/A | 0.315 | 0.059 | 0.170 | 0.038 | < 0.001 |
| | CT2Rep | **0.444** | 0.402 | N/A | 0.310 | 0.160 | 0.435 | 0.128 | |
| | M3D | 0.436 | 0.400 | N/A | 0.326 | 0.148 | **0.407** | 0.090 | |
| | CT-CHAT | 0.436<br>[0.425, 0.446] | 0.212<br>[0.208, 0.216] | 0.235<br>[0.182, 0.296] | **0.353**<br>**[0.344, 0.363]** | 0.165<br>[0.131, 0.202] | 0.356<br>[0.249, 0.480] | 0.148<br>[0.122, 0.177] | |
| | CT-CHAT (w/ nodule attributes) | N/A | N/A | N/A | N/A | 0.305 | 0.382 | 0.268 | |
| | MedVista3D | 0.474 | 0.252 | 0.349 | 0.386 | N/A | N/A | N/A | |
| | Reg2RG | 0.438<br>[0.428, 0.448] | 0.210<br>[0.206, 0.214] | 0.054<br>[0.043, 0.069] | 0.238<br>[0.235, 0.240] | 0.172<br>[0.129, 0.218] | 0.346<br>[0.242, 0.467] | 0.129<br>[0.097, 0.165] | |
| | T3D | N/A | N/A | N/A | 0.378 | **0.274** | 0.355 | 0.207 | |
| | CT-GRAPH | 0.406<br>[0.396, 0.416] | 0.198<br>[0.194, 0.201] | 0.052<br>[0.040, 0.067] | 0.251<br>[0.248, 0.254] | 0.255<br>[0.207, 0.304] | 0.286<br>[0.223, 0.353] | **0.258**<br>**[0.212, 0.307]** | |
| | BTB3D | 0.413<br>[0.406, 0.420] | 0.186<br>[0.183, 0.189] | 0.014<br>[0.011, 0.019] | 0.264<br>[0.259, 0.268] | 0.205<br>[0.165, 0.247] | 0.205<br>[0.162, 0.253] | 0.229<br>[0.183, 0.278] | |
| | **EXACT-CHAT** | 0.495<br>[0.485, 0.504] | 0.222<br>[0.218, 0.226] | **0.141**<br>**[0.111, 0.178]** | 0.328<br>[0.320, 0.336] | 0.310<br>[0.274, 0.347] | 0.410<br>[0.291, 0.541] | 0.371<br>[0.336, 0.408] | |
| **External Validation**<br><br>**RAD-ChestCT Dataset**<br>**(*n* = 3,630)** | RadFM | N/A | N/A | N/A | N/A | 0.049<br>[0.041, 0.058] | 0.302<br>[0.140, 0.387] | 0.054<br>[0.048, 0.060] | 0.056 |
| | CT2Rep | N/A | N/A | N/A | N/A | 0.133 | 0.299 | 0.139 | |
| | M3D | N/A | N/A | N/A | N/A | 0.113<br>[0.091, 0.138] | 0.269<br>[0.214, 0.330] | 0.080<br>[0.064, 0.098] | |
| | CT-CHAT | N/A | N/A | N/A | N/A | 0.142<br>[0.128, 0.156] | 0.209<br>[0.162, 0.257] | 0.186<br>[0.171, 0.202] | |
| | Merlin | N/A | N/A | N/A | N/A | 0.057<br>[0.049, 0.065] | 0.178<br>[0.146, 0.209] | 0.080<br>[0.075, 0.085] | |
| | BTB3D | N/A | N/A | N/A | N/A | 0.241<br>[0.215, 0.267] | 0.270<br>[0.242, 0.299] | 0.272<br>[0.243, 0.304] | |
| | Reg2RG | N/A | N/A | N/A | N/A | 0.113<br>[0.093, 0.134] | 0.277<br>[0.205, 0.355] | 0.082<br>[0.068, 0.098] | |
| | Hulu-Med | N/A | N/A | N/A | N/A | 0.279<br>[0.249, 0.309] | 0.398<br>[0.355, 0.441] | **0.254**<br>**[0.226, 0.284]** | |
| | CT-GRAPH | N/A | N/A | N/A | N/A | **0.270**<br>**[0.240, 0.300]** | **0.368**<br>**[0.326, 0.410]** | 0.243<br>[0.216, 0.270] | |
| | **EXACT-CHAT** | N/A | N/A | N/A | N/A | 0.289<br>[0.265, 0.313] | 0.469<br>[0.327, 0.546] | 0.298<br>[0.275, 0.321] | |
| **External Validation**<br><br>**MianYang Dataset**<br>**(n = 505)** | RadFM | 0.000<br>[0.000, 0.000] | 0.008<br>[0.008, 0.009] | 0.000<br>[0.000, 0.000] | 0.021<br>[0.020, 0.023] | 0.023<br>[0.009, 0.044] | 0.143<br>[0.052, 0.185] | 0.046<br>[0.021, 0.076] | < 0.001 |
| | M3D | 0.000<br>[0.000, 0.001] | 0.023<br>[0.022, 0.024] | 0.000<br>[0.000, 0.000] | 0.057<br>[0.055, 0.060] | 0.068<br>[0.027, 0.118] | 0.162<br>[0.063, 0.292] | 0.055<br>[0.019, 0.102] | |
| | CT-CHAT | 0.299<br>[0.288, 0.310] | **0.177**<br>**[0.174, 0.180]** | 0.003<br>[0.002, 0.006] | 0.299<br>[0.294, 0.303] | 0.073<br>[0.045, 0.104] | 0.119<br>[0.086, 0.152] | 0.088<br>[0.053, 0.128] | |
| | Merlin | 0.113<br>[0.101, 0.125] | 0.062<br>[0.060, 0.065] | 0.001<br>[0.000, 0.001] | 0.116<br>[0.114, 0.118] | 0.024<br>[0.021, 0.028] | 0.074<br>[0.013, 0.076] | 0.059<br>[0.057, 0.060] | |
| | BTB3D | 0.436<br>[0.425, 0.443] | 0.184<br>[0.180, 0.188] | 0.008<br>[0.005, 0.011] | **0.261**<br>**[0.256, 0.267]** | 0.162<br>[0.107, 0.222] | 0.190<br>[0.134, 0.252] | 0.218<br>[0.133, 0.315] | |
| | Reg2RG | 0.281<br>[0.269, 0.292] | 0.154<br>[0.150, 0.157] | **0.006**<br>**[0.004, 0.008]** | 0.209<br>[0.206, 0.212] | 0.086<br>[0.041, 0.145] | 0.169<br>[0.076, 0.292] | 0.066<br>[0.029, 0.119] | |
| | Hulu-Med | 0.154<br>[0.136, 0.172] | 0.106<br>[0.100, 0.112] | 0.003<br>[0.001, 0.006] | 0.180<br>[0.172, 0.188] | **0.175**<br>**[0.095, 0.266]** | **0.265**<br>**[0.135, 0.424]** | 0.176<br>[0.080, 0.287] | |
| | CT-GRAPH | **0.308**<br>**[0.295, 0.320]** | 0.175<br>[0.172, 0.178] | 0.006<br>[0.004, 0.010] | 0.221<br>[0.217, 0.225] | 0.200<br>[0.122, 0.295] | 0.293<br>[0.136, 0.449] | **0.198**<br>**[0.117, 0.314]** | |
| | **EXACT-CHAT** | 0.451<br>[0.441, 0.461] | 0.209<br>[0.205, 0.212] | 0.015<br>[0.010, 0.020] | 0.301<br>[0.297, 0.305] | 0.290<br>[0.221, 0.359] | 0.326<br>[0.221, 0.435] | 0.367<br>[0.307, 0.429] | |

**Supplementary Table 6 | Report generation performance before post-hoc refinement.** Performance of EXACT-CHAT and baseline models before post-hoc refinement on CT-RATE, RAD-ChestCT and MianYang. NLG metrics include BLEU-1, METEOR, CIDEr and ROUGE-L; clinical efficacy metrics include RadBERT-F1, RadBERT-Precision and RadBERT-Recall. NLG metrics were evaluated on CT-RATE and MianYang. For RAD-ChestCT, reference reports were unavailable, and clinical efficacy was evaluated by comparing abnormality labels extracted from generated reports with dataset-provided labels. Independently evaluated results are reported with 95% CIs estimated using 2,000 bootstrap resamples; results obtained from published studies are

shown without CIs. Red text indicates the highest value, blue text the second highest, and bold text the third highest for each metric within each dataset. *P* values were calculated using two-sided bootstrap hypothesis tests with 2,000 resamples and compare EXACT-CHAT with CT-GRAPH on CT-RATE, Hulu-Med on RAD-ChestCT and CT-GRAPH on MianYang for RadBERT-F1. CT, computed tomography; NLG, natural language generation; BLEU, bilingual evaluation understudy; METEOR, Metric for Evaluation of Translation with Explicit ORdering; CIDEr, Consensus-based Image Description Evaluation; ROUGE, Recall-Oriented Understudy for Gisting Evaluation; CI, confidence interval; N/A, not available.

|  | Models (Refined) | BLEU-1 | METEOR | CIDEr | ROUGE-L | RadBERT-F1 Score | RadBERT-Precision | RadBERT-Recall | *P*-value |
|---|---|---|---|---|---|---|---|---|---|
| **Internal Validation CT-RATE Dataset (*n* = 1,564)** | CT-CHAT | 0.508 [0.500, 0.515] | **0.228 [0.224, 0.231]** | 0.076 [0.055, 0.102] | 0.339 [0.331, 0.346] | 0.427 [0.384, 0.469] | 0.341 [0.301, 0.382] | 0.647 [0.589, 0.703] | <0.001 |
| | Reg2RG | **0.449 [0.444, 0.454]** | 0.232 [0.229, 0.235] | **0.016 [0.012, 0.020]** | **0.299 [0.295, 0.304]** | 0.275 [0.239, 0.311] | **0.218 [0.183, 0.254]** | **0.609 [0.566, 0.652]** | |
| | BTB3D | 0.381 [0.375, 0.388] | 0.207 [0.205, 0.209] | 0.012 [0.009, 0.015] | 0.263 [0.260, 0.267] | **0.279 [0.244, 0.313]** | 0.209 [0.181, 0.237] | 0.577 [0.521, 0.632] | |
| | **EXACT-CHAT** | 0.522 [0.516, 0.529] | 0.232 [0.229, 0.236] | 0.080 [0.062, 0.100] | 0.320 [0.313, 0.327] | 0.501 [0.456, 0.543] | 0.414 [0.368, 0.460] | 0.730 [0.677, 0.780] | |
| **External Validation RAD-ChestCT Dataset (*n* = 3,630)** | RadFM | N/A | N/A | N/A | N/A | 0.242 [0.223, 0.262] | 0.271 [0.240, 0.307] | 0.368 [0.344, 0.392] | < 0.001 |
| | M3D | N/A | N/A | N/A | N/A | 0.335 [0.312, 0.358] | 0.292 [0.269, 0.316] | 0.505 [0.473, 0.538] | |
| | CT-CHAT | N/A | N/A | N/A | N/A | **0.361 [0.339, 0.383]** | **0.336 [0.301, 0.372]** | 0.566 [0.534, 0.597] | |
| | Merlin | N/A | N/A | N/A | N/A | 0.329 [0.305, 0.353] | 0.314 [0.289, 0.340] | 0.480 [0.445, 0.515] | |
| | BTB3D | N/A | N/A | N/A | N/A | 0.323 [0.301, 0.344] | 0.267 [0.247, 0.286] | 0.539 [0.507, 0.571] | |
| | Reg2RG | N/A | N/A | N/A | N/A | 0.255 [0.236, 0.274] | 0.268 [0.242, 0.295] | 0.627 [0.608, 0.646] | |
| | Hulu-Med | N/A | N/A | N/A | N/A | 0.419 [0.394, 0.443] | 0.379 [0.355, 0.403] | 0.622 [0.588, 0.656] | |
| | **EXACT-CHAT** | N/A | N/A | N/A | N/A | 0.441 [0.416, 0.467] | 0.406 [0.380, 0.433] | **0.610 [0.576, 0.643]** | |
| **External Validation MianYang Dataset (*n* = 505)** | RadFM | 0.373 [0.360, 0.386] | 0.186 [0.182, 0.190] | 0.010 [0.005, 0.015] | **0.292 [0.286, 0.298]** | 0.196 [0.126, 0.268] | 0.218 [0.149, 0.292] | 0.401 [0.288, 0.515] | < 0.001 |
| | M3D | 0.366 [0.354, 0.378] | 0.166 [0.162, 0.170] | 0.009 [0.006, 0.012] | 0.247 [0.242, 0.252] | 0.206 [0.152, 0.268] | 0.202 [0.133, 0.298] | 0.563 [0.482, 0.644] | |
| | CT-CHAT | **0.443 [0.434, 0.452]** | **0.209 [0.206, 0.212]** | **0.016 [0.011, 0.022]** | 0.332 [0.326, 0.336] | **0.299 [0.234, 0.364]** | **0.235 [0.186, 0.286]** | **0.648 [0.540, 0.751]** | |
| | Merlin | 0.327 [0.312, 0.343] | 0.145 [0.140, 0.150] | 0.008 [0.005, 0.013] | 0.212 [0.206, 0.219] | 0.267 [0.200, 0.336] | 0.218 [0.167, 0.278] | 0.532 [0.419, 0.651] | |
| | Reg2RG | 0.451 [0.441, 0.461] | 0.200 [0.197, 0.203] | 0.017 [0.012, 0.023] | 0.270 [0.266, 0.274] | 0.232 [0.180, 0.290] | 0.179 [0.139, 0.226] | 0.569 [0.430, 0.683] | |
| | Hulu-Med | 0.352 [0.336, 0.366] | 0.163 [0.158, 0.168] | 0.011 [0.007, 0.016] | 0.242 [0.235, 0.249] | 0.327 [0.252, 0.403] | 0.254 [0.195, 0.320] | 0.629 [0.515, 0.740] | |
| | BTB3D | 0.424 [0.416, 0.433] | 0.226 [0.224, 0.229] | 0.009 [0.006, 0.012] | 0.291 [0.286, 0.295] | 0.256 [0.204, 0.309] | 0.187 [0.150, 0.226] | 0.674 [0.565, 0.778] | |
| | **EXACT-CHAT** | 0.503 [0.495, 0.512] | 0.223 [0.220, 0.226] | 0.024 [0.018, 0.031] | 0.308 [0.304, 0.311] | 0.410 [0.319, 0.497] | 0.338 [0.258, 0.422] | 0.667 [0.545, 0.786] | |

**Supplementary Table 7 | Report generation performance after disease-conditioned post-hoc refinement.** Performance of EXACT-CHAT and the baseline models listed for each dataset after disease-conditioned post-hoc refinement on CT-RATE, RAD-ChestCT and MianYang. Reports were refined using the same GPT-4.1 procedure, conditioned on disease predictions produced by the corresponding model. NLG metrics include BLEU-1, METEOR, CIDEr and ROUGE-L; clinical efficacy metrics include RadBERT-F1, RadBERT-Precision and RadBERT-Recall. NLG metrics were evaluated on CT-RATE and MianYang. For RAD-ChestCT, reference reports were unavailable, and clinical efficacy was evaluated by comparing abnormality labels extracted from generated reports with dataset-provided labels. Values are metric estimates with 95% CIs estimated using 2,000 bootstrap resamples. Red text indicates the highest value, blue text the second highest, and bold text the third highest for each metric within each dataset. *P* values were calculated using two-sided bootstrap hypothesis tests with 2,000 resamples and compare EXACT-CHAT with CT-CHAT on CT-RATE and Hulu-Med on RAD-ChestCT and MianYang for RadBERT-F1. CT, computed tomography; NLG, natural language generation; BLEU, bilingual evaluation understudy; METEOR, Metric for Evaluation of Translation with Explicit ORdering; CIDEr, Consensus-based Image Description Evaluation; ROUGE, Recall-Oriented Understudy for Gisting Evaluation; CI, confidence interval; N/A, not available.

| **Task** | **Models** | **Metric [95% CI]** | | ***P*-value** |
|---|---|---|---|---|
| | | **Dice** | **AUPR** | |
| Lung cancer segmentation<br><br>**22 development patients (18 training, 4 validation)**<br><br>**400 test cases** | BiomedParse-v2 | 0.118<br>[0.101, 0.136] | 0.176<br>[0.155, 0.196] | Dice: 0.001<br><br>AUPR: <0.001 |
| | SegMamba-Rand-Init | 0.138<br>[0.118, 0.160] | 0.140<br>[0.119, 0.161] | |
| | RWKV-UNet | 0.254<br>[0.233, 0.276] | 0.217<br>[0.196, 0.240] | |
| | **EXACT-Seg** | **0.291<br>[0.268, 0.315]** | **0.325<br>[0.299, 0.352]** | |

**Supplementary Table 8 | Lung tumour segmentation performance on NSCLC-Radiomics.** Models were evaluated on 400 held-out test patients. The development set comprised 18 training patients and four validation patients. The EXACT encoder was frozen during adaptation; SegMamba and RWKV-UNet were trained end-to-end, whereas BiomedParse-v2 was evaluated using pre-trained weights and the fixed prompt “lung tumor”. Values are patient-level means with 95% bootstrap confidence intervals. Bold indicates the best result. *P* values compare EXACT-Seg with RWKV-UNet using paired two-sided *t*-tests. AUPR, area under the precision–recall curve.

| **Task** | **Models** | **Metric [mean ± SD]** | | ***P*-value** |
|---|---|---|---|---|
| | | **C-index** | **2-year AUROC** | |
| Survival prediction | FMCIB | 0.551 ± 0.027 | 0.622 ± 0.065 | |
| **LUNG1 cohort** | **FMCIB + EXACT (global)** | 0.558 ± 0.028 | 0.638 ± 0.066 | C-index: 0.202 |
| **Five repeats of five-fold cross-validation** | **FMCIB + EXACT (tumour-pooled)** | **0.568 ± 0.033** | **0.646 ± 0.069** | AUROC: 0.251 |

**Supplementary Table 9 | Survival prediction using frozen image representations on NSCLC-Radiomics.** Overall survival analysis included 422 patients; two-year survival analysis included 420 patients after excluding two patients censored before 730 days. FMCIB features were evaluated alone or combined with EXACT global or tumour-pooled features, with both encoders frozen. Manual gross tumour volumes were used for FMCIB tumour localization and EXACT tumour pooling. Values are means ± SD of test-set performance across 25 training–test splits from five repeats of five-fold cross-validation. All methods used identical task-specific splits. Bold indicates the best result. *P* values compare the tumour-pooled and global fusion configurations using paired two-sided t-tests on test-set performance from the same training–test splits. C-index, concordance index; AUROC, area under the receiver operating characteristic curve; SD, standard deviation.

| Metric | A: EXACT | B: Without confinement at inference | C: Retrained without confinement | Δ (A−C) | *p* (A vs C) | Δ (A−B) | *p* (A vs B) |
|---|---|---|---|---|---|---|---|
| **Zero-shot classification; CT-RATE dataset, *n* = 1564** | | | | | | | |
| **AUROC** | **0.830 [0.822, 0.837]** | 0.671 [0.661, 0.680] | 0.811 [0.803, 0.819] | +0.018 [+0.013, +0.024] | <0.001 | +0.159 [+0.149, +0.168] | <0.001 |
| **F1 Score** | **0.836 [0.826, 0.844]** | 0.736 [0.726, 0.753] | 0.820 [0.813, 0.832] | +0.015 [+0.002, +0.023] | 0.023 | +0.100 [+0.080, +0.111] | <0.001 |
| **Accuracy** | **0.768 [0.758, 0.778]** | 0.641 [0.632, 0.660] | 0.748 [0.739, 0.762] | +0.020 [+0.004, +0.030] | 0.009 | +0.127 [+0.104, +0.138] | <0.001 |
| **Zero-shot localization; COVID-19 dataset, *n* = 20** | | | | | | | |
| **DSC** | **0.435 [0.337, 0.516]** | 0.012 [0.008, 0.016] | 0.030 [0.017, 0.045] | +0.405 [+0.305, +0.491] | < 0.001 | +0.423 [+0.332, +0.502] | < 0.001 |
| **HIT@5% (DSC definition)** | **0.950 [0.850, 1.000]** | 0.000 [0.000, 0.000] | 0.200 [0.050, 0.400] | +0.750 [+0.550, +0.900] | < 0.001 | +0.950 [+0.850, +1.000] | < 0.001 |
| **HIT@10% (DSC definition)** | **0.850 [0.700, 1.000]** | 0.000 [0.000, 0.000] | 0.100 [0.000, 0.250] | +0.750 [+0.500, +0.950] | < 0.001 | +0.850 [+0.700, +1.000] | < 0.001 |
| **AUPR** | **0.530 [0.436, 0.601]** | 0.042 [0.026, 0.059] | 0.250 [0.200, 0.294] | +0.280 [+0.223, +0.327] | < 0.001 | +0.488 [+0.398, +0.554] | < 0.001 |
| **HIT@5% (AUPR definition)** | **0.950 [0.850, 1.000]** | 0.350 [0.150, 0.550] | 0.900 [0.750, 1.000] | +0.050 [+0.000, +0.150] | 0.675 | +0.600 [+0.400, +0.800] | < 0.001 |
| **HIT@10% (AUPR definition)** | **0.900 [0.750, 1.000]** | 0.100 [0.000, 0.250] | 0.900 [0.750, 1.000] | +0.000 [+0.000, +0.000] | 1.000 | +0.800 [+0.600, +0.950] | < 0.001 |

**Supplementary Table 10 | Contribution of anatomical confinement to zero-shot classification and anomaly localization.** EXACT was compared with two variants to assess the contribution of its anatomical constraints: (1) removal of anatomical masks only at inference, with the trained weights unchanged, and (2) retraining and evaluation without anatomical confinement. Columns A, B and C correspond to EXACT, inference-only removal and retraining without confinement, respectively. Classification was evaluated on CT-RATE (*n* = 1,564 scans), with AUROC, F1 score and accuracy macro-averaged across 18 findings. Localization was evaluated on COVID-19 (*n* = 20 cases) using the voxel-wise mean of the high-resolution anomaly maps for atelectasis, lung opacity, mosaic attenuation pattern and consolidation. DSC was calculated after binarization at an absolute threshold of 0.005, whereas AUPR was calculated from continuous maps. HIT@5% and HIT@10% denote the proportions of cases with the specified metric exceeding 0.05 and 0.10, respectively. Values are point estimates with 95% confidence intervals in brackets. Differences (Δ) are calculated as EXACT minus the corresponding ablated variant; positive values favour EXACT. For the classification metrics, confidence intervals for each configuration were obtained using bootstrap resampling (1,000 resamples), and confidence intervals for the differences and P values using paired bootstrap resampling (2,000 resamples). Confidence intervals for DSC and AUPR, and confidence intervals and P values for the HIT rates, were obtained using paired bootstrap resampling (2,000 resamples). P values for DSC and AUPR were obtained using paired permutation tests (20,000 permutations). Bootstrap resampling for the per-configuration classification confidence intervals used a random seed of 0; all other resampling used a random seed of 42. AUROC, area under the receiver operating characteristic curve; AUPR, area under the precision–recall curve; DSC, Dice similarity coefficient; HIT, hit rate.